\documentclass[runningheads]{llncs}
\usepackage{graphicx}
\usepackage{amsmath,amssymb} %
\usepackage{color}
\usepackage{comment}

\usepackage{epsfig}
\usepackage{multirow}
\usepackage{pifont}
\usepackage{subcaption}
\usepackage{cite} 
\usepackage{adjustbox}
\usepackage{threeparttable}
\usepackage{xspace}
\usepackage{xcolor}
\usepackage{booktabs}
\usepackage{amsfonts}
\usepackage{wasysym}
\usepackage{placeins}
\usepackage{tabularx}
\usepackage{enumitem}
\usepackage{bbm}
\usepackage[hidelinks]{hyperref}
\usepackage[normalem]{ulem}

\usepackage{floatrow}
\usepackage{algorithm}
\usepackage{algorithmicx}
\usepackage[noend]{algpseudocode}

\newcommand{\ourmodel}{ILVM} %
\newcommand{\ourmodelshort}{\textsc{ILVM}}
\newcommand{\ourdataset}{\textsc{ATG4D}}
\newcommand{\nuscenes}{\textsc{nuScenes}}

\newcolumntype{s}{>{\centering\arraybackslash}X}

\makeatletter
\DeclareRobustCommand\onedot{\futurelet\@let@token\@onedot}
\def\@onedot{\ifx\@let@token.\else.\null\fi\xspace}

\makeatother

\DeclareMathOperator*{\argmin}{arg\,min}

\makeatletter
\newcommand{\printfnsymbol}[1]{%
  \textsuperscript{\@fnsymbol{#1}}%
}
\makeatother

\begin{document}
\pagestyle{headings}
\mainmatter
\def\ECCVSubNumber{4386}  %

\title{Implicit Latent Variable Model for Scene-Consistent Motion Forecasting} %

\titlerunning{Implicit Latent Variable Model for Scene-Consistent Motion Forecasting}

\authorrunning{S. Casas, C. Gulino, S. Suo, R. Liao, K. Luo, R. Urtasun}
\author{Sergio Casas\thanks{Denotes equal contribution}$^{1, 2}$, Cole Gulino\printfnsymbol{1}$^{1}$, Simon Suo\printfnsymbol{1}$^{1, 2}$, \\ Katie Luo$^{1}$, Renjie Liao$^{1, 2}$, Raquel Urtasun$^{1, 2}$%
\institute{Uber ATG$^1$, University of Toronto$^2$\\
\{sergio.casas, cgulino, suo, katie.luo, rjliao, urtasun\}@uber.com}
}
\maketitle

\begin{abstract}
In order to plan a safe maneuver an autonomous vehicle must accurately perceive its environment, and understand the interactions among traffic participants. 
In this paper, we aim to learn scene-consistent motion forecasts of complex urban traffic directly from sensor data.
In particular, we propose to characterize the joint distribution over future trajectories via an implicit latent variable model.
We model the scene as an interaction graph and employ powerful graph neural networks to learn a distributed latent representation of the scene.
Coupled with a deterministic decoder, we obtain trajectory samples that are consistent across traffic participants, achieving state-of-the-art results in motion forecasting and interaction understanding.
Last but not least, we demonstrate that our motion forecasts result in safer and more comfortable motion planning.

\end{abstract}

\section{Introduction}
Self driving vehicles (SDV) have the potential to make a broad impact in our society, providing a safer and more efficient solution to transportation. 
A critical component for autonomous driving is  the ability to perceive the world and  forecast all possible future instantiations  of the scene. 
3D perception algorithms  have improved incredibly fast in recent years \cite{qi2017pointnet, yang2018pixor, lang2019pointpillars, liang2019multi, zhou2019end, radar}, yielding very accurate  object detections surrounding the SDV. However, producing multi-modal motion forecasts that precisely capture multiple plausible futures consistently for all actors in the scene remains a very open problem. 

\begin{figure}[t]
    \centering
    \begin{subfigure}{0.48\textwidth}
      \centering
      \includegraphics[width=\linewidth, trim={0cm, 0cm, 3cm, 0cm}, clip]{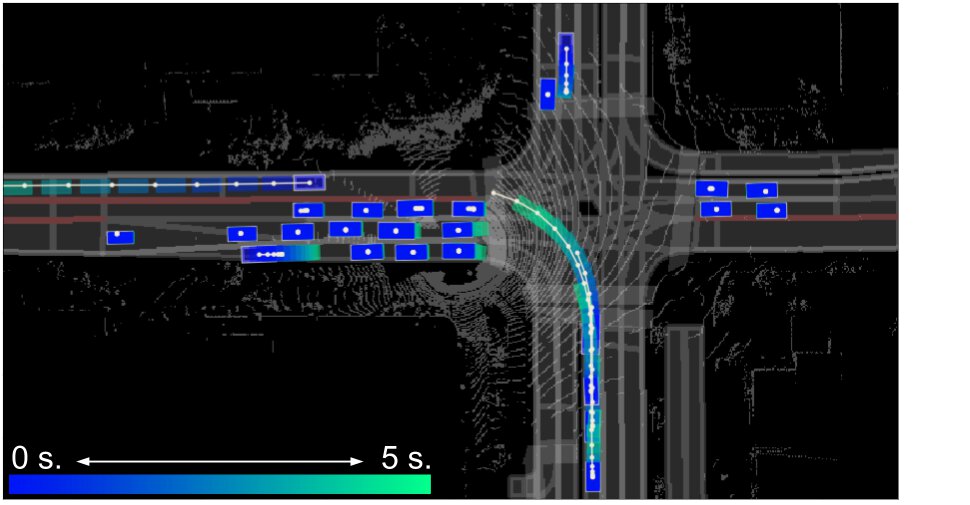}
      \captionsetup{width=\linewidth}
      \caption{Sample 1: protected left turn}
      \label{fig:motivating:protected}
    \end{subfigure}%
    \hspace{0.3em}
    \begin{subfigure}{.48\textwidth}
      \centering
      \includegraphics[width=\linewidth, trim={0cm, 0cm, 3cm, 0cm}, clip]{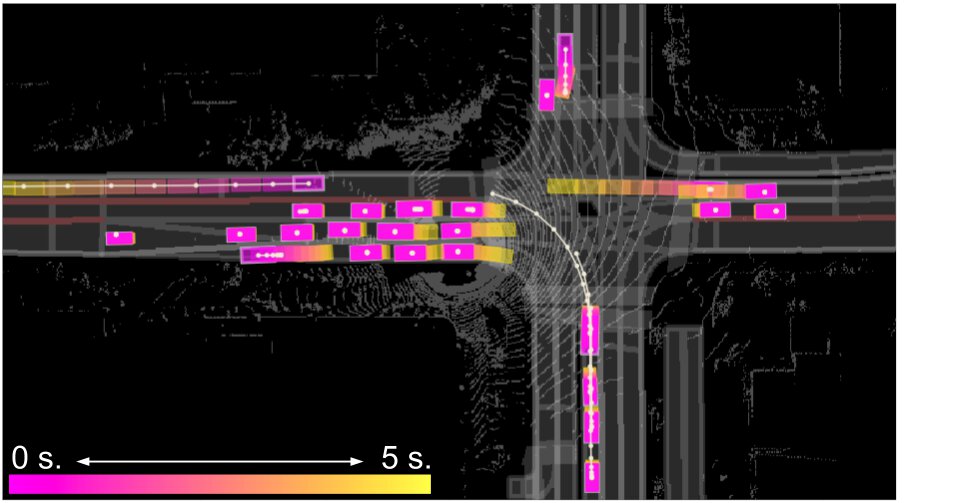}
      \captionsetup{width=\linewidth}
      \caption{Sample 2: horizontal traffic flow}
      \label{fig:motivating:straight}
    \end{subfigure}
    \centering
    \caption{\textbf{Two scene-consistent future trajectory samples from our model}. Ground truth trajectories are shown as white polylines.}
    \label{fig:motivating}
\end{figure}

The complexity is immense: the future is inherently uncertain as actor behaviors are influenced not only by their own individual goals and intentions but also by the other actors' actions.
For instance, an actor at an intersection may choose to turn right or go straight due to its own destination, and yield or go if the behavior of a nearby traffic participant is aggressive or conservative. 
Moreover, unobserved traffic rules such as the future traffic light states heavily affect the traffic (see  Fig.\ref{fig:motivating}).
It is clear that all these  aspects cannot be directly observed and require complex reasoning about the scene as a whole, including its geometry, topology and the interaction between multiple agents.

In an autonomy system, detections and motion forecasts for other actors in the scene are typically passed as obstacles to a motion-planner \cite{ratliff2006maximum,sadat2019jointly} in order to plan a safe maneuver.
Importantly, the distribution over future trajectories needs to cover the ground-truth for the plan to be safe, but also must exhibit low enough entropy such that a comfortable ride with reasonable progress is achieved.
Thus  in complex urban environments the SDV should reason about multiple futures separately \cite{hardy2013contingency, klingelschmitt2015managing, hubmann2018automated}, and plan proactively by understanding how its own  actions might influence other actors' behaviors \cite{okamoto2018ddt, henaff2019model}.
Furthermore, as self-driving vehicles get closer to full autonomy, closed-loop simulation is becoming increasingly critical not only for testing but also for training. In a self-driving simulator \cite{carla, autonovi-sim, martinez2017grand}, smart-actor models \cite{treiber2000congested, behrisch2011sumo, horizon-gail, Bhattacharyya_2018} 
are responsible for generating stochastic joint behaviors that are realistic at a scene-level, with actors obeying to underlying scene dynamics with   complex interactions.

These applications require learning a joint distribution over actors' future trajectories that characterizes how the scene might unroll as a whole. Since this is generally intractable, many motion forecasting approaches \cite{chai2019multipath,cui2018multimodal,casas2019spatially,Hong_2019_CVPR} assume marginal independence across actors' future trajectories,  failing to get scene-consistent futures.
Alternatively, auto-regressive formulations \cite{2019arXiv190501296R,tang2019multiple} model interactions at the output level, but require sequential sampling which results in slow inference and compounding errors \cite{ross2011reduction}.

To overcome these challenges, we propose a novel way to characterize the joint distribution over motion forecasts via an implicit latent variable model (ILVM). 
We aim to recover a latent space that can summarize all the unobserved scene dynamics given input sensor data. 
This is challenging given that
(i) modern roads present very complex geometries and topologies that make every intersection unique, 
(ii) the dynamic environment is only partially observed through sensor returns, and
(iii) the number of actors in a scene is variable.
To address these, we model the scene as an interaction graph \cite{kipf2018neural,casas2019spatially,ivanovic2019trajectron,li2020end}, where nodes are traffic participants. We then partition the scene latent space into a distributed representation among actors.
We leverage graph neural networks (GNN) \cite{battaglia2018relational} both to encode the full scene into the latent space as well as to decode latent samples into socially consistent future trajectories. We frame the decoding of all actors' trajectories as a deterministic mapping from the inputs and scene latent samples, making the latent variables capture all the stochasticity in our generative process. Furthermore, this allows us to perform  efficient inference via parallel sampling.

We show that our ILVM significantly outperforms the motion forecasting state-of-the-art in \ourdataset{} \cite{yang2018pixor} and \nuscenes{} \cite{caesar2019nuscenes}.  
We observe that our ILVM is able to generate scene-consistent samples (see Fig.~\ref{fig:motivating}) while producing less entropic joint distributions that also better cover the ground-truth.
Moreover,  when using our scene-consistent motion forecasts, a state-of-the-art motion planner \cite{sadat2019jointly} can plan safer and more comfortable trajectories.

\section{Related Work} \label{related}

\begin{figure}[t]
    \centering
    \begin{subfigure}[t]{.30\textwidth}
        \centering
        \includegraphics[width=.80\textwidth]{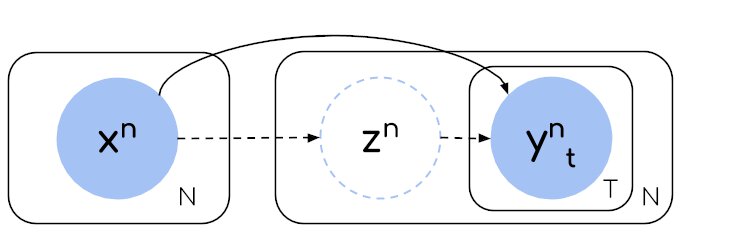}
        \caption{Independent output}
        \label{figure:graphical_models:independent}
    \end{subfigure}
    \begin{subfigure}[t]{.35\textwidth}
        \centering
        \includegraphics[width=\textwidth]{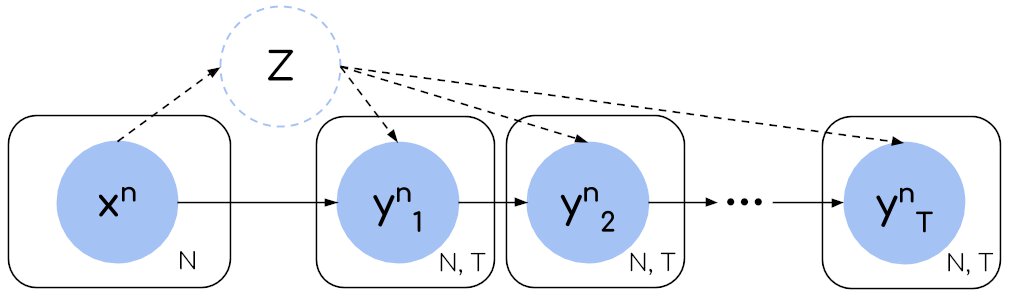}
        \caption{Social auto-regressive}
        \label{figure:graphical_models:autoregressive}
    \end{subfigure}
    \begin{subfigure}[t]{.30\textwidth}
        \centering
        \includegraphics[width=0.8\textwidth]{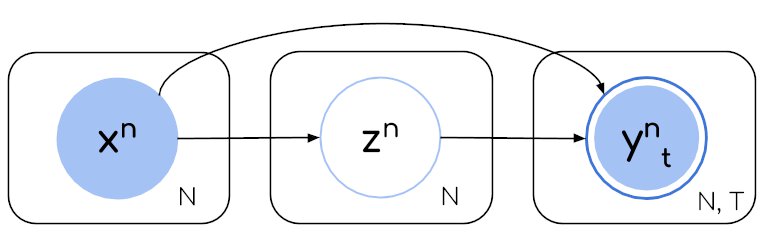}
        \caption{ILVM (Ours)}
        \label{figure:graphical_models:joint_latent}
    \end{subfigure}
    \caption{\textbf{Graphical models of trajectory distribution}. Dashed arrows/circles denote that only some approaches within the group use those components. Double circle in (c) denotes that it is a deterministic mapping of its inputs. 
    }
    \label{figure:graphical_models}
\end{figure}

In this section, we review recent advances in motion forecasting, with a  focus on realistic approaches that predict from sensor data, explicitly reason about the multi-modality of the output distribution, or model multi-agent interactions.

In traditional self-driving stacks, an object detection module is responsible for recognizing other traffic participants in the scene, followed by a motion forecasting module that predicts how the scene might unroll given the current state of each actor. However, the actor state is typically a very compact representation that includes pose, velocity, and acceleration. As a consequence, it is hard to incorporate uncertainty due to sensor noise or occlusion. 

We follow the works of \cite{luo2018fast, casas2018intentnet,zeng2019end}, which unified these two tasks by having a single fully convolutional backbone network predict both the current and future states for each pixel in a bird's eye view grid directly from a voxelized LiDAR point-cloud and semantic raster of an HD map. 
This approach naturally propagates uncertainty between the two tasks in the feature space, without the need of explicit intermediate representations. While these models reason about uncertainty in sensor observations, they neglect inherent uncertainty in the actors' future behavior. 
\cite{casas2019spatially, li2020end} add agent-agent interaction reasoning to this framework. \cite{casas2019spatially} introduces spatially-aware graph neural networks that aggregate features from neighboring actors in the scene to predict a single trajectory per actor with gaussian waypoints, assuming marginal independence across actors.
This approach is still limited in expressivity since (i) a uni-modal characterization of the future is insufficient for downstream motion planning to make safe decisions, and (ii) modeling the marginal distribution per actor cannot provide trajectory samples that are consistent across actors.

Another research stream \cite{alahi2016social, hoshen2017vain, ma2017forecasting, kipf2018neural, djuric2018motion, rhinehart2018r2p2, 2019arXiv190501296R, tang2019multiple, le2017coordinated} has focused on the problem of multi-agent trajectory prediction from perfect perception, i.e., assuming that the ground-truth past trajectory of all actors' is given. Unfortunately, this is not realistic in self-driving vehicles, which rely on imperfect perception with noise that translates into failures such as false positive and false negative detections and id switches in tracking. Nonetheless, these methods have proposed  output parameterizations that can predict multi-modal distributions over future trajectories, which are applicable to our end-to-end perception and prediction setting.

Various factorizations of the joint distribution over $N$ actors' trajectories $p(Y|X) = p(y_1, \cdots, y_N|x_1, \cdots, x_N)$ with different levels of independence assumptions have been proposed to sidestep the intractability of $p(Y|X)$.
The simplest approximation is to assume \emph{independent futures} across actors and time steps $p(Y|X) = \prod_n \prod_t p(y_n^t | X)$, as shown in Fig.~\ref{figure:graphical_models:independent}.
Some approaches directly regress the parameters of a mixture of Gaussians over time \cite{chai2019multipath, cui2018multimodal, lgn}, which provides efficient sampling but can suffer from low expressivity and unstable optimization. 
Non-parametric approaches \cite{phan2019covernet, kim2017probabilistic, jain2019discrete, ridel2020scene} have also been proposed to characterize the multi-modality of one actor's individual behavior. These approaches either score trajectory samples from a finite set \cite{phan2019covernet,dsdnet} with limited coverage or predict an occupancy grid  at different future horizons \cite{kim2017probabilistic, jain2019discrete, ridel2020scene}, which is very memory consuming.
\cite{rhinehart2018r2p2} proposed to learn a one-step policy that predicts the next waypoint based on the previous history, avoiding the time independence assumption.
Variational methods \cite{lee2017desire, Hong_2019_CVPR} inspired by \cite{kingma2013autoencoding, sohn2015learning} have also been proposed to learn an actor independent latent space to capture unobserved actor dynamics such as goals. 
Unfortunately, none of these methods can accurately characterize the joint distribution in interactive situations, since the generative process is independent per actor.

An alternative approach to better characterize the behavior of multiple actors jointly is \emph{autoregressive generation} with social mechanisms \cite{alahi2016social, 2019arXiv190501296R}, which predict the distribution over the next trajectory waypoint of each actor conditioned on the previous states of all actors $p(Y|X) = \prod_n \prod_t p\left(y_n^t | Y^{0:t-1}, X\right)$. This approach has been enhanced by introducing latent variables \cite{kipf2018neural, tang2019multiple, ivanovic2019trajectron}, as in Fig.~\ref {figure:graphical_models:autoregressive}.
In particular, \cite{kipf2018neural} introduces discrete latent variables to model pairwise relationships in an interaction graph, while in  \cite{tang2019multiple,ivanovic2019trajectron} they capture per-actor high-level actions.
Autoregressive approaches, however, suffer from compounding errors \cite{ross2011reduction, lamb_professor_forcing, osa2018algorithmic}. During training, the model is  fed the ground-truth $Y^{0:t-1}$, while during inference, the model must rely on approximate samples from the learned distribution.
While scheduled sampling \cite{bengio2015scheduled} has been proposed to mitigate this issue, the objective function underlying this method is improper \cite{huszar2015not} and pushes the conditional distributions $p(y_n^t | Y^{0:t-1})$ to model the marginal distributions $p(y_n^t)$ instead.
Moreover, these methods require sequential sampling, which is not amenable to real-time applications such as self-driving.

In contrast to previous works, we propose to model interaction in a scene latent space that captures all sources of uncertainty, and use a deterministic decoder to characterize an implicit joint distribution over all actors' future trajectories without any independence assumptions at the output level, as shown in Fig. \ref{figure:graphical_models:joint_latent}. This design features efficient parallel sampling, high expressivity and yields trajectory samples that are substantially more consistent across actors.

\section{Scene Level Reasoning for Motion Forecasting} \label{method}

In this section we introduce our approach to model \textbf{the joint distribution $P(Y|X)$ over $N$ actors' future trajectories} $Y = \begin{Bmatrix} y_1, y_2, \cdots, y_N \end{Bmatrix}$ given 
each actor's local context $X = \begin{Bmatrix} x_1, x_2, \cdots, x_N \end{Bmatrix}$ extracted from sensor data and HD maps. 
An actor's trajectory $y_n$ is composed of 2D waypoints over time $y_n^t$ in the coordinate frame defined by the actor's current position and heading.
In the following, we first explain our implicit latent variable model, then introduce our concrete architecture including the actor feature extraction from sensor data, and finally explain how to train our model in an end-to-end manner.

\subsection{Implicit Latent Variable Model with Deterministic Decoder} \label{latent_spagnn}

We formulate the generative process of future trajectories over actors with a latent variable model: 
{\small
$$P(Y|X) = \int_Z P(Y|X, Z) P(Z|X) dZ$$ 
}
where $Z$ is a latent variable that captures unobserved scene dynamics such as actor goals and style, multi-agent interactions, or future traffic light states.

We propose to use a \textbf{deterministic mapping $Y = f(X, Z)$} to implicitly characterize $P(Y | X, Z)$, instead of explicitly representing it in a parametric form. This approach allows us to avoid factorizing $P(Y|X,Z)$ (as in Fig.~\ref{figure:graphical_models:independent} or Fig.~\ref{figure:graphical_models:autoregressive}) and sidestep the associated shortcomings discussed in Section \ref{related}. In this framework, generating scene-consistent future trajectories $Y$ across actors is simple and  highly efficient, since it only requires one stage of parallel sampling:
\begin{enumerate}
    \item Draw latent scene samples from prior $Z \sim P(Z|X)$
    \item Decode with the deterministic decoder $Y = f(X, Z)$ 
\end{enumerate}

We emphasize that this modeling choice encourages
the latent $Z$ to capture \emph{all} stochasticity in our generative process. To this end, we leverage a \emph{continuous latent $Z$} for high expressivity.
This stands in contrast to previous methods \cite{kipf2018neural,tang2019multiple, ivanovic2019trajectron}, where discrete latent $Z$ are employed to model discrete high-level actions or pairwise interactions, and an explicit $P(Y|X,Z)$ to represent continuous uncertainty.

Producing a latent space that can capture all the uncertainties in any scenario is challenging: scenarios vary drastically in the number of actors $N$, the road topology as well as traffic rules. To mitigate this challenge, we propose to partition the scene latent as $Z=\begin{Bmatrix}z_1, z_2, \cdots, z_N\end{Bmatrix}$, obtaining a distributed representation where $z_n$ is anchored to actor $n$ in an interaction graph with  traffic participants as nodes.
The distributed representation has the benefit of naturally scaling the capacity of the latent space as the number of actors grow.
Furthermore, the anchoring gives the model an inductive bias that eases the learning of a scene latent space.
Intuitively, each latent $z_n$ encodes unobserved dynamics most relevant to actor $n$, including interactions with neighboring actors and traffic rules that apply in its locality.
We represent each $z_n$ as a diagonal multivariate gaussian
$
z_n \sim \mathcal{N}\left(\left[\mu_n^1(X), \cdots, \mu_n^D(X) \right], \text{diag}\left(\left[\sigma_n^1(X), \cdots, \sigma_n^D(X) \right]\right) \right)
$, as is common with variational models \cite{kingma2013autoencoding, sohn2015learning}. We emphasize that although factorized, the latent space is not marginally independent across actors since each $z_n$ is conditioned on all $x_1, \cdots, x_N$ as shown in the graphical model in Fig.~\ref{figure:graphical_models:joint_latent}.

\begin{figure}[t]
    \centering
    \includegraphics[width=\textwidth]{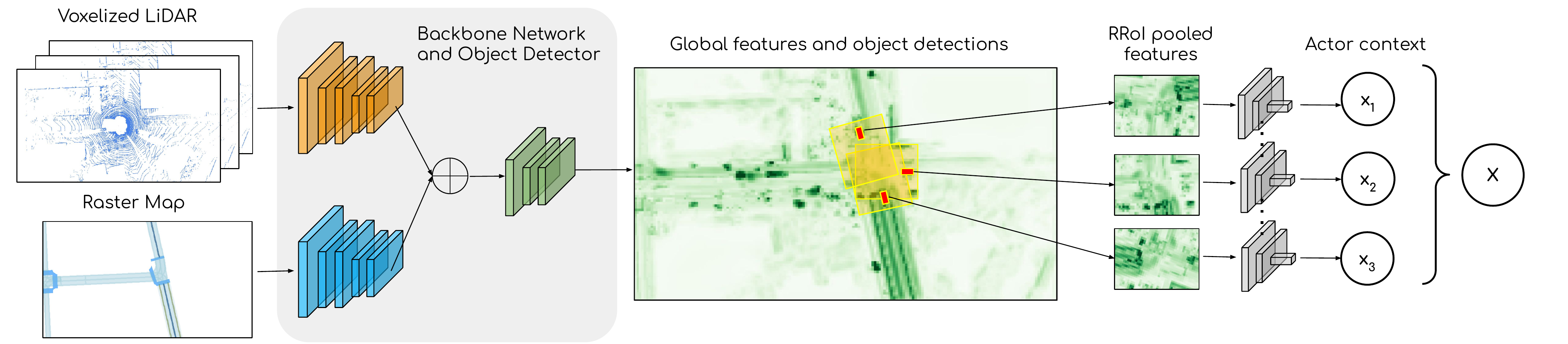}
    \caption{\textbf{Actor Feature Extraction}. Given LiDAR and maps, our backbone CNN detects the actors in the scene, and individual feature vectors per actor are extracted via RRoI Align \cite{ma2018arbitrary}, followed by a CNN with spatial pooling.}
    \label{fig:actor_features}
\end{figure}

Since integration over $Z$ is intractable, we exploit amortized variational inference \cite{kingma2013autoencoding, sohn2015learning}. 
By introducing an encoder distribution $Q(Z|X, Y)$ to approximate the true posterior $P(Z|X, Y)$, the learning problem can be reformulated as a maximization of the Evidence Lower BOund (ELBO). 
Please visit the supplementary for a more thorough description of variational inference.

\subsection{Joint Perception and Motion Forecasting Architecture}

\begin{figure}[t]
    \centering
    \includegraphics[width=\textwidth]{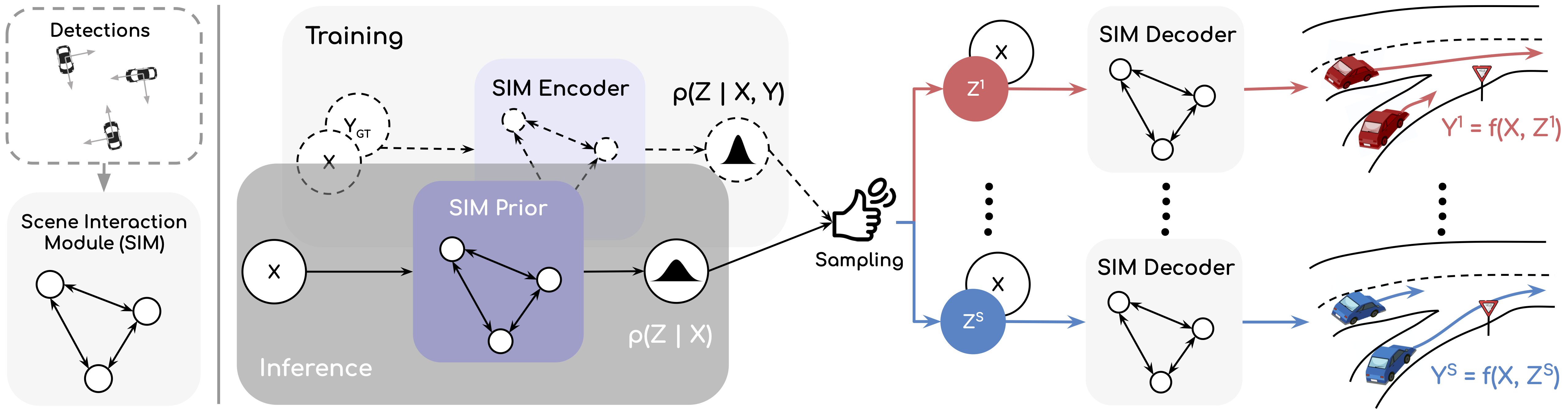}
    \caption{\textbf{Our Implicit Latent Variable Model} encodes the scene into a latent space, from which it can efficiently sample multiple future realizations in parallel, each with socially consistent trajectories.}
    \label{fig:model}
\end{figure}

Our architecture consists of an actor feature extractor that detects objects in the scene and provides rich representations of each actor (Fig.~\ref{fig:actor_features}), encoder/prior modules that infer a scene latent space at training/inference respectively, and a decoder that predicts the actors' future trajectories (Fig.~\ref{fig:model}). 
To implement the prior, encoder and decoder modules, we leverage a flexible scene interaction module (SIM) as our building block for relational reasoning (Alg.~\ref{alg:sim_main}). 


\medskip\noindent\textbf{Actor Feature Extractor:} 
Fig.~\ref{fig:actor_features} shows how we extract per actor features $X=\begin{Bmatrix}x_1, x_2, \cdots, x_N\end{Bmatrix}$ from raw sensor data and HD maps in a differentiable manner, such that perception and motion forecasting can be trained jointly end-to-end. 
We use a CNN-based perception backbone network architecture inspired by \cite{yang2018pixor,casas2018intentnet} to extract rich geometrical and motion features about the whole scene from a past history of voxelized LiDAR point clouds and a raster map. 
We then detect \cite{yang2018pixor} the traffic participants in the scene, and apply Rotated Region of Interest Align \cite{ma2018arbitrary} to the backbone features around each object detection, providing the local context for all actors, as proposed by \cite{casas2019spatially}. 
As mentioned at the beginning of Section \ref{method}, this will be the input to our motion forecasting module. This contrasts with previous approaches (e.g.,  \cite{djuric2018motion, chai2019multipath,2019arXiv190501296R,tang2019multiple}) that assume past trajectories for each actor are given. We refer the reader to the supplementary material for more details about our perception module, including the backbone architecture and detection parameterization. 


\medskip\noindent\textbf{Scene Interaction Module (SIM):}
This is a core building block of our encoder, prior, and decoder networks, as shown in Fig.~\ref{fig:model}. 
Once we have extracted individual actor features, we can frame the scene as a fully-connected interaction graph where each traffic participant is a node.
Inspired by \cite{casas2019spatially}, we use a spatially-aware graph neural network to model multi-agent dynamics, as described in Alg.~\ref{alg:sim_main}.
Our SIM performs a single round of message passing to update the nodes' representation, taking into account spatiotemporal relationships. 

\medskip\noindent\textbf{Encoder:}
To approximate the true posterior latent distribution $P(Z|X,Y)$, we introduce an approximate posterior $q_\phi(Z|X,Y)$, implemented by our SIM and parameterized by $\phi$. This network is also commonly known as recognition network, since it receives the target output variable $Y$ as an input, and thus it can \textit{recognize} the scene dynamics that are unobserved by the prior $p_{\gamma}(Z|X)$. Note that the encoder can only be used during training, since it requires access to the ground-truth future trajectories.
We initialize the node representations as $h_n$ = $\text{MLP}(x_n \oplus \text{GRU}(y_n))$, where $\oplus$ denotes concatenation along the feature dimension.
After running one round of message passing, the scene interaction module predicts the distribution over scene latent variables $Z=\begin{Bmatrix}z_1, z_2, \cdots, z_N\end{Bmatrix}$.
We stress that despite anchoring each partition of the scene latent to an actor, each individual $z_n$ contains information about the full scene, since each final node representation is dependent on the whole input $X$ because of the message propagation in the fully-connected interaction graph. 

\begin{algorithm*}
    \caption{Scene Interaction Module (SIM)} \label{alg:sim_main}
    \begin{flushleft}
    \textbf{Input:}
    Initial hidden state for all of the actors in the scene $\begin{Bmatrix}h_0, h_1, \cdots, h_N\end{Bmatrix}$. BEV coordinates (centroid and heading) of the detected bounding boxes $\begin{Bmatrix} c_0,c_1, ..., c_N \end{Bmatrix}$.

    \textbf{Output:} 
    Feature vector per node $\begin{Bmatrix}o_0, o_1, \cdots, o_N\end{Bmatrix}$.
    \end{flushleft}
    \begin{algorithmic}[1]
    \State Construct fully-connected interaction graph $G = (V, E)$ from detections
    \State Compute pairwise coordinate transformations $\mathcal{T} (c_{u}, c_{v})$, $\forall (u, v) \in E $
    \For {$ (u, v) \in E $} \Comment{Compute message for every edge in the graph in parallel}
        \State $m_{u \rightarrow v} = \text{MLP}\left(h_{u}, h_{v}, \mathcal{T} (c_{u}, c_{v}) \right)$
    \EndFor
    \For {$ v \in V $} \Comment{Update node states in parallel}
        \State $a_{v}=\text{MaxPooling}\left(\left\{m_{u \rightarrow v} : u \in \mathbf{N}(v)\right\}\right)$ \Comment{Aggregate messages from neighbors}
        \State $h'_{v}=\text{GRU}\left(h_{v}, a_{v}\right)$ \Comment{Update the hidden state}
        \State $o_{v}=\text{MLP}\left(h'_{v}\right)$ \Comment{Compute outputs}
    \EndFor
    \State\Return $\begin{Bmatrix}o_0, o_1, \cdots, o_N\end{Bmatrix}$
    \end{algorithmic}
\end{algorithm*}

\medskip\noindent\textbf{Prior:}
The prior network $p_{\gamma}(Z|X)$ is responsible for approximating the prior distribution of the scene latent variable $Z$ at inference time. 
Similar to the encoder, we model the scene-level latent space with our SIM, where the only difference is that the initial node representations in the graph propagation are  $h_n = \text{MLP}(x_n)$, since $y_n$ is not available at inference time. 

\medskip\noindent\textbf{Deterministic Decoder:}
Recall that our scene latent has been partitioned into a distributed representation $Z=\begin{Bmatrix}z_1, z_2, \cdots, z_N\end{Bmatrix}$. %
To leverage actor features and distributed latents from the whole scene, we parameterize the decoder with another SIM.
We can then predict the $s$-th realization of the future at a scene level via message passing, where each actor trajectory $y_n^s$ takes into account a sample from all the partitions of the scene latent $Z^s=\begin{Bmatrix}z_1^s, \cdots, z_n^s\end{Bmatrix}$ as well as all actors' features $X$, enabling reasoning about multi-agent interactions such as car following, yielding, etc. 
More precisely, given each actor context $x_n$, we initialize its node representation for the decoder graph propagation as $h^{s}_n = \text{MLP}(x_n \oplus z_n^s)$. 
After a round of message passing in our SIM, $h_n^{'s}$ contains an updated representation of actor $n$ that takes into account the underlying dynamics of the whole scene summarized in $Z^s$.
Finally, the $s$-th trajectory sample for actor $n$ is deterministically decoded $y_n^s = \text{MLP}(h_n^{'s})$ by the SIM output function, without additional sampling steps. The trajectory-level scene sample is simply the collection of all actor trajectories $Y^s = \begin{Bmatrix}y_1^s, \dots , y_N^s\end{Bmatrix}$. We can generate $S$ possible futures for all actors in the scene in parallel by batching $S$ scene latent samples.


In this fashion, our model implicitly characterizes the joint  distribution over actors' trajectories, achieving superior scene-level consistency. In the experiments section we ablate the design choices in the encoder, prior and decoder, and show that although all of them are important, the deterministic decoder is the key contribution towards socially-consistent trajectories.

\subsection{Learning}
\label{section:learning}

Our perception and prediction model can be trained end-to-end using stochastic gradient descent. In particular, we minimize a multi-task loss for detection and motion forecasting: %
$\mathcal{L} = \mathcal{L}_{\text{det}} + \lambda \cdot \mathcal{L}_{\text{forecast}}$

\medskip\noindent\textbf{Detection:} For the detection classification branch we employ a binary cross entropy loss with hard negative mining $\mathcal{L}_{\text{cla}}$. 
We select all positive examples from the ground-truth and 3 times as many negative examples. 
For box fitting, we apply a smooth $\ell_1$ loss $\mathcal{L}_{\text{reg}}$ to each of the 5 parameters $(x_i, y_i, w_i, h_i, \phi_i)$ of the bounding boxes anchored to a positive example $i$. The overall detection loss is a linear combination $\mathcal{L}_{\text{det}} = \mathcal{L}_{\text{cla}} + \alpha \cdot \mathcal{L}_{\text{reg}}$. 


\medskip\noindent\textbf{Motion Forecasting:}
We adapt the variational learning objective of the CVAE framework \cite{sohn2015learning} and optimize the evidence-based lower bound (ELBO) of the log-likelihood $\log P(Y | X)$. 
In our case, due to the deterministic decoder leading to an implicit distribution over $Y$, we use Huber loss $\ell_{\delta}$ as the reconstruction loss, and reweight the KL term with $\beta$ as proposed by \cite{higgins2017beta}:
\begin{align*}
\mathcal{L}_{\text{forecast}} = \sum_n^N \sum_t^T \ell_{\delta}(y_n^t - y_{n, GT}^t) + \beta \cdot \text{KL}\left(q_{\phi}\left(Z | X, Y_{GT} \right)|| p_{\gamma}\left(Z | X \right ) \right)
\end{align*}
where the first term minimizes the reconstruction error between all the trajectories in the scene $Y = \{y_n^t | \forall n, t\} = f_\theta(Z)$, $Z \sim q_{\phi}\left(Z | X, Y_{GT} \right)$ and their corresponding ground-truth $Y_{GT}$, and the second term brings the privileged \emph{posterior} $q_\phi(Z | X, Y_{GT})$ and the approximate \emph{prior} $p_\gamma(Z | X)$ distributions closer.

\section{Experimental Evaluation}

In this section, we first explain the metrics and baselines we use for evaluation. 
Next, we compare our model against state-of-the-art motion forecasting algorithms on predicting the future 5 second trajectories on two real-world datasets: \ourdataset{} \cite{yang2018pixor} and \nuscenes{} \cite{caesar2019nuscenes} (see supplementary for details).
Then, we measure the impact on  motion planning. 
Finally, we carry out an ablation study to understand which part of our model contributes the most.

\subsection{Scene Level Motion Forecasting Metrics}

Previous methods use sample quality metrics at the actor level such as the popular minimum/mean average displacement error (minADE/meanADE). However, these metrics only evaluate the quality of the underlying marginal distribution per actor. For instance, minADE takes the trajectory sample that best fits the ground-truth of each actor independently,  which does not measure the consistency between different actors’ sample trajectories and can be easily cheated by predicting high entropy distributions that cover all the space but are not precise.

\newcolumntype{s}{>{\centering\arraybackslash}X}
\begin{table}[t]
	\centering
	\begin{threeparttable}
        \begin{tabularx}{\textwidth}{
                        s |
                        l |  %
                        s s s s s  %
                        }
		    \toprule
                Type &
                Model & 
                $\mathrm{SCR}_{5s}$ (\%)  & min SFDE(m)  & min SADE(m)  & mean SFDE(m)  & mean SADE(m) \\ %
            \midrule
                \multirow{5}{1.5cm}{Indep. Output} 
                & SpAGNN \cite{casas2019spatially}          & {8.19}& {2.83} & {1.34}  & {4.37} & {1.92} \\
                & RulesRoad \cite{Hong_2019_CVPR}           & 6.66  & 2.71   & 1.32    & 4.21   & 1.84   \\
                & MTP \cite{cui2018multimodal}              & {3.98}& {1.91} & {0.95}  & {3.11} & {1.37} \\ 
                & MultiPath \cite{chai2019multipath}        & {4.41}& {1.97} & {0.95}  & {3.14} & {1.36} \\
                & R2P2-MA \cite{rhinehart2018r2p2}          & {4.63}& {2.13} & {1.09}  & {3.27} & {1.49} \\
                \hline
                \multirow{4}{1.5cm}{Social Auto-regressive} 
                & SocialLSTM \cite{alahi2016social}         & {6.13}& {2.75} & {1.38}  & {4.05} & {1.83} \\
                & NRI \cite{kipf2018neural}                 & 7.00	& 2.68   & 1.43    & 3.81   & 1.74   \\
                & ESP \cite{2019arXiv190501296R}            & {2.67}& {1.91} & {0.97}  & {2.84} & {1.29} \\
                & MFP \cite{tang2019multiple}	            & {5.15}& {2.35} & {1.13}  & {3.35} & {1.45} \\
            \midrule
                & \ourmodelshort{}   				     & \textbf{0.70}  & \textbf{1.53} & \textbf{0.76} & \textbf{2.27} & \textbf{1.02} \\
	        \bottomrule
		\end{tabularx}
    \end{threeparttable}
    \caption{\textbf{[\ourdataset] Scene-level motion forecasting} ($S=15$ samples)}
	\label{table:tor4d_main}
\end{table}

We propose scene-level sample quality metrics to evaluate how well the models capture the joint distribution over future outcomes. 
To this end, we define a scene-level counterpart of the popular minimum/mean average displacement error. We emphasize that in this context, each scene sample $s \in 1, ..., S$ is a collection of $N$ future trajectories, one for each actor in the scene.
{\small
\begin{align*}
\mathrm{minSADE} & = \min_{ s \in 1 \dots S} \frac{1}{N T} \sum_{n=1}^{N} 
\sum_{t=1}^T||y^t_{n, GT} - y^t_{n, s}||^2\\
\mathrm{meanSADE} & = \frac{1}{N TS} \sum_{s=1}^S
\frac{}{}\sum_{n=1}^{N} 
\sum_{t=1}^T||y^t_{n, GT} - y^t_{n, s}||^2
\end{align*}
}
We also compute their final counterparts minSFDE and meanSFDE, which evaluate only the motion forecasts at the final timestep (i.e. at 5 seconds). 

Furthermore, to evaluate the consistency of the motion forecasts we propose to measure the scene collision rate ($\mathrm{SCR}$). It measures the percentage of trajectory samples that collide with any other trajectory in the same scene sample $s$. Two trajectory samples are considered in collision if the overlap between their future bounding boxes at any time step is higher than a small IOU threshold $\varepsilon_{IOU}$. To compute this, we first obtain the bounding boxes for future time steps $\{b^t_{i, s}\}$. The size of the bounding boxes are the same as their object detections and the future headings are extracted by finite differences on the trajectory samples.
{\small
$$
\mathrm{SCR}_T = \frac{1}{NS} \sum_{s=1}^S
\frac{}{}\sum_{i=1}^{N} \min \left(1, \sum_{j>i}^{N} 
\sum_{t=1}^T \mathbbm{1} \left[IoU(b^t_{i, s}, b^t_{j, s}) > \varepsilon_{IOU}\right] \right)
$$
}
Finally, to perform a fair comparison in motion forecasting metrics, which are evaluated on true positive detections, we follow \cite{casas2019spatially} and operate the object detector at 90\% recall point for all models in \ourdataset{} and 80\% in \nuscenes{}.

\begin{table}[t]
	\centering
	\begin{threeparttable}
        \begin{tabularx}{\textwidth}{
                        s |
                        l |  %
                        s 
                        s s s s  %
                        }
            \toprule
                Type  &
                Model & 
                $\mathrm{SCR}_{5s}$ (\%)  & min SFDE(m)  & min SADE(m)  & mean SFDE(m)  & mean SADE(m)  \\ %
            \midrule
            \multirow{5}{1.5cm}{Indep. Output} 
            & SpAGNN \cite{casas2019spatially}    & 7.54    & 2.07       & 1.00     & {3.85}    & {1.82}  \\
            & RulesRoad \cite{Hong_2019_CVPR}     & 5.67    &  2.10      & 1.01     & 3.55      & 1.67    \\
            & MTP \cite{cui2018multimodal}        & 8.68    & 1.86       & 0.91     & 3.86      & {1.85}  \\ 
            & MultiPath \cite{chai2019multipath}  & {7.31}  & {2.01}     & {0.95}   & {3.50}    & {1.65}  \\
            & R2P2-MA \cite{rhinehart2018r2p2}    & {4.56}  & {2.25}     & {1.08}   & {3.47}    & {1.67}  \\
            \hline
            \multirow{4}{1.5cm}{Social Auto-regressive}
            & SocialLSTM \cite{alahi2016social}   & 6.45    & 2.71       & 	1.33     & 4.20    & 2.05 \\
            & NRI \cite{kipf2018neural}           & 5.98	& 2.54	     & 1.28	     & 3.91	    & 1.88  \\
            & ESP \cite{2019arXiv190501296R}      & {5.09}  & {2.16}     & {1.07}    & {3.46}   & {1.67} \\
            & MFP \cite{tang2019multiple}	      & {4.94}  & {2.74}     & {1.30}    & {4.11}   & {1.95} \\
        \midrule
            & \ourmodelshort{}   				   & \textbf{1.91} & \textbf{1.84}  & \textbf{0.86}  & \textbf{2.99} & \textbf{1.43}    \\
	        \bottomrule
		\end{tabularx}
    \end{threeparttable}
	\caption{\textbf{[\nuscenes{}] Scene-level motion forecasting} ($S=15$ samples)}
	\label{table:nuscenes_main}
\end{table}

\subsection{Baselines}\label{section:baselines}
In this section, we discuss the state of the art motion forecasting models that we use as baselines. It is important to note that most baselines are designed for motion forecasting given perfect perception, i.e., ground-truth past trajectories. However, this is not realistic in self-driving vehicles, which rely on imperfect noisy perception. Thus, we adapt them to the realistic setting by replacing their past trajectory encoders with our extracted actor features (see Fig.~\ref{fig:actor_features}) and training end-to-end with our perception backbone (see supplementary for details).


\medskip\noindent\textbf{Independent output:} We benchmark against \textsc{SpAGNN} \cite{casas2019spatially}, \textsc{MTP} \cite{cui2018multimodal}, \textsc{MultiPath} \cite{chai2019multipath}, \textsc{RulesRoad} \cite{Hong_2019_CVPR}, and \textsc{R2P2-MA} \cite{rhinehart2018r2p2}. Since the trajectory sampling process from these models is independent per actor, we define a scene sample $s$ by drawing one sample for each actor in the scene.

\medskip\noindent\textbf{Social auto-regressive:}
We compare against \textsc{SocialLSTM} \cite{alahi2016social}, \textsc{ESP} \cite{2019arXiv190501296R}, \textsc{MFP} \cite{tang2019multiple}, and \textsc{NRI} \cite{kipf2018neural}.
It is worth sharing that for these baselines to achieve competitive results we had to perturb the ground-truth trajectories with white noise during training. 
This is because these models suffer from a distributional shift between training and inference, as explained in Section \ref{related}.
We note that white noise was more effective than teacher forcing \cite{lamb_professor_forcing} or scheduled sampling \cite{bengio2015scheduled}.

\subsection{Motion Forecasting Results}
\label{sec:prediction_results}

Experimental results for motion forecasting in the \ourdataset{} dataset (with $S=15$ samples) are shown in Table \ref{table:tor4d_main}. Our \textsc{ILVM} outperforms the baselines across all metrics. Very notably, it \textbf{achieves a 75\% reduction in collision rate} with respect to the strongest baseline in this metric (\textsc{ESP} \cite{2019arXiv190501296R}), thus highlighting the better characterization of the joint distribution across actors (which also translates into scene-consistent samples). Our model is also much more precise (20\% reduction in  meanSFDE) while exhibiting better coverage of the ground-truth data (19\% reduction in minSFDE). We include an analysis of how the minSADE and minSFDE vary across different number of samples $S$ in the supplementary.

\begin{figure}[t]
    \centering
    \begin{tabular} {@{}c@{\hspace{.1em}}c@{\hspace{.5em}}c@{\hspace{.5em}}c}
        {} & \textbf{MultiPath} & \textbf{ESP} & \textbf{Ours (ILVM)} \\
        \rotatebox[origin=c]{90}{\textbf{Sample 1}} &
        \raisebox{-0.5\height}{\includegraphics[width=0.325\linewidth, trim={3cm, 3cm, 3cm, 1cm}, clip]{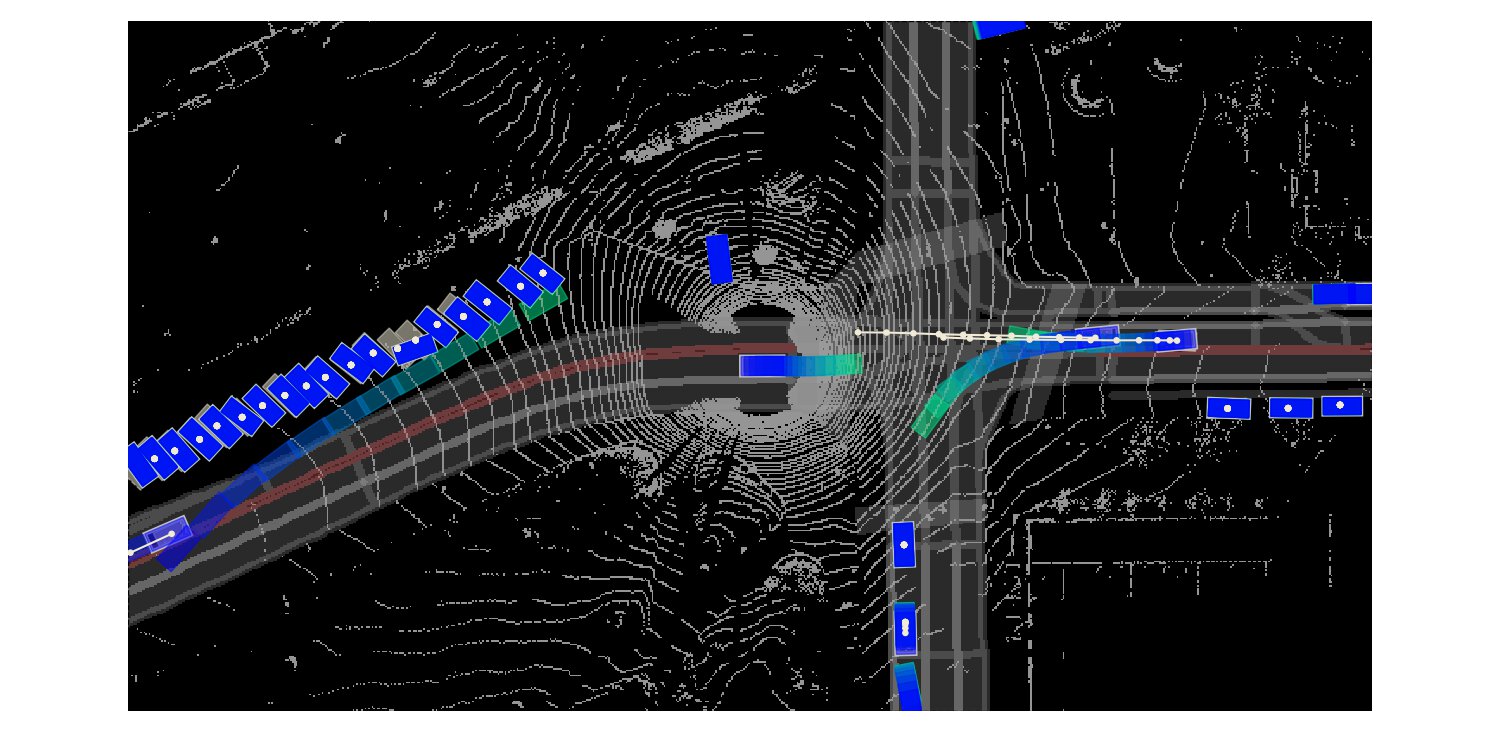}} &
        \raisebox{-0.5\height}{\includegraphics[width=0.325\linewidth, trim={3cm, 3cm, 3cm, 1cm}, clip]{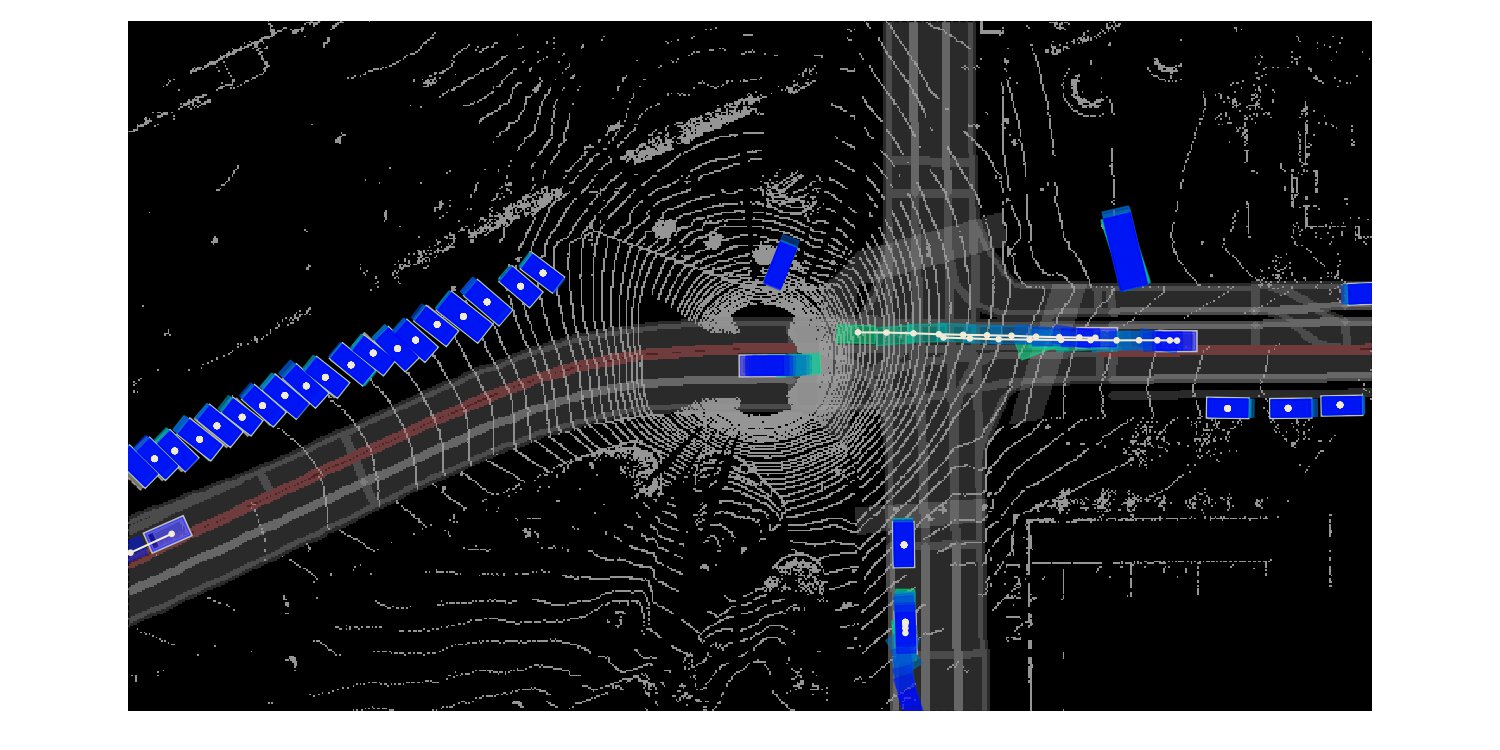}} &
        \raisebox{-0.5\height}{\includegraphics[width=0.325\linewidth, trim={3cm, 3cm, 3cm, 1cm}, clip]{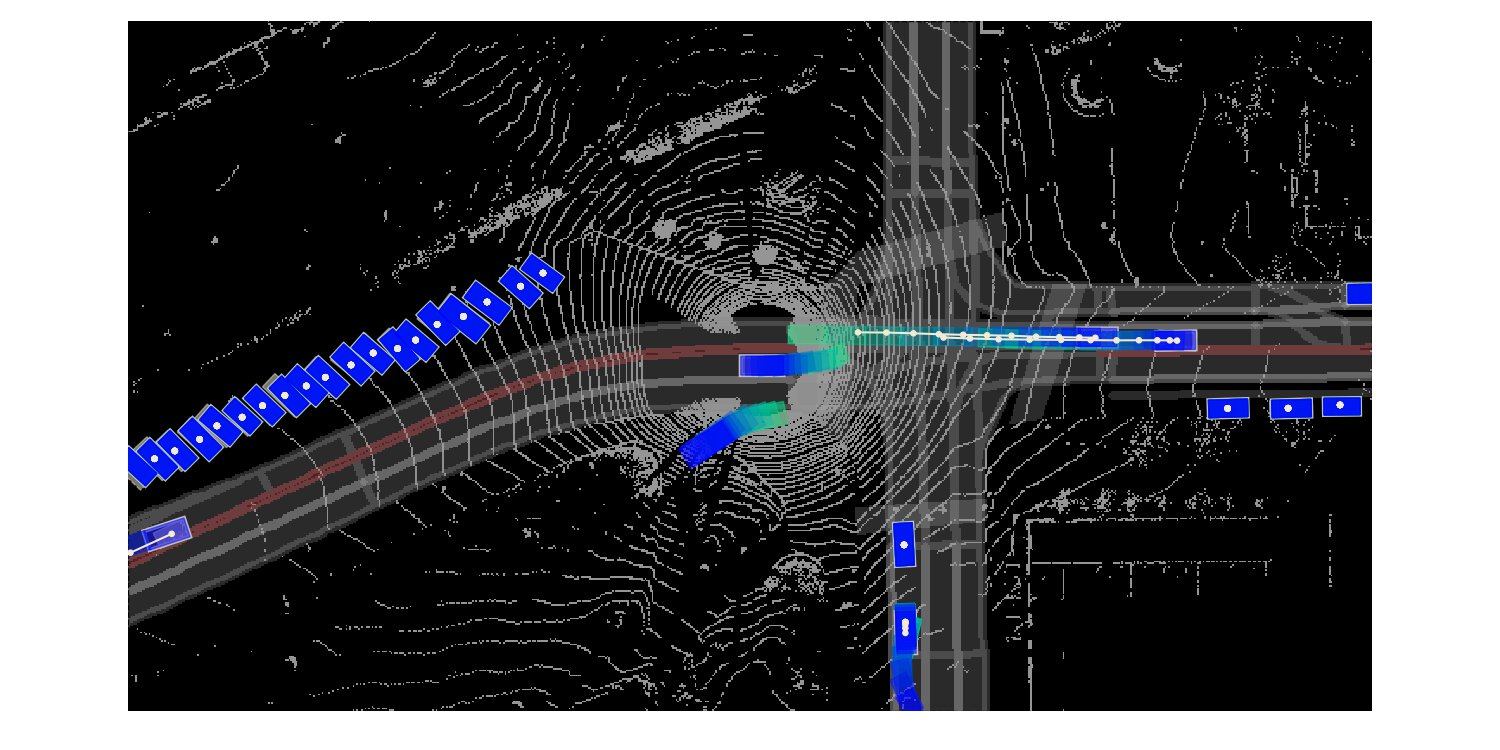}} \vspace{.5em} \\
        \rotatebox[origin=c]{90}{\textbf{Sample 2}} &
        \raisebox{-0.5\height}{\includegraphics[width=0.325\linewidth, trim={3cm, 3cm, 3cm, 1cm}, clip]{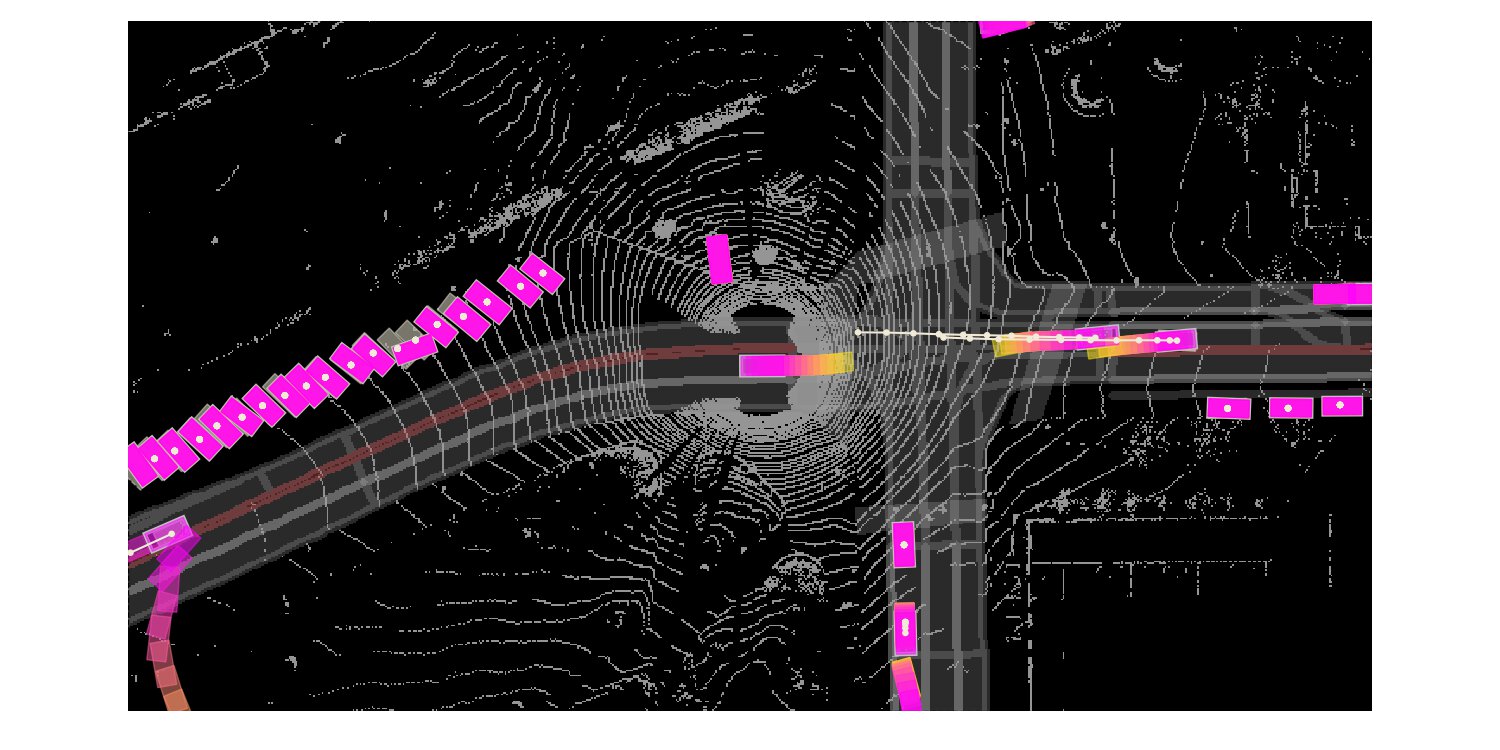}} &
        \raisebox{-0.5\height}{\includegraphics[width=0.325\linewidth, trim={3cm, 3cm, 3cm, 1cm}, clip]{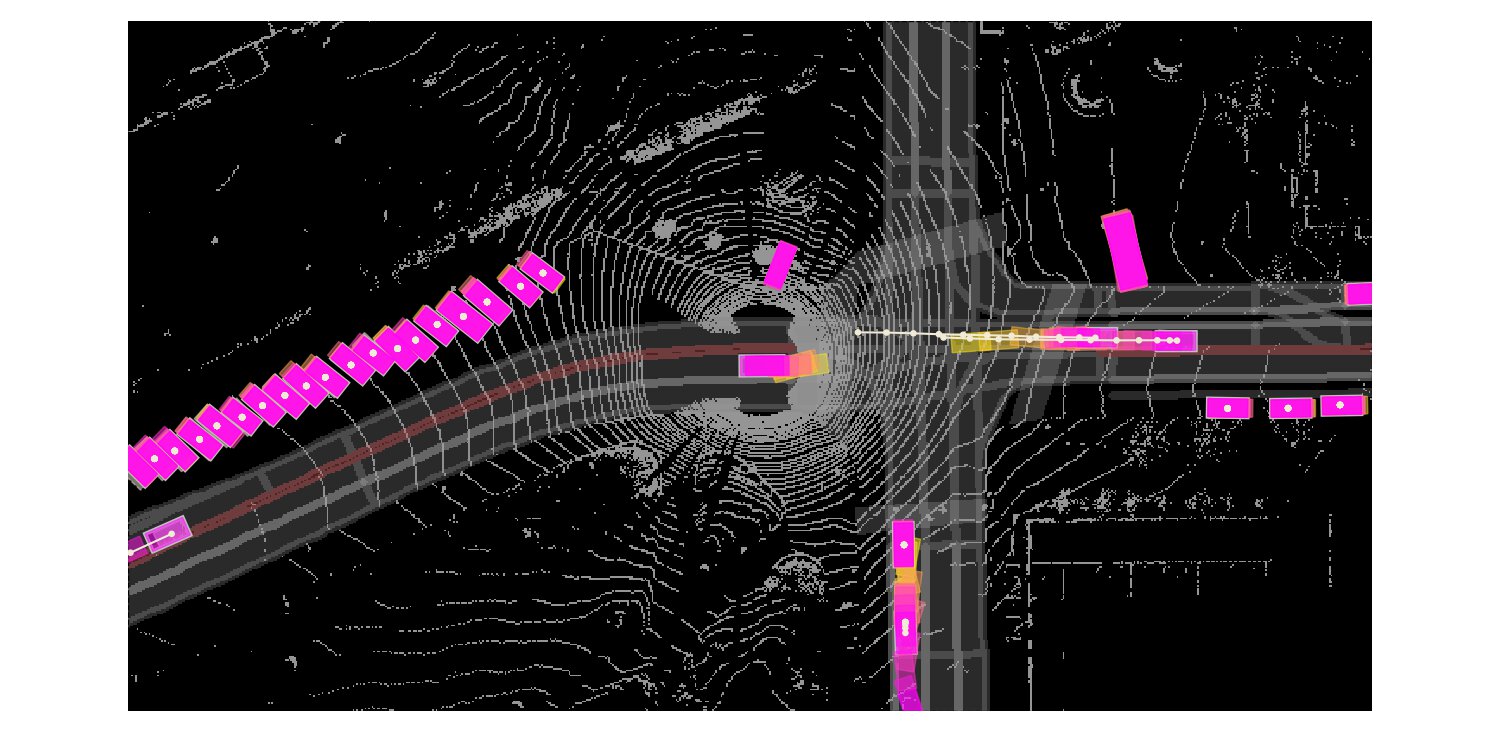}} &
        \raisebox{-0.5\height}{\includegraphics[width=0.325\linewidth, trim={3cm, 3cm, 3cm, 1cm}, clip]{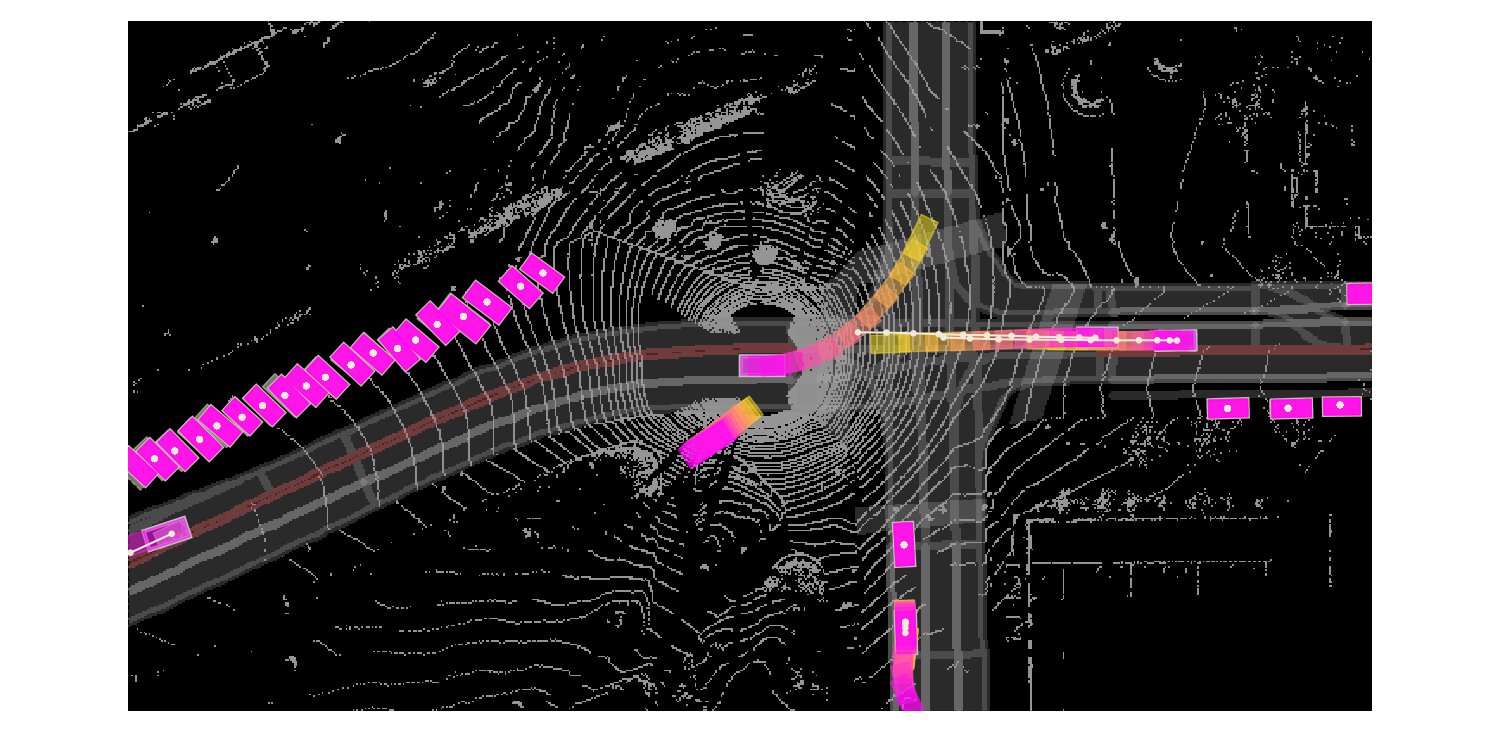}} \\
    \end{tabular}
    \caption{\textbf{Scene-level samples}. Our latent variable model captures underlying scene dynamics at the intersection level (i.e. yield vs. go)
    }
    \label{fig:scene_samples_1}
\end{figure}

Fig.~\ref{fig:scene_samples_1} shows individual samples. We heuristically select the two most distinct samples for visualization to show diverse realizations of the future. The baseline models capture variations in individual actors' future, but do not capture the yielding interaction at the intersection, which our model does. In addition, Fig.~\ref{fig:qualitative_prediction} showcases the full distribution learned by the models. More concretely, this plot shows a Monte Carlo estimation of the marginal distribution per actor, where 50 samples are drawn from each model. Transparency in the plots illustrates the probability density at a given location.
These examples support the same conclusion taken from the quantitative results and  highlight the ability of our model to understand complex road geometries and the multi-modal behaviors they induce. This is particularly interesting since all models share the same representation of the environment and backbone architecture. 

To show that our improvements generalize to a dataset with a different distribution of motions and road topologies, we validate our method on \nuscenes{}. Table \ref{table:nuscenes_main}, shows that \ourmodelshort{} brings improvements over the baselines across all metrics. In particular, we observe significant gains in scene-consistency ($\mathrm{SCR}$) and precision metrics ($\mathrm{meanSADE}$ and $\mathrm{meanSFDE}$).

\subsection{Motion Planning Results}
\begin{figure}[t]
    \centering
    \begin{tabular} {@{}c@{\hspace{.1em}}c@{\hspace{.5em}}c@{\hspace{.5em}}c}
        {} & \textbf{MultiPath} & \textbf{ESP} & \textbf{Ours (ILVM)} \\
        \rotatebox[origin=c]{90}{\textbf{Scenario 1}} &
        \raisebox{-0.5\height}{\includegraphics[width=0.325\linewidth, trim={3.5cm, 3cm, 3cm, 1cm}, clip]{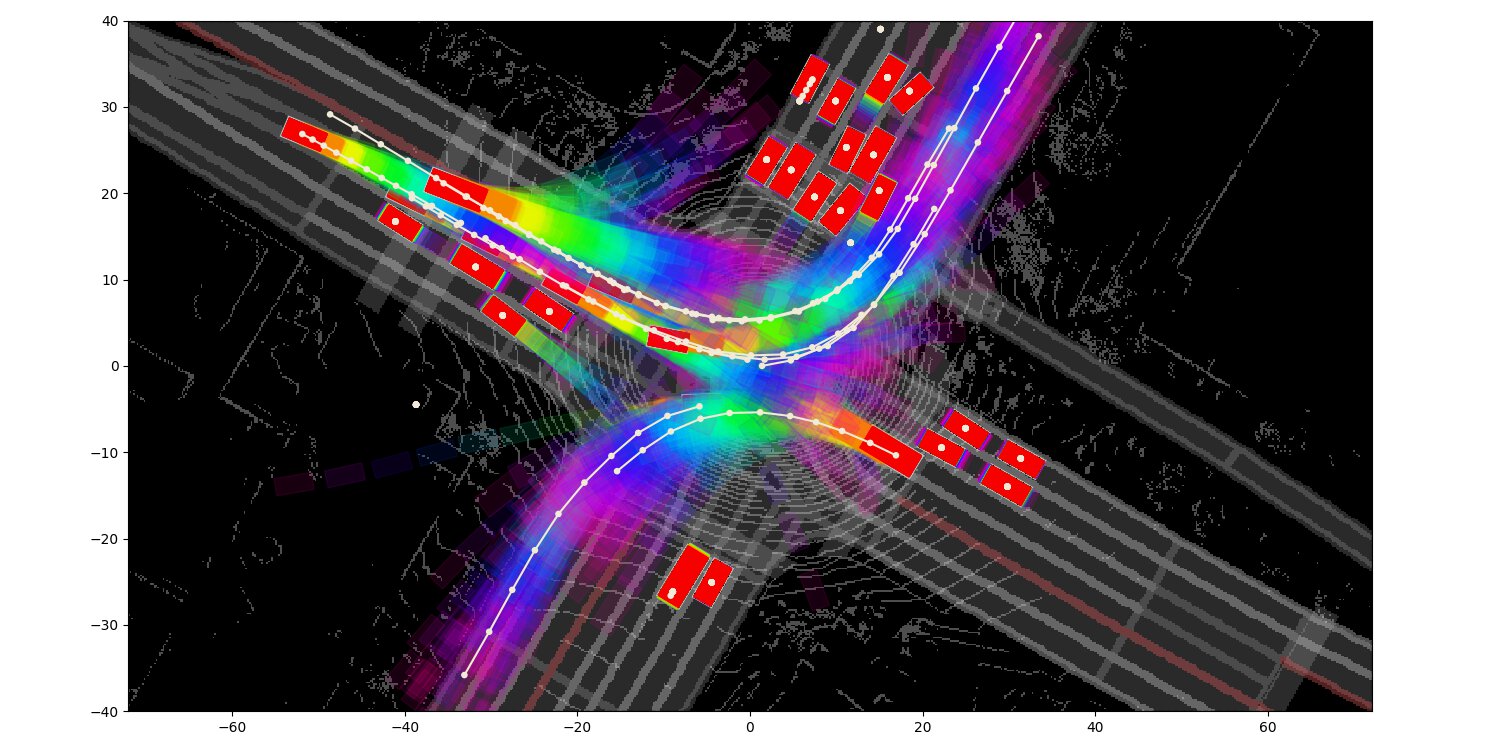}} &
        \raisebox{-0.5\height}{\includegraphics[width=0.325\linewidth, trim={3.5cm, 3cm, 3cm, 1cm}, clip]{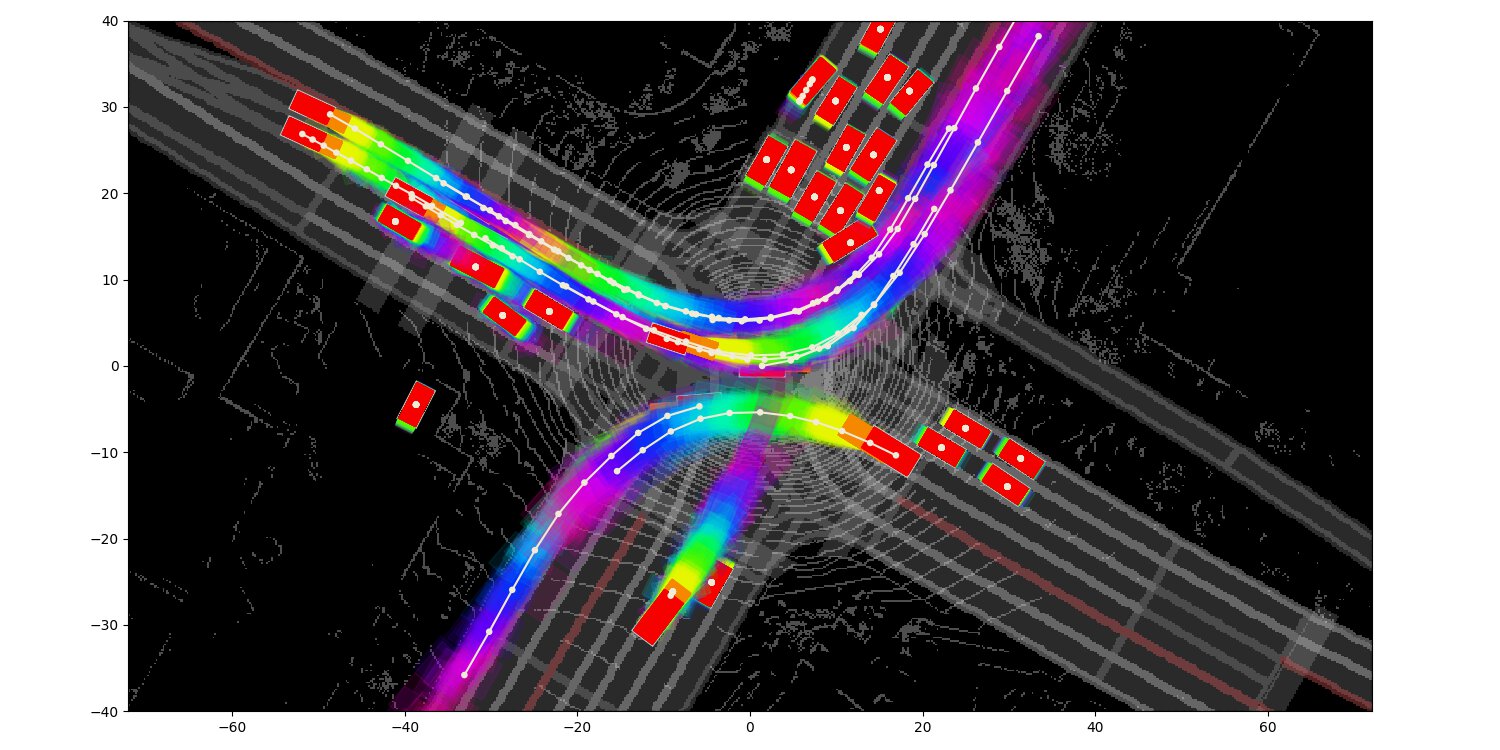}} &
        \raisebox{-0.5\height}{\includegraphics[width=0.325\linewidth, trim={3.5cm, 3cm, 3cm, 1cm}, clip]{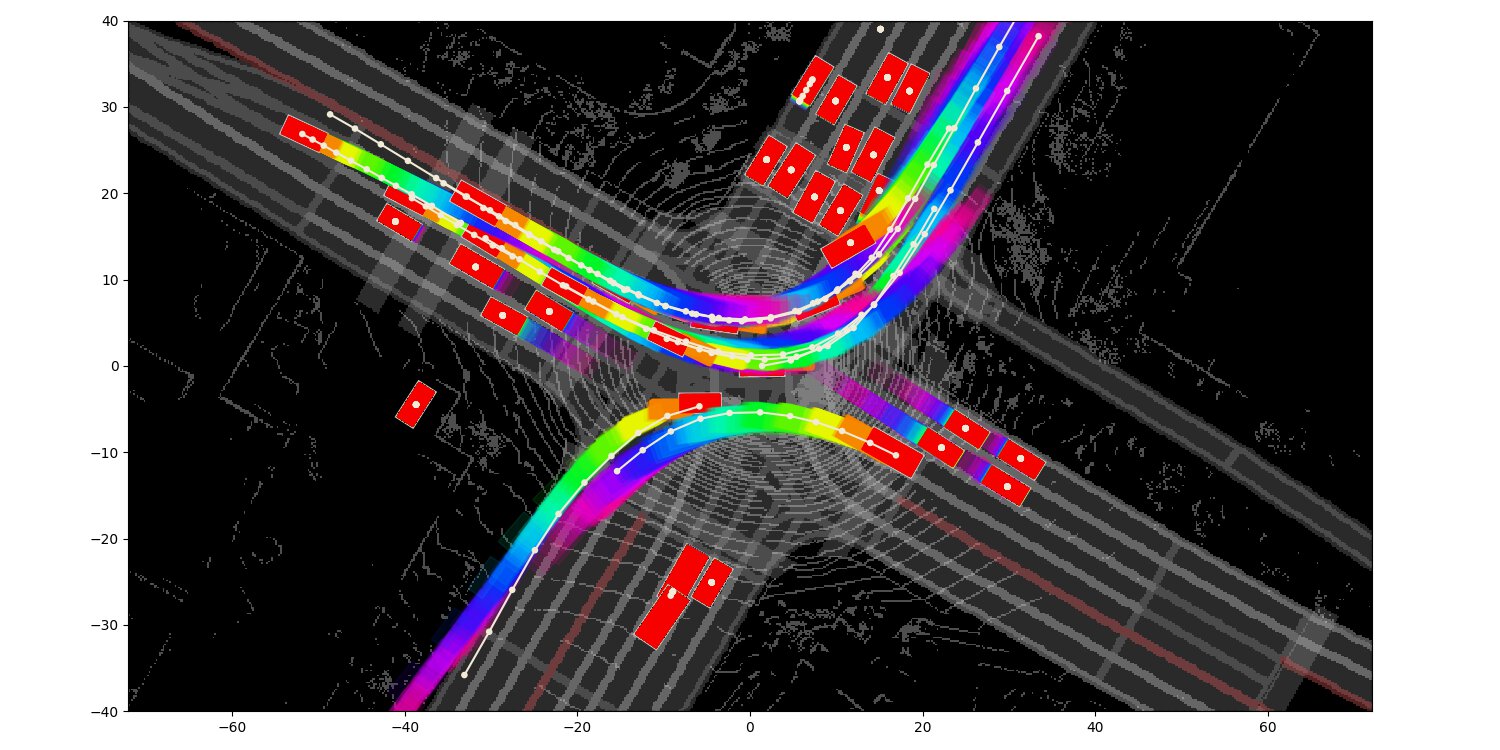}} \vspace{.5em} \\
        \rotatebox[origin=c]{90}{\textbf{Scenario 2}} &
        \raisebox{-0.5\height}{\includegraphics[width=0.325\linewidth, trim={3.5cm, 3cm, 3cm, 1cm}, clip]{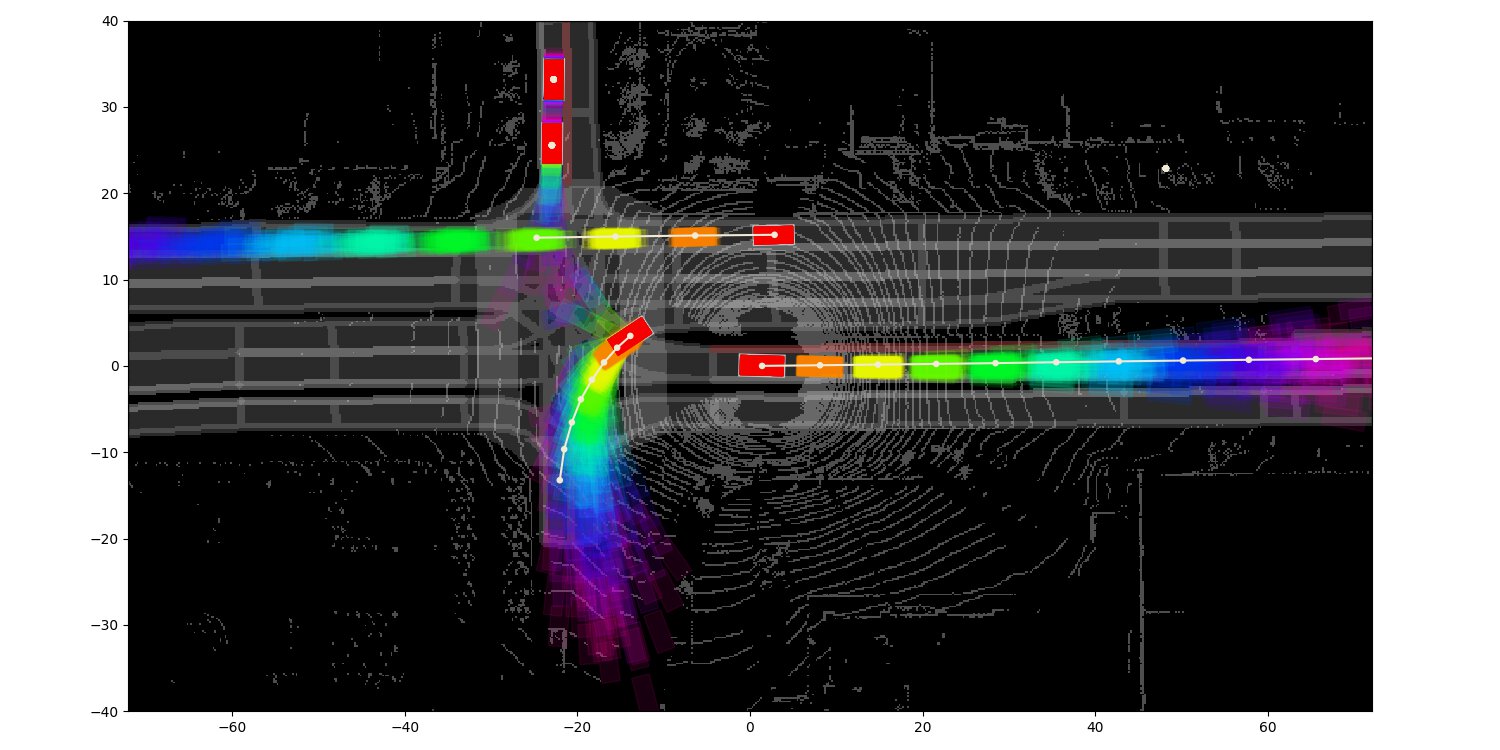}} &
        \raisebox{-0.5\height}{\includegraphics[width=0.325\linewidth, trim={3.5cm, 3cm, 3cm, 1cm}, clip]{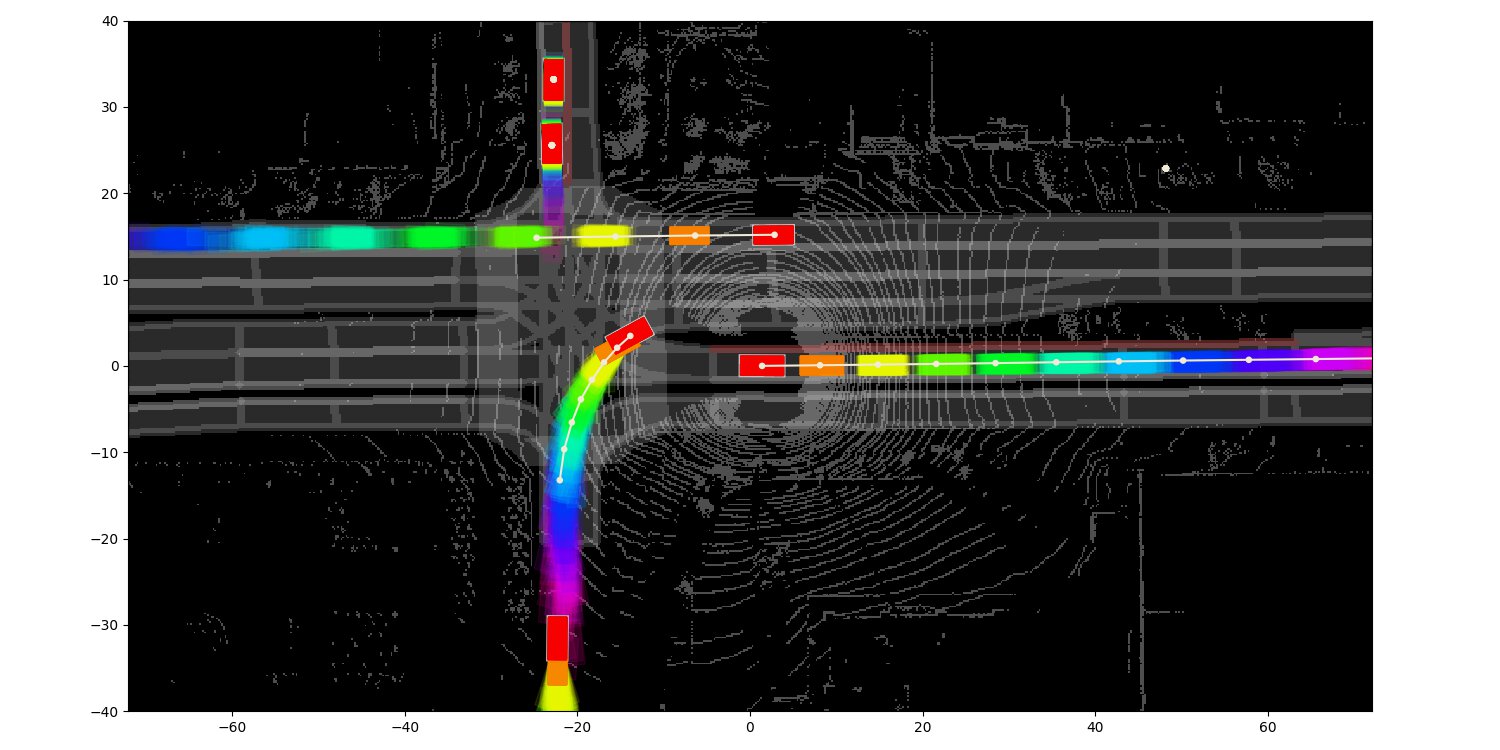}} &
        \raisebox{-0.5\height}{\includegraphics[width=0.325\linewidth, trim={3.5cm, 3cm, 3cm, 1cm}, clip]{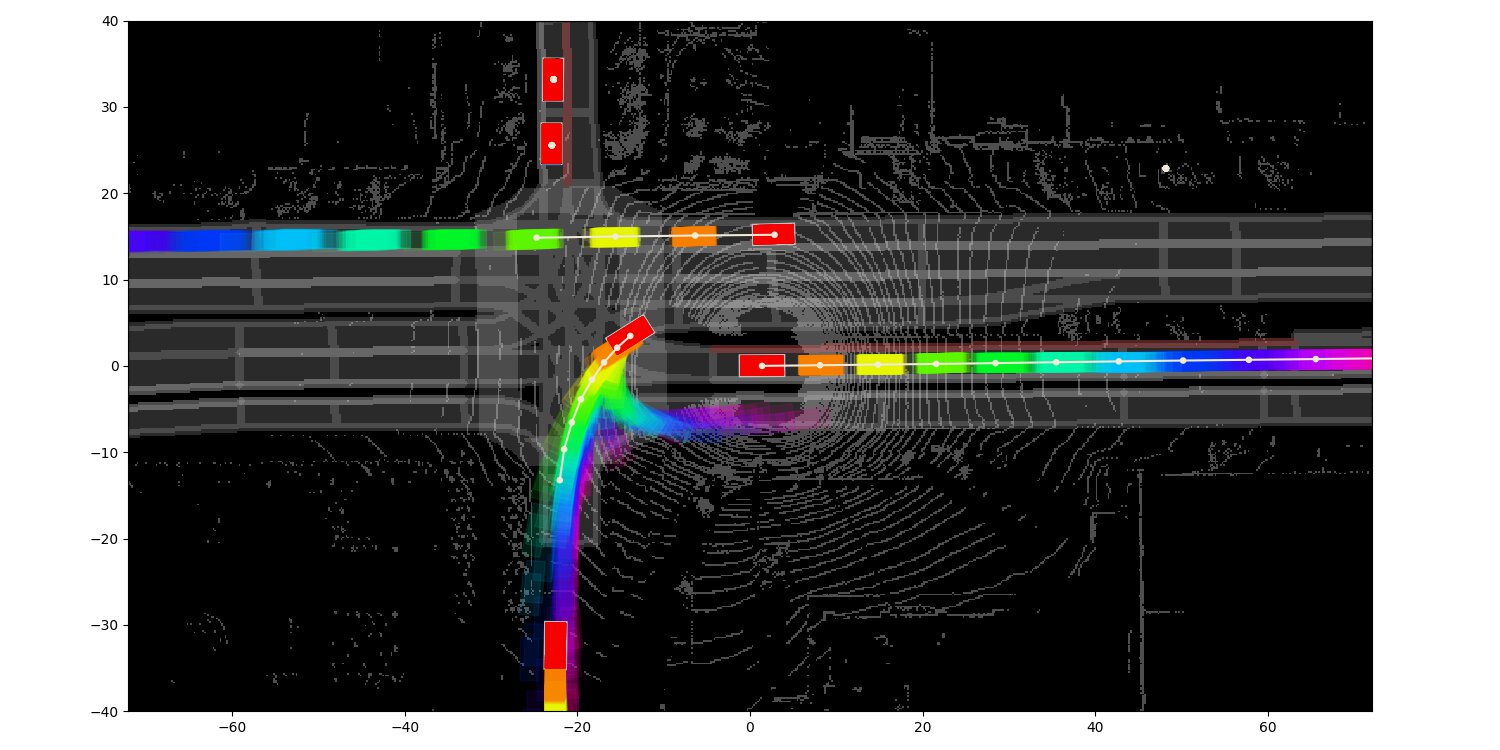}} \vspace{.5em} \\
        \rotatebox[origin=c]{90}{\textbf{Scenario 3}} &
        \raisebox{-0.5\height}{\includegraphics[width=0.325\linewidth, trim={3.5cm, 3cm, 3cm, 1cm}, clip]{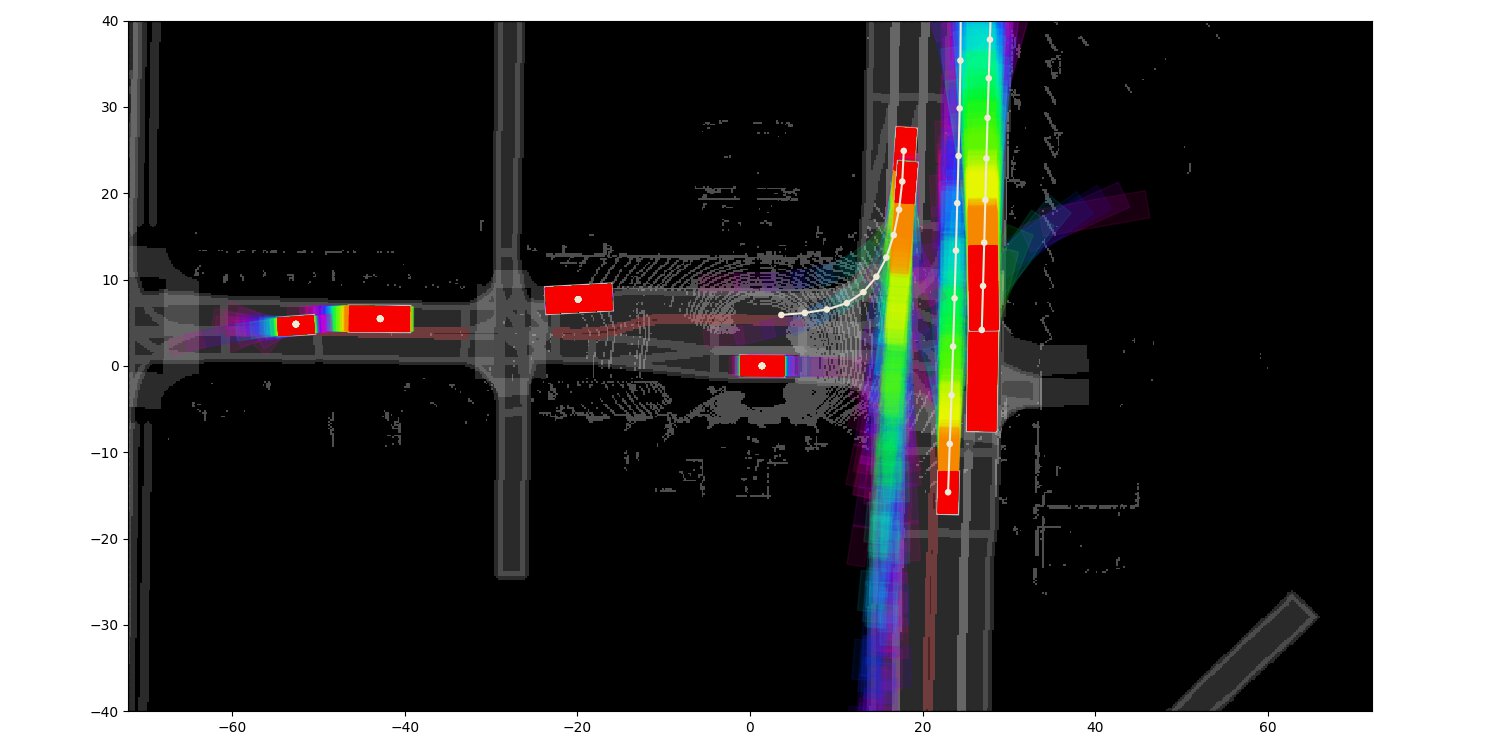}} &
        \raisebox{-0.5\height}{\includegraphics[width=0.325\linewidth, trim={3.5cm, 3cm, 3cm, 1cm}, clip]{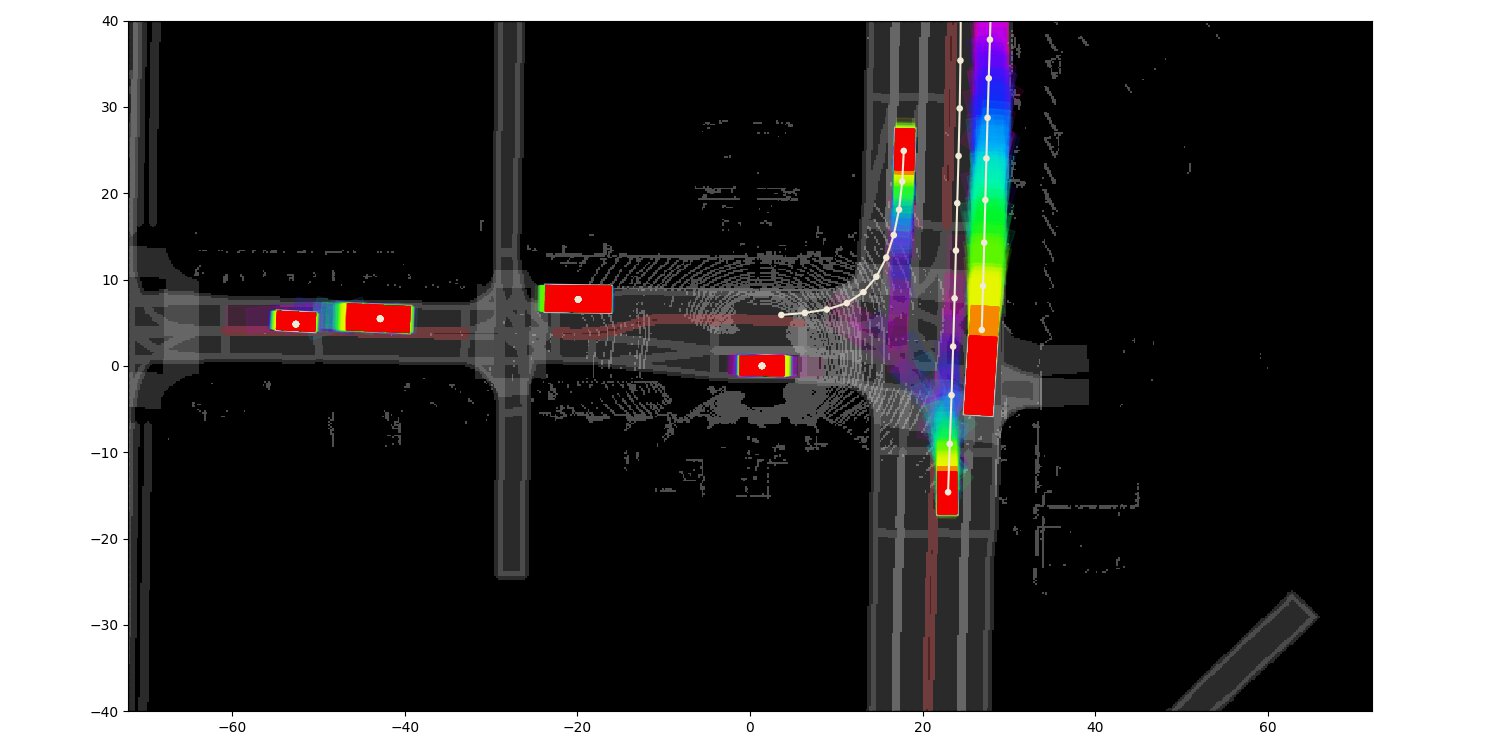}}  & 
        \raisebox{-0.5\height}{\includegraphics[width=0.325\linewidth, trim={3.5cm, 3cm, 3cm, 1cm}, clip]{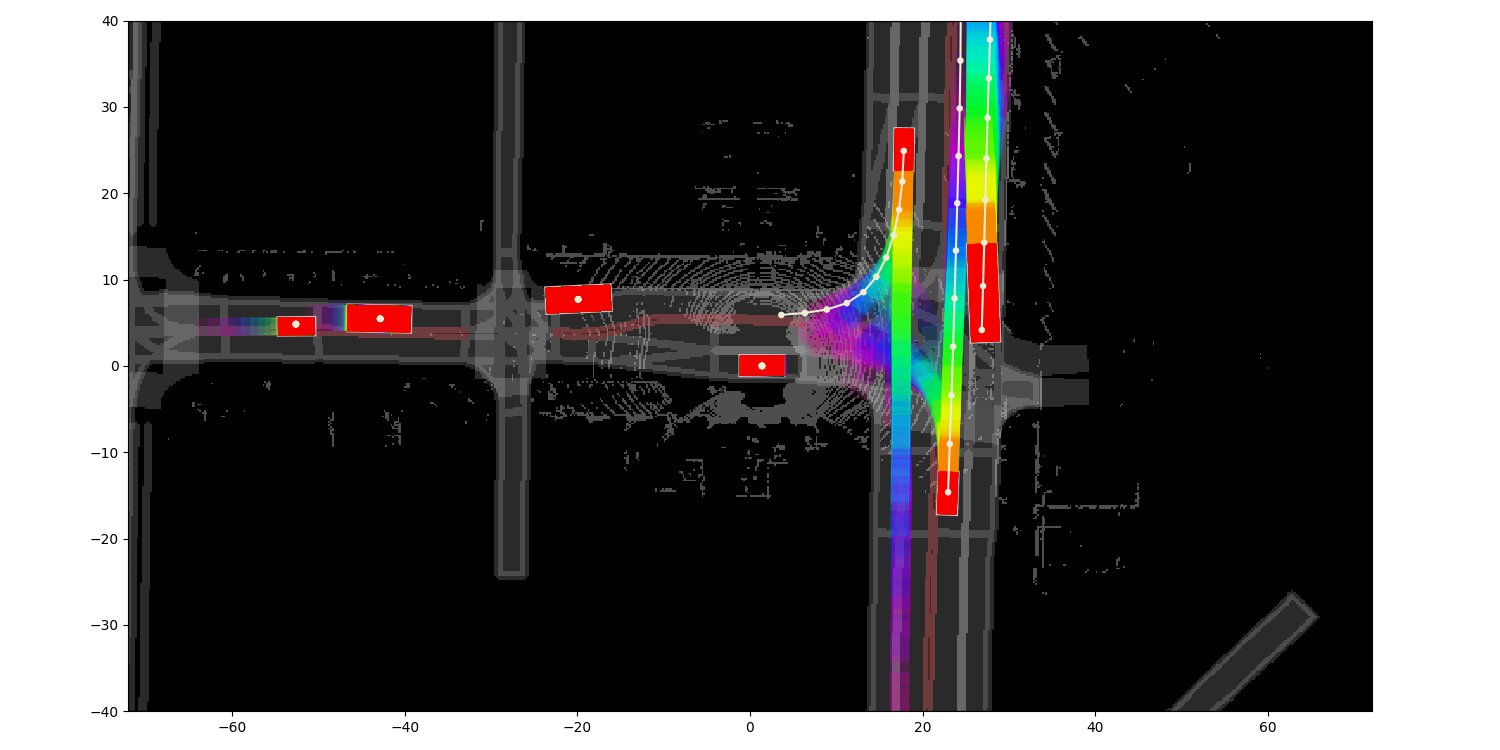}} \\
    \end{tabular}
    \caption{\textbf{Motion forecasting visualizations of 50 samples}. Time is encoded in the rainbow color map ranging from red (0s) to pink (5s).}
    \label{fig:qualitative_prediction}
\end{figure}

To validate the system-level impact of different perception and prediction models, we use the state-of-the-art learnable motion planner of \cite{plt} to plan a trajectory for the SDV ($\tau_{\text{SDV}}$):
{\small
$$
\tau_{\text{SDV}} = \argmin_{\tau \in \mathrm{T}} \mathbb{E}_{p(Y|X)} \left[ c(\tau, Y \setminus y_{\text{SDV}}) \right] \approx \argmin_{\tau \in \mathrm{T}} c(\tau, \begin{Bmatrix} Y^s \setminus y^s_{\text{SDV}} \colon \forall s \in 1 \dots S \end{Bmatrix})
$$
} 
where $p(Y|X)$ is the distribution over future trajectories output by the perception and prediction model, $\mathrm{T}$ is a predefined set of SDV trajectories given the map and high-level route, and $c$ is a costing function that measures safety and comfort taking into account the motion forecasts for the rest of the vehicles.
More concretely, the motion planner receives a Monte Carlo estimate of the future trajectory distribution with $S = 50$ sample trajectories (see Fig.\ref{fig:qualitative_prediction}) for every detected vehicle (excluding the SDV), which are considered obstacles in order to approximate the expected cost of plans $\tau \in \mathrm{T}$.

The experiments in Table~\ref{table:planning} measure how different motion forecasts translate into the safety and comfort of the SDV trajectory ($\tau_{\text{SDV}}$), an impact often overlooked by previous works. Our motion forecasts (ILVM) enable the motion planner to execute significantly safer and more comfortable trajectories. 
We notice that the ego-motion plans make similar progress across models, but our approach produces the closest trajectories to the expert demonstrations (lowest $\ell_2$ distance at 5 seconds into the future), while yielding much fewer collisions.
We include planning qualitative results in our supplementary material.

\subsection{Ablation Study}

\medskip\noindent\textbf{Implicit vs. Explicit Decoder:}
We ablate ILVM ($\mathcal{M}_0$ in Table~\ref{table:tor4d_ablation}) by replacing the proposed implicit decoder with an explicit decoder that produces a full covariance bi-variate Gaussian per waypoint, and the reconstruction loss with Negative Log Likelihood. This gives us $\mathcal{M}_1$, where ancestral sampling is used for inference: first sample latent, then sample output. Here, we can see that assuming conditional independence across actor at the output level significantly degrades all aspects of the motion forecasting performance. Most notably, the high scene collision rate shows that the samples are no longer socially consistent.

\begin{table*}[t]
	\centering
	\begin{threeparttable}
        \begin{tabularx}{\linewidth}{
                        s |
                        >{\centering\arraybackslash}l |   
                        >{\centering\arraybackslash}X |  
                        >{\centering\arraybackslash}X | 
                        >{\centering\arraybackslash}X | 
                        >{\centering\arraybackslash}X | 
                        >{\centering\arraybackslash}X  
                        }
            \toprule
                Type & 
                Model & 
                \multicolumn{1}{c|}{Collision} & 
                \multicolumn{1}{c|}{L2 human} &
                \multicolumn{1}{c|}{Lat. acc.} &
                \multicolumn{1}{c|} {Jerk} & 
                \multicolumn{1}{c}{Progress} \\
                {} & 
                {} &
                \multicolumn{1}{c|}{(\% up to 5s)} & 
                \multicolumn{1}{c|}{(m @ 5s)} & 
                \multicolumn{1}{c|}{(m/$s^2$)} &
                \multicolumn{1}{c|}{(m/$s^3$)} &
                \multicolumn{1}{c}{(m @ 5s)} \\
            \midrule
                \multirow{5}{1.5cm}{Indep. Output} 
                & \textsc{SpAGNN} \cite{casas2019spatially}           & 4.19  & 5.98  & 2.94 & 2.90 & 32.37  \\
                & \textsc{RulesRoad} \cite{Hong_2019_CVPR}            & 4.04  & 5.83  & 2.84 & 2.76 & 32.50 \\
                & \textsc{MTP} \cite{cui2018multimodal}               & 3.10  & 5.67  & 2.83 & 2.66 & 33.14 \\
                & \textsc{MultiPath} \cite{chai2019multipath}         & 3.30  & 5.58  & 2.73 & 2.57 & 32.99 \\
                & \textsc{R2P2-MA} \cite{2019arXiv190501296R}         & 3.71  & 5.65  & 2.84 & 2.53 & 33.90 \\
                \hline
                \multirow{4}{1.5cm}{Social Auto-regressive}
                & \textsc{SocialLSTM} \cite{alahi2016social}          & 4.22  & 5.92  & 2.76 & 2.66 & 32.60 \\
                & \textsc{NRI} \cite{kipf2018neural}                  & 4.94  &	5.73  &	2.78 & 2.55 & 33.43 \\
                & \textsc{ESP} \cite{2019arXiv190501296R}             & 3.13  & 5.48  & 2.76 & 2.44 & 33.74 \\
                & \textsc{MFP} \cite{tang2019multiple}                        & 4.14  & 5.57  & 2.61 & 2.43 & 32.94 \\
            \midrule
                & \ourmodelshort{}                                    & \textbf{2.64}  & \textbf{5.33}  & \textbf{2.59} & \textbf{2.30} & 33.72  \\
	        \bottomrule
		\end{tabularx}
    \end{threeparttable}
    \caption{\textbf{[\ourdataset{}] System Level Performance (ego-motion planning)}}
	\label{table:planning}
\end{table*}

\begin{table}[t]
	\centering
	\begin{threeparttable}
        \begin{tabularx}{\textwidth}{
                        s| s s s s|  %
                        s s s s s %
                        }
		    \toprule
                ID & Learned Prior & Implicit Output & SIM \mbox{Encoder} & SIM \mbox{Decoder} &  %
                $\mathrm{SCR}_{5s}$  & min SFDE  & min SADE  & mean SFDE  & mean SADE  \\ %
            \midrule
            $\mathcal{M}_0$ & \checkmark & \checkmark & \checkmark & \checkmark    & \textbf{0.70}  & \textbf{1.53} & \textbf{0.76} & \textbf{2.27} & \textbf{1.02}  \\
            \hline
            $\mathcal{M}_1$ & \checkmark &            & \checkmark & \checkmark    & 8.46  & 2.66  & 1.31  & 4.17 & 1.80 \\
            $\mathcal{M}_2$ &            & \checkmark & \checkmark &  \checkmark   & 1.10	& 1.53 & 0.76	& 2.43 & 1.08  \\
            $\mathcal{M}_3$ & \checkmark & \checkmark &            & \checkmark    & 1.03 & 1.57 & 0.78 & 2.42 & 1.08  \\
            $\mathcal{M}_4$ & \checkmark & \checkmark & \checkmark &               & 1.52 & 1.67 & 0.81 & 2.44 & 1.09  \\
            $\mathcal{M}_5$ & \checkmark & \checkmark &            &               & 1.74 & 1.66 & 0.81 & 2.43 & 1.08  \\
	        \bottomrule
		\end{tabularx}
    \end{threeparttable}
	\caption{\textbf{[\ourdataset{}] Motion Forecasting Ablation Study} ($S=15$ samples)}
	\label{table:tor4d_ablation}
\end{table}

\medskip\noindent\textbf{Learned vs. Fixed Prior:}
A comparison between $\mathcal{M}_0$ and $\mathcal{M}_2$ in Table~\ref{table:tor4d_ablation} shows that using a learned prior network $P(Z|X)$ achieves a better precision diversity trade-off  compared to using a fixed prior distribution of isotropic Gaussians.

\medskip\noindent\textbf{ILVM architecture:} In Table~\ref{table:tor4d_ablation}, $\mathcal{M}_3$ ablates the SIM encoder and prior networks by replacing them with MLPs that model $p(z_n|x_n)$ and $p(z_n|x_n, y_n)$ at the actor-level, respectively. $\mathcal{M}_4$ replaces the SIM decoder by an MLP per actor $y_n^s = \text{MLP}(X, z_n^s)$. Finally, $\mathcal{M}_5$ applies the changes in $\mathcal{M}_3$ and $\mathcal{M}_4$. These experiments show that both the graph based prior/encoder and decoder are important for our latent variable model. In particular, the large gap in scene level collision demonstrates that our proposed SIM encoder and decoder capture scene-level understanding that is not present in the ablations with independent assumptions at the latent or output level.


\section{Conclusion and Future Work}
We have proposed a latent variable model to obtain an implicit joint distribution over actor trajectories  that characterizes the 
dependencies 
over their future behaviors. Our model achieves fast parallel sampling of the joint trajectory space and produces scene-consistent motion forecasts. We have demonstrated the effectiveness of our method on two challenging datasets by significantly improving over  state-of-the-art motion forecasting models on scene-level sample quality metrics. Our method achieves much more precise predictions that are more socially consistent. We also show that our method produces significant improvements in motion planning, even though the planner does not make explicit use of the strong consistency of our scenes. We leave it to future work to design a motion planner to better utilize joint distributions over trajectories.

\clearpage
\bibliographystyle{splncs04}
\bibliography{eccv2020submission}

\clearpage
\appendix

\vspace*{0.2cm}
{ \noindent \Large \textbf{Supplementary Material}} \\

In the following supplementary materials, we provide: additional discussions of our method in the broader context of implicit generative models in Section~\ref{sec:implicit}, details about the datasets used in Section~\ref{sec:datasets}, more implementation details in Section~\ref{sec:implementation}, more in-depth evaluation details and quantitative results in Section~\ref{sec:results}, and finally more visualizations in Section~\ref{sec:visualizations}.

\section{Further Discussion on Implicit Generative Models}
\label{sec:implicit}
\subsection{Implicit Generative Models}
Typical probabilistic models for motion forecasting define an \emph{explicit} parameterized output distribution over each actor $n$ and trajectory waypoints across time $t$, $y^n_t$. Examples are the methods proposed in \cite{cui2018multimodal, casas2019spatially, chai2019multipath}, which parameterize their output distribution as a mixture of Gaussians, which can be sampled efficiently and provides a likelihood evaluation but assumes 1) independence across actors, and 2) a particular shape of the output distribution. In contrast, implicit generative models define a output distribution $p_{\theta}(Y)$ \emph{implicitly} by specifying a latent distribution $p(Z)$ from which we can sample,
followed by a mapping $f_{\theta}: \mathcal{Z} \rightarrow \mathcal{Y}$, which we refer to as the decoder. 

In particular, we can characterize the decoder in two ways: \begin{enumerate}
    \item via a specified and tractable conditional likelihood $p_{\theta}(Y|Z)$. In this case, many tools are available for inference and learning. Variational inference, and in particular the variational auto-encoder (VAE) \cite{kingma2013auto}, is a common choice.
    \item via a stochastic sampling procedure where $p(Y|Z)$ is not specified. In this case, likelihood-free inference methods are required for learning. Density estimation by comparison has been proposed \cite{mohamed2016learning} using either density ratio (GAN) or density difference (MMD).
    These methods, however,  are generally more difficult to optimize.
\end{enumerate}

In our model, we define $f_{\theta}$ as a deterministic function (parameterized by a graph neural network), since we wish the latent $Z$ to capture all the uncertainty in a scene and have $Y$ be deterministic given $Z$. To sidestep the difficulty of likelihood-free inference, particularly in a complex model where we optimize both perception and motion forecasting end-to-end, we make a mild assumption to leverage variational inference for learning. In the following sections, we provide a preliminary on variational inference before providing a detailed analysis of our model and learning approach.

\subsection{Variational Inference}
The variational auto-encoder (VAE) \cite{kingma2013autoencoding} specifies a directed graphical model with latent variables $z$ and output variables $y$. 
Conditional variational auto-encoder (CVAE) \cite{sohn2015learning} extends this formulation to the conditional generative setting, with additional input variables $x$. Now, for a given observation $x$, 
$z$ is drawn from the conditional prior distribution $p(z|x)$ and output $y$ is generated from the distribution $p(y|x,z)$.

The learning objective for the conditional generative model is maximizing the conditional log likelihood $\log p(y|x)$. 
But since marginalizing over continuous latent variables $z$ is intractable, it is typical to apply the Stochastic Gradient Variational Bayes (SGVB) framework and optimize the following evidence variational lower bound (ELBO) instead:
\begin{align*}
    \log p(y|x) \ge \mathbb{E}_{q_{\phi}(z|x, y)}[\log p_{\theta}(y|x,z)] - KL(q_{\phi}(z|x, y)||p_{\gamma}(z|x))
\end{align*}
Here, $q_{\phi}(z|x, y)$ is the learned approximate posterior, $p_{\gamma}(z|x)$ the learned approximate prior, and $p_{\theta}(y|x, z)$ the learned decoder.

Followup works have proposed further modifications to this objective to encourage disentanglement, prevent posterior collapse, and improve training stability. In this work, we follow \cite{higgins2017beta} in extending the ELBO objective with an additional hyperparameter $\beta$:
\begin{align*}
    L_{ELBO} = - \mathbb{E}_{q_{\phi}(z|x, y)}[\log p_{\theta}(y|x,z)] + \beta KL(q_{\phi}(z|x, y)||p_{\gamma}(z|x))
\end{align*}

\subsection{Analysis of Our Method}

We recall that to capture the joint distribution over all the actor trajectories we employ a deterministic decoder $Y = f(X, Z)$, letting the latent variable $Z$ capture all the stochasticity. 
Thus, instead of optimizing the likelihood based reconstruction objective that appears in the ELBO, we opted for a Huber loss on the trajectory waypoints. 
This choice can be interpreted as an assumption of $p_{\theta}(Y|X,Z)$ being a Gaussian/Laplacian with fixed diagonal covariance. 
For simplicity, let's assume our Huber loss is always active within the L2 segment, but the following derivation could be easily done with a Laplacian as well.
In this view, we can further interpret $\beta$ as the variance of the underlying Gaussian, as follows:

\begin{align*}
    \mathcal{L}_{forecast} &= || Y - Y_{GT}||_2^2  + \beta KL(q_{\phi}(Z|X,Y)||p_{\gamma}(Z|X)) \\
    & \propto \frac 1 {\beta} || Y - Y_{GT}||_2^2  + KL(q_{\phi}(Z|X,Y)||p_{\gamma}(Z|X)) \\
    & \propto \mathbb{E}_{q_{\phi}(Z|X,Y)}\left[\log \mathcal{N}(Y_{GT} |f_{\theta}(X, Z), \frac \beta 2 I) \right] + KL(q_{\phi}(Z|X,Y)||p_{\gamma}(Z|X))
\end{align*}

To see this, recall the log likelihood of Gaussian with diagonal covariance:
\begin{align*}
    \log \mathcal{N}(Y_{GT} | \mu, \sigma^2I) & = - \frac 1 {2\sigma^2} \sum_{n=1}^N\sum_{t=1}^T(y_{n, GT}^t - \mu^t)^2 - c_{\sigma} \\
    \log \mathcal{N}(Y_{GT} | \mu, \sigma^2I=\frac \beta 2 I) & = - \frac 1 {\beta} \sum_{n=1}^N\sum_{t=1}^T(y_{n, GT}^t - \mu^t)^2 - c_{\sigma}
\end{align*}
where $c_{\sigma}$ can be neglected since it is constant relative to $\mu$, and thus does not contribute to its gradient.

Empirically we found $\beta=0.05$ to yield the best results. Under the interpretation above, this would translate into using a fixed variance of $2.5cm$ while learning our model. We note that this is extremely small in the context of motion forecasts (where vehicles can easily travel 50 meters in 5 seconds), and thus consistent with our goal of approximating $Y$ as a deterministic mapping from $X$ and $Z$, letting $Z$ capture (nearly) all the uncertainty at a scene level.

\section{Datasets} \label{sec:datasets}
We benchmark our approach on two datasets: \ourdataset{} \cite{yang2018pixor} and \nuscenes{} \cite{caesar2019nuscenes}. This allow us to test the effectiveness of our approach in two vehicle platforms with different LiDAR sensors and maps, driving in multiple cities across the world.

\medskip\noindent\textbf{\ourdataset{}}
Our dataset contains more than one million frames collected over several cities in North America with a 64-beam, roof-mounted LiDAR. 
Our labels are very precise 3D bounding box tracks with a maximum distance from the self-driving vehicle of 100 meters. There are 6500 snippets in total, each 25 seconds long. In each city, we have access to high definition maps capturing the geometry and the topology of each road network.
Following previous works in joint perception and motion forecasting \cite{luo2018fast,casas2018intentnet,casas2019spatially} we consider a rectangular region centered around the self-driving vehicle that spans 144 meters along the direction of its heading and 80 meters across. In these experiments, the model is given one second of LiDAR history and has to predict 5 seconds into the future. 

\medskip\noindent\textbf{\nuscenes{}}
This dataset consists of 1,000 snippets of 20 seconds each, collected in Boston and Singapore (right-side vs. left-side driving). Their 32-beam LiDAR captures a sparser point cloud than the one in \ourdataset{}, making object detection more challenging. 
High definition maps are also provided.
We use the evaluation setup proposed in their perception benchmark, where the previous 10 LiDAR sweeps (0.5 seconds) are fed to the model, and the region of interest is a circle of 50 meters radius around the SDV. The prediction horizon is 5 seconds.
\section{Implementation Details}
\label{sec:implementation}
In this section, first we provide implementation details about each component in our joint perception and motion forecasting model. We then discuss the required adaptations for the baselines.

\subsection{ILVM Details}

\subsubsection{LiDAR Pointcloud Parameterization:}
Following \cite{yang2018pixor}, we use a voxelized representation of the 3D LiDAR point cloud in Bird's Eye View (BEV) as the main input to our model. As in its follow-up work \cite{yang2018hdnet}, we normalize the height dimension with dense ground-height information provided by HD maps for \ourdataset{} dataset only (\nuscenes{} does not provide this information). 
To exploit motion cues, we leverage multiple LiDAR sweeps by compensating the ego-motion (i.e. projecting the past sweeps to the coordinate frame of the current sweep), as proposed by \cite{luo2018fast}.
Following \cite{casas2018intentnet}, we ravel the height and time dimension into the channel dimension, to use 2D convolution to process spatial-temporal information efficiently.
The final representation is a 3D occupancy tensor of dimensions $(\frac{ L }{\Delta L}, \frac{ W} {\Delta W}, \frac {H \cdot T} {\Delta H \cdot \Delta T})$. Here, $L=144$, $W=80$, and $H=5$ are the spatial dimensions in meters. $\Delta L = \Delta W = \Delta H = 0.2$ m/pixel are the resolutions for the spatial dimensions, $T=5$ seconds is the prediction horizon, and $\Delta T = 0.5$ seconds/time-step is the time resolution.
\vspace{-0.15cm}

\vspace{-0.15cm}
\subsubsection{High-Definition Maps Parameterization:}
We use a rasterized map representation encoding traffic elements such as intersections, lanes, roads, and traffic lights.
Elements with different semantics are encoded into different channels in the raster, as proposed by \cite{casas2018intentnet}. 

The map elements we rasterize are the following: drivable surface polygons, road polygons, intersection polygons, vehicle lane polygons going straight, dedicated left and right vehicle lane polygons, dedicated bike lane polygons, dedicated bus lane polygons, centerline markers for all lanes, lane dividers for all lanes with semantics (allowed to cross, not allowed to cross, might be allowed to cross). This gives us a total of 13 different map channels combining these elements.
\vspace{-0.15cm}

\vspace{-0.15cm}
\subsubsection{Shared Perception Backbone:}
We use a lightweight backbone network adapted from \cite{yang2018pixor} for feature extraction.
In particular, we instantiate two separate streams such that the voxelized LiDAR and rasterized map are processed separately first.
The resulting features from both streams are then concatenated feature-wise since they share the same spatial resolution, and finally fused by a convolutional header. Our LiDAR backbone uses 2, 2, 3, and 6 layers in its 4 residual blocks. The convolutions in the residual blocks of our LiDAR backbone have 32, 64, 128 and 256 filters with a stride of 1, 2, 2, 2 respectively. The backbone that processes the high-definition maps uses 2, 2, 3, and 3 layers in its 4 residual blocks. 
The convolutions in the residual blocks of our map backbone have 16, 32, 64 and 128 filters with a stride of 1, 2, 2, 2 respectively. For both backbones, the final feature map is a multi-resolution concatenation of the outputs of each residual block, as explained in \cite{engelcke2017vote3deep}. This gives us 4x down-sampled features with respect to the input. 
The header network consists of 4 convolution layers with 256 filters per layer.
We use GroupNorm \cite{wu2018group} because of our small batch size (number of frames) per GPU. These extracted features inform both the downstream detection and motion forecasting networks, explained next.
\vspace{-0.15cm}

\vspace{-0.15cm}
\subsubsection{Object Detection:}
We use two convolutional layers to output a classification (i.e. confidence) score and a bounding box for each anchor location following the output parameterization proposed in \cite{yang2018pixor}, which are finally reduced to the final set of candidates by applying non-maximal suppression (NMS) with an IoU of 0.1, and finally thresholding low probability detections (given by the desired common recall).
\vspace{-0.15cm}

\vspace{-0.15cm}
\subsubsection{Actor Feature Extraction:}
To arrive at the final actor level features $x_n$, we apply rotated ROI Align \cite{huang2018improving} to extract fixed size spatial feature maps for bounding boxes with arbitrary shapes and rotations from our global feature map extracted by the backbone. We pool a region around each actor in its frame with an axis defined by the actor's centroid orientation. The region in BEV space spans for 10m backwards, 70m in front, and 40m to both sides of the actor.
After applying the rotated ROI Align operator, we get a feature map for each actor of size 40 x 40 x 256. We then apply a 4-layer down-sampling convolutional network followed by max-pooling along the spatial dimensions to reduce the feature map to a 512-dimensional feature vector per actor. The convolutional network uses a dilation factor of 2 for the convolutional layers to enlarge the receptive field for the per-actor features, which we found to be important. We use ReLU as the non-linearity and GroupNorm for normalization.
\vspace{-0.15cm}

\vspace{-0.15cm}
\subsubsection{Scene Interaction Module:}
Our scene interaction module is inspired by \cite{casas2019spatially}, and is used in our Prior, Encoder, and Decoder networks. 
Our edge or message function consists of a 3-layer MLP that takes as input the hidden states of the 2 terminal nodes at each edge in the graph at the previous propagation step as well as the projected coordinates of their corresponding bounding boxes.
We use feature-wise max-pooling as our aggregate function in order to be more robust to changes in the graph topology between training and inference, since at training we use the ground-truth bounding boxes but at inference employ the detected bounding boxes.
To update the hidden states we use a GRU cell.
Finally, to output the results from the graph propagations, we use a 2-layer MLP.

\subsubsection{Motion Forecasting:}
The inference of our motion forecasting model is explained step-by-step in Algorithm~\ref{alg:ilvm}. 
Our $\text{Prior}_{\gamma}$ and $\text{Encoder}_{\phi}$ modules are both composed of 2 SIMs with different parameters, one that predicts the latent means $Z_{\mu}$ and one that predicts the latent sigmas $Z_{\sigma}$. 
We use the same input, hidden, and output dimension of 64 for these SIMs. 
To obtain the 64-dimensional input $H^0$ to the ${Prior}_{\gamma}$ SIMs, we use a 2-layer MLP to summarize the 512-dimensional actor feature $X$.
For $\text{Encoder}_{\phi}$, we use an additional 2-layer MLP to embed the ground truth future trajectories $Y_{GT}$ into a 64-dimensional embedding first, then summarize the concatenated 576-dimensional vector into the 64-dimensional input $H^0$ to the SIMs.
The $\text{Decoder}_\theta$ is implemented with a single SIM, which takes a 576-dimensional input $H^0$ (i.e. direct concatenation of 512-dimensional actor features $X$ and 64-dimensional latent sample $Z^s$), and outputs a 20-dimensional vector $Y^s$ (i.e. 10 waypoints with $(x, y)$ coordinates) for each actor.
Although we have described the algorithm as sequential over scenes $1 \dots S$ for clarity in the algorithm, the sampling and decoding of all scenes can be done in parallel.

\begin{algorithm*}[h!]
    \caption{Motion Forecasting} \label{alg:ilvm}
    \textbf{Input:} 
    Actor features $X=\begin{Bmatrix}x_1, x_2, \cdots, x_N\end{Bmatrix}$. BEV locations of object detections $C = \begin{Bmatrix} c_0,c_1, ...,c_N \end{Bmatrix}$
    Number of scene samples to generate $S$.
    
    \textbf{Output:} Scene trajectory samples in bird's-eye-view space $\begin{Bmatrix}Y^1, Y^2, \cdots, Y^S\end{Bmatrix}$, where $Y^s = \begin{Bmatrix} y_1^{s}, y_2^{s}, ..., y_N^{s} \end{Bmatrix}$ ($N$ is the number of detected actors).
    
    \begin{algorithmic}[1]
    \State $\begin{Bmatrix}Z_\mu, Z_\sigma \end{Bmatrix} \gets \text{Prior}_{\gamma}(X, C)$ \Comment{Use SIM modules to output latent distribution}
    \For{$s=1, ..., S$} \Comment{Run for all requested number of samples}
        \State $Z^s  \sim \mathcal{N}\left(\begin{Bmatrix}Z_\mu, Z_\sigma \cdot I \end{Bmatrix}\right)$ \Comment{Sample a scene latent from diagonal gaussian}
        \State $H^s = \begin{Bmatrix}\text{MLP}(x_n \oplus z_n^s): \forall n \in 1 \dots N \end{Bmatrix}$
        \State $Y^s = \text{Decoder}_\theta (H^s, C)$ \Comment{Use SIM module to decode trajectory sample}
    \EndFor
    \State\Return $\begin{Bmatrix}Y^1, Y^2, \cdots, Y^S\end{Bmatrix}$
    \end{algorithmic}
\end{algorithm*}

\subsubsection{Optimization Details:}

We use the Adam optimizer \cite{kingma2014adam} with an initial learning rate of 1.25e-5 and no weight decay. To weigh the multi-task objective, we use $[\alpha, \lambda, \beta] = [0.1, 0.5, 0.05]$. We follow \cite{cyclic_annealing} in using a cyclic annealing schedule for $\beta$. More specifically, we perform warmup for 40k steps in 10k step cycles.

\subsection{Baseline Details}
Here we provide the implementation details behind how we updated our baseline models to meet our perception and prediction setting.
There are basically two options for comparison: 
\begin{enumerate}[label=(\alph*)]
    \item use an off-the-shelf detector and tracker to provide past trajectories, or
    \item replace their past trajectory encoders by our backbone and per actor feature extraction. 
\end{enumerate}
SpAGNN \cite{casas2019spatially} showed that (a) using an off-the-shelf tracker (Unscented Kalman Filter + Hungarian matching) results in much worse performance than option (b), so we stick to the latter for a fair comparison where all methods use the same architecture to extract actor features $X$ from sensor data, which is trained end-to-end with the motion forecasting module for each baseline.

\subsubsection{Explicit Marginal Likelihood Models:}
SpAGNN \cite{casas2019spatially} was originally proposed in the joint perception and prediction setting and therefore does not require any adaptation. We adapt MTP \cite{cui2018multimodal} and MultiPath \cite{chai2019multipath} to use our backbone network, object detection and per actor feature extraction and then apply their proposed mixture of trajectories output parameterization, where each way-point is a gaussian. 

A detail worth noting is that these baselines do not propose a way to get temporally consistent samples, since the gaussians are independent across time (the models are not auto-regressive). Thus, we introduce a heuristic sampler to get temporally consistent samples from this model. The sampled trajectories are extracted using the re-parameterization trick for a bi-variate normal:
$$y^s_{n, t} = \mu_{n, t}  + A_{n, t} \cdot \varepsilon^s_n$$
where the model predicts a normal distribution $\mathcal{N}\left(y^s_{n, t} | \mu_{n, t}, \Sigma_{n, t} \right)$ per waypoint $t$, $(A_{n, t})^T \cdot A_{n, t} = \Sigma_{n, t}$ is the cholesky decomposition of the covariance matrix, and $\varepsilon^s_n \sim \mathcal{N}(0, I)$ is the noise sampled from a standard bi-variate normal distribution. Note that the noise $\varepsilon^s_n$ is constant across time $t$ for a given sample $s$ and actor $n$. Intuitively, having a constant noise across time steps allows us to sample waypoints whose relative location with respect to its predicted mean and covariance is constant across time (i.e. translated by the predicted mean and scaled by the predicted covariance per time).

\subsubsection{Autoregressive Models}

While many papers that utilize auto-regressive models \cite{2019arXiv190501296R, rhinehart2018r2p2, tang2019multiple} use fully observed states as dynamic inputs, we extend these models to the joint detection and motion forecasting task. For all auto-regressive models, we use the \emph{detection backbone} and the \emph{actor feature extraction} modules that we use for all models, including ours.

Due to the compounding error problem \cite{ross2011reduction} found in auto-regressive models, we had to make some adjustments to the training procedure to account for the noise in the $t - 1$ conditioning space. Typically during training for auto-regressive models, a one-step prediction distribution given the previous ground-truth value $p\left(y_{t+1} | x, y_{t, GT}\right)$ is learned. This can cause a catastrophic mismatch between the input distribution that the model sees during training and that it sees during inference. To help simulate the noise it sees during inference, we add gaussian noise to the conditioning state $\tilde{y} = y_{GT} + \epsilon$ where $\epsilon \sim \mathcal{N}\left(0, I \cdot \alpha \right)$. The parameter $\alpha$ defines to the amount of noise we expect in meters between time-steps (we use a value of 0.2m in our experiments.)
We note that we also tried scheduled sampling \cite{bengio2015scheduled}, but adding white noise worked better.

For ESP \cite{2019arXiv190501296R}, we extended our \emph{R2P2-MA} implementation with the "whisker" indexing technique to get added context into the feature map at the location of the conditioning state $\tilde{y}_t$. Due to memory constraints, we had to limit the radii of the whiskers to $[1m, 2m, 4m]$ while keeping the seven angle bins. We also reproduced the social context conditions but with a minor modification. While the original paper specified a fixed number of actors, we used k-nearest neighbors to select a set amount of $M=4$ neighbors to gather social features and model the distribution $p\left(y_{n, t+1} | x, y_{n, t}, y_{1,t}, y_{2,t}, \cdots, y_{M,t} \right)$. Lastly, We note that the originally proposed SocialLSTM and NRI do not leverage any sensor or map data, but since we share the feature extraction architecture for all models, their adaptations do have access to these cues making their methods more powerful than originally proposed.

\section{Additional Evaluation Details and Results}
\label{sec:results}
In this section, we present additional evaluation details and quantitative results on \emph{detection} and \emph{motion forecasting}.

\subsection{Detection} 

\begin{figure}[t]
    \centering
    \includegraphics[width=\textwidth]{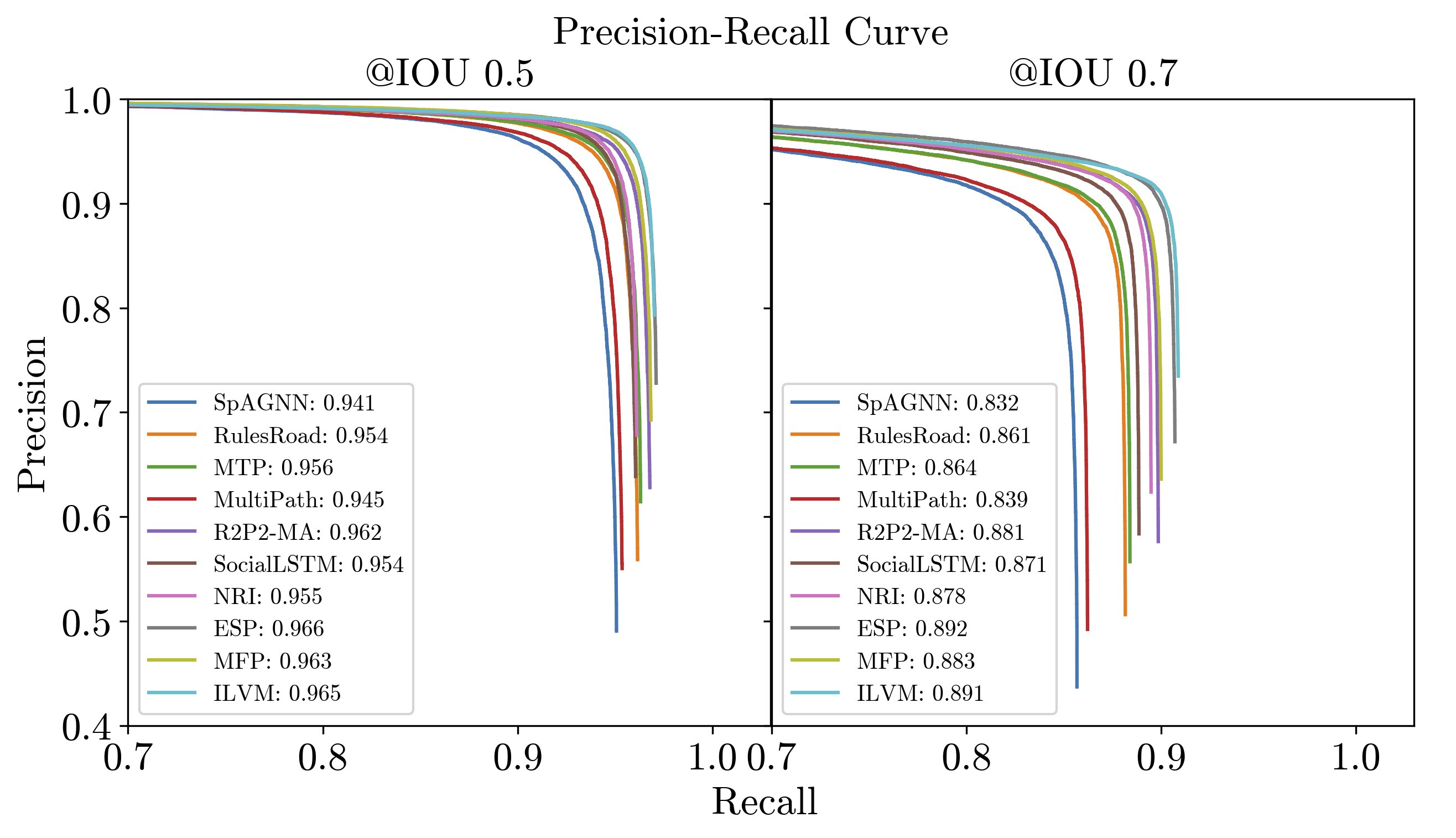}
    \caption{Precision-Recall curve at IoU 0.5 and 0.7. Legend shows mAP (mean Average Precision) for each model. Note: horizontal axis starts at 0.7 recall.}
    \label{fig:detection}
\end{figure}

Fig.~\ref{fig:detection} shows that our model achieves the best detection performance at both IoU thresholds.
Since all models have the same backbone and detection header, we conjecture that our learning objective eases the joint optimization of both detection and motion forecasting.

\label{sec:detection}

\subsection{Motion Forecasting}
For \ourdataset{} experiments, we operate the object detector at a 90\% common recall point.
In \nuscenes{} experiments, we operate the object detector at an 80\% common recall point, since detection is more challenging in this dataset due to the sparser 32-beam LiDAR sensor (as opposed to 64-beam in \ourdataset{}).

\subsubsection{Diversity vs. Precision in multimodal prediction}
\begin{figure}[h]
    \centering
    \includegraphics[width=\textwidth]{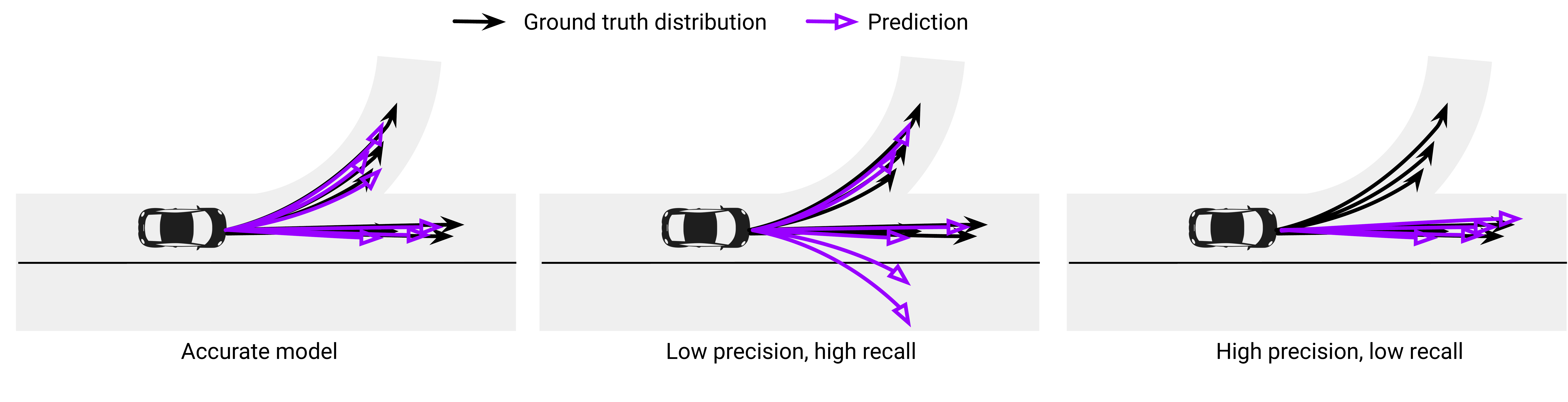}
    \caption{Diversity vs. Precision for an individual actor}
    \label{fig:diversity_precision}
\end{figure}

Fig.~\ref{fig:diversity_precision} showcases three different predictions that exhibit different qualities, which we use to illustrate the language used throughout the paper. On the left we show an accurate model that can nicely capture the bimodal distribution due to the branching map topology. On the middle, a prediction model predicts high diversity samples (high recall), but has low precision as it predicts unrealistic samples that are out of distribution. On the right, we show a high-precision prediction, meaning that all the samples are within the true data distribution, but low recall or diversity, meaning that it misses some modes of the ground-truth distribution.

\subsubsection{Actor-level vs. Scene-level metrics} 
\begin{figure}[h]
    \centering
    \includegraphics[width=\textwidth]{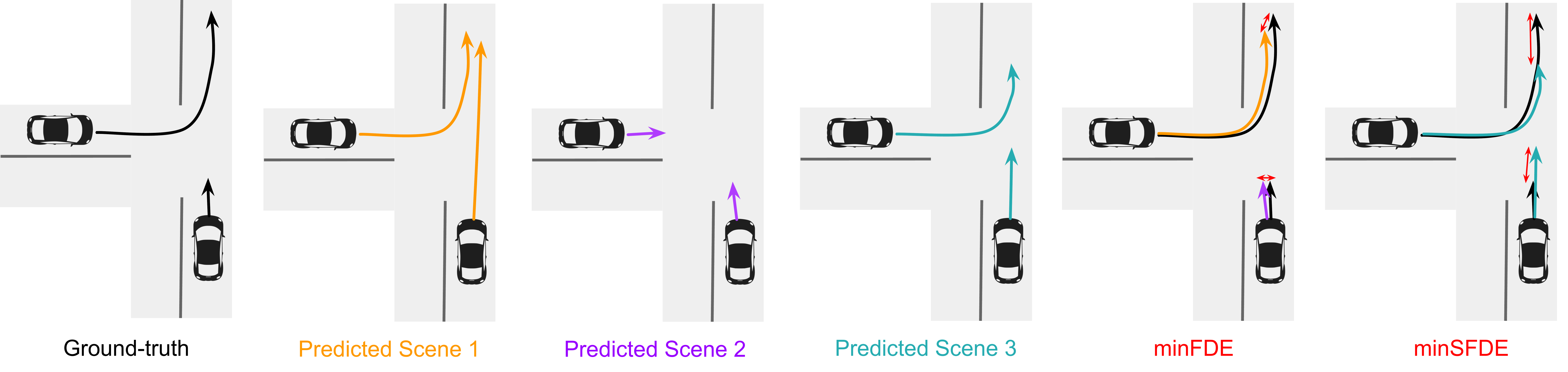}
    \caption{Coverage metrics: actor-level (minFDE) vs. scene-level (minSFDE). The first column shows the groun-truth trajectories, the next 3 are possible futures predicted by the model, the last 2 show the trajectory samples selected by each metric (which are different), as well as their error (red arrows).}
    \label{fig:minFDE_minSFDE}
\end{figure}

Fig.~\ref{fig:minFDE_minSFDE} motivates the need for scene-level metrics to evaluate the characterization of the joint distribution over actors. In particular, minFDE (actor-level minimum displacement error) will take the minimum error trajectory for each actor regardless of which scene prediction it belongs to. In contrast, minSFDE (our proposed scene-level counterpart) takes the trajectories from the predicted scene with less average error across vehicles, thus selecting the scene that is most consistent with the ground-truth as a whole.

\subsubsection{Scene Consistency -- Sample Collisions} 
\begin{figure}[h!]
    \centering
    \includegraphics[width=.75\textwidth]{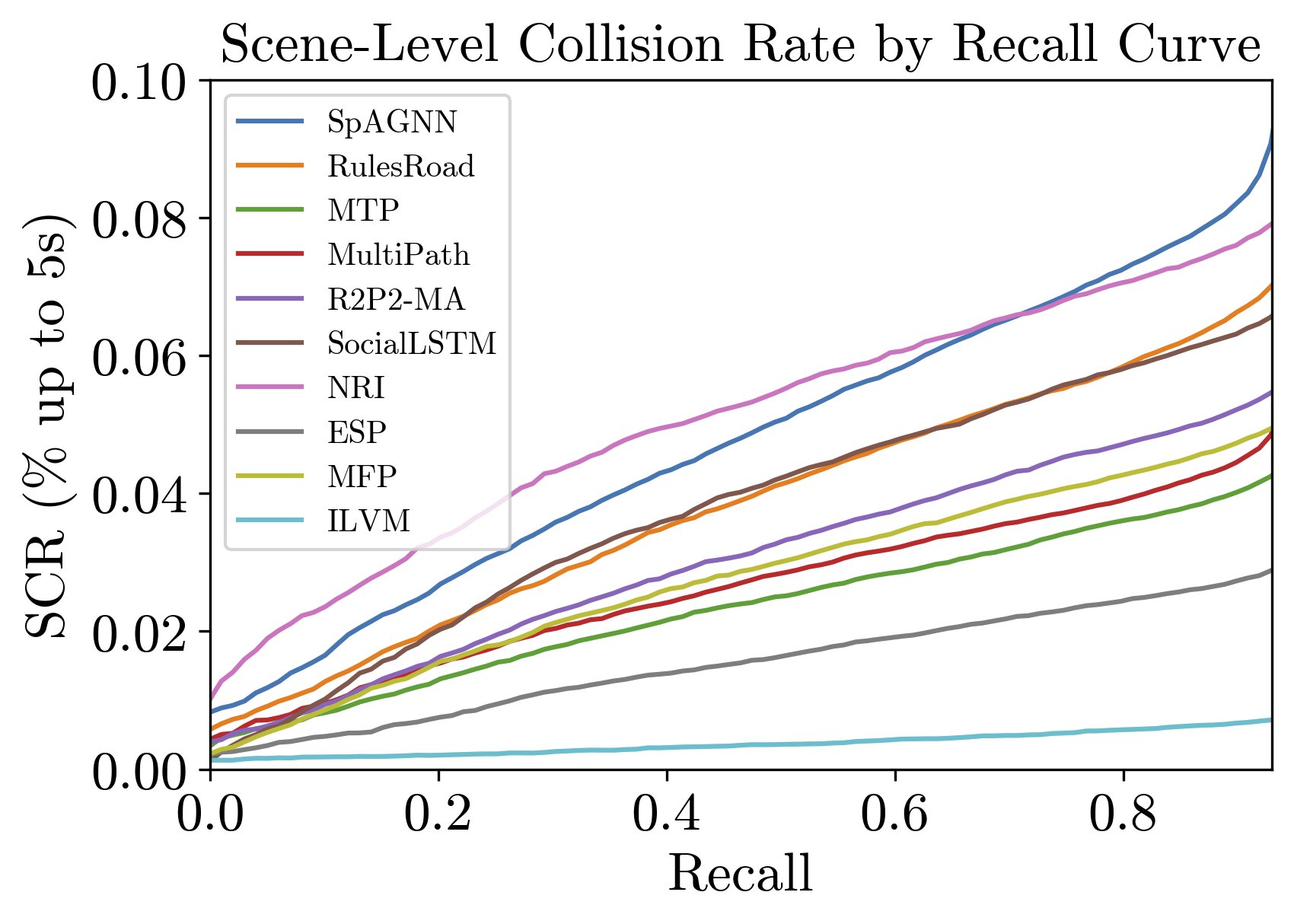}
    \caption{\textsc{ILVM} models social interaction consistently well at different recall levels.}
    \label{fig:collision_by_recall}
\end{figure}

Here, we demonstrate that our models produce scene-level samples that are more socially consistent regardless of which recall point we operate our object detector. 
Figure~\ref{fig:collision_by_recall} shows our Scene Collision Rate (SCR) at different detection recall points (also known as operating point).
As the recall point is chosen to be higher and higher there are more low probability actors in the scene which greatly increases the chances of a predicted collision, as expected.
When analyzing the results, it is clear that just sharing social features as \cite{casas2019spatially} does is not enough to create scene consistent samples.
Models that do joint sampling such as \ourmodel{} and ESP \cite{2019arXiv190501296R} do markedly better on this measure. 
Interestingly, \ourmodel{} barely sees an increase in the amount of collisions as recall increases, which shows that our model is able to generate scene consistent samples no matter the complexity of the scene.

\subsubsection{Sample Quality -- Cumulative Hit Rate:} 
So far in the motion forecasting literature, actor-level precision (meanFDE) and recall (minFDE) metrics of the trajectory samples have been proposed, but no attempt has been made to combine them in a single metric, despite the fact that the previous metrics have evident drawbacks.
For instance, meanFDE disregards the fact that multiple plausible futures could be very far apart and overly penalizes multi-modality, while a low minFDE can be achieved by just predicting fanned out distributions that cover big spaces.
Here, we propose to use a cumulative Hit Rate curve, where the horizontal axis corresponds to the L2 error threshold, and the vertical axis to the percentage of samples that fall under such error.
Moreover, we extend this notion to also capture failures in the object detector, by considering that false positive and false negative detections always have error higher than the threshold, thus obtaining a holistic metric for joint perception and prediction. 
We now define this metric mathematically. We use $\hat{y}$ to denote the ground truth future $y_{GT}$.
 
\begin{align*}
    \text{Hit Rate}\left(y, \hat{y}, t, \epsilon \right) = \frac{1}{N S}\sum_{n=1}^N\sum_{s=1}^S \text{Hit}\left( y_{n}^{t, s}, \hat{y}_{n}^t, y_{n}^0, \hat{y}_{n}^0, \epsilon \right ) \\
\end{align*}
\begin{align*}
\text{Hit}\left( y_{n}^{t, s}, \hat{y}_{n}^t, y_{n}^0, \hat{y}_{n}^0, \epsilon \right ) = \left\{\begin{matrix}
    1 && \text{if } \text{IoU}\left(y^0_n, \hat{y}^0_n \right ) > 0.5 \text{ and } ||y_{n}^{t, s} - \hat{y}_{n}^t||_2 < \epsilon \\ 
    0 && \text{otherwise}
    \end{matrix}\right.
\end{align*}

\begin{figure}[h!]
    \centering
    \includegraphics[width=\textwidth]{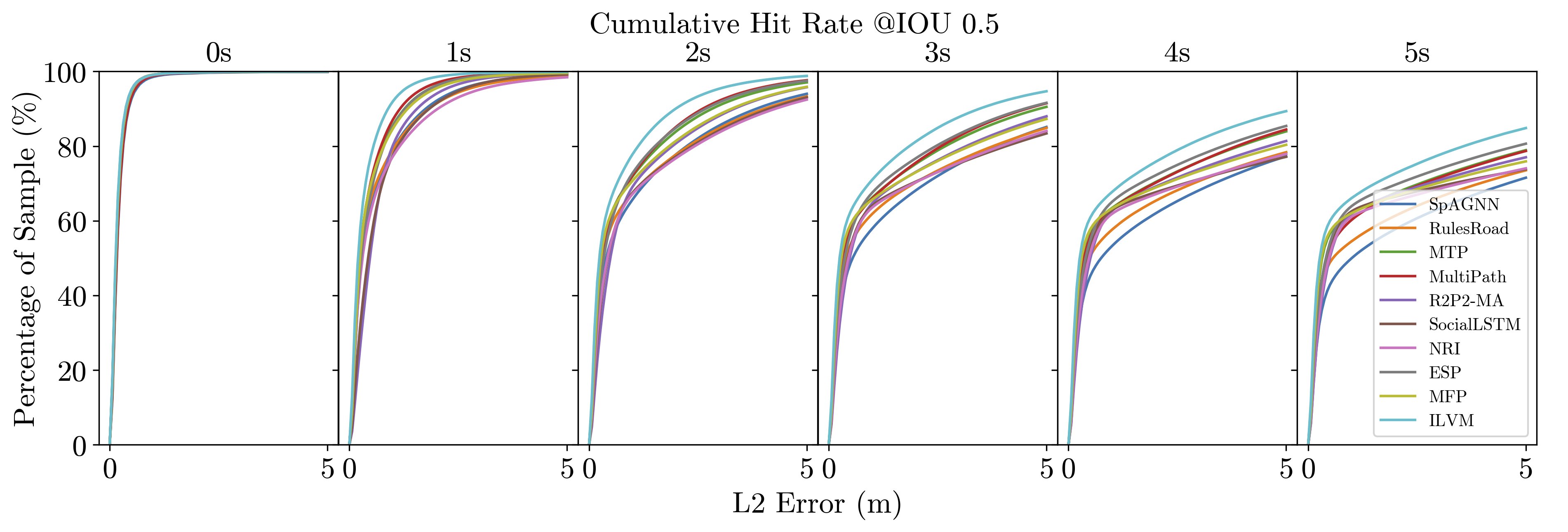}
    \caption{\textsc{ILVM} obtains the best hit rate at all time-steps in the prediction horizon.}
    \label{fig:sample_l2_cdf}
\end{figure}

Thus $\text{Hit Rate}$ finds the percentage of samples that are true positive detections and have an L2 error below a threshold $\epsilon$. We sweep $\epsilon$ values of 0.0m to 5.0m to get the broader curve which gives us the distribution on how likely each model is to get a detection and sample close to the ground-truth. We do not compute the metric above 5 meter error since we consider that to always be a bad sample or "miss". Fig.~\ref{fig:sample_l2_cdf} shows that Our ILVM significantly outperforms all baselines in cumulative hit rate across across all time steps.

\subsubsection{Sample Quality -- Breakdown} 

We define Along-Track and Cross-Track distance as the longitudinal and lateral distance after projecting motion forecasts into the ground-truth actor trajectory coordinate. This breakdown is important, since lateral error is semantically more significant than longitudinal error for the downstream task of ego-motion planning.

\begin{figure}[h!]
    \centering
    \includegraphics[width=\textwidth]{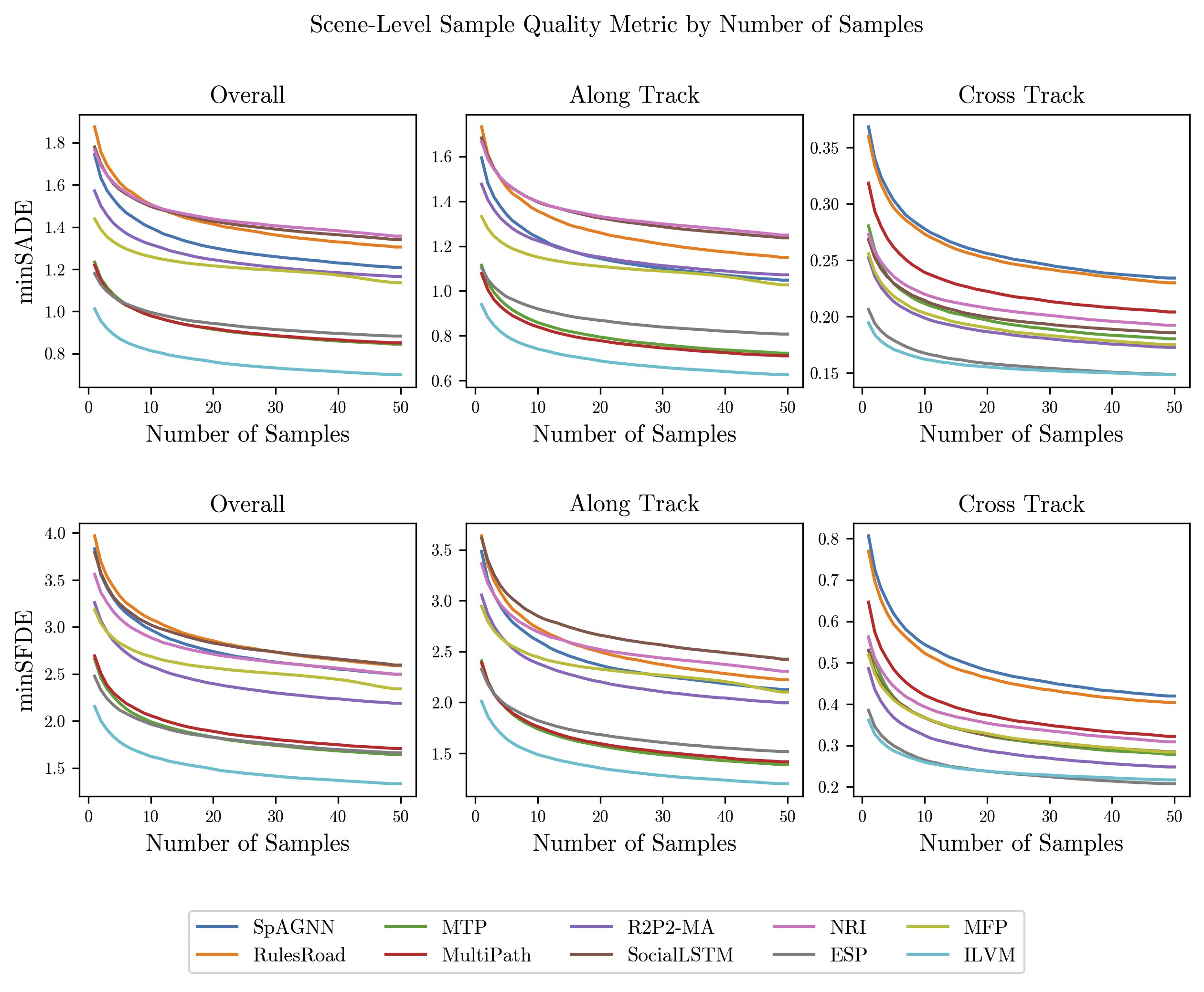}
    \caption{Our \textsc{ILVM} outperforms more significantly in along track error than cross track error for both scene-level minADE and minFDE metrics.}
    \label{fig:breakdown_by_samples}
\end{figure}

Fig.~\ref{fig:breakdown_by_samples} provides an in-depth analysis of the sample quality of our motion forecasts by examining the error breakdown between Along-Track and Cross-Track.
Furthermore, we highlight the robustness of our model in recovering ground-truth scenes when given different number of samples.

The results showcase that model rankings may not be consistent in the breakdown of Along-Track vs. Cross-Track. In particular, we find the main contributor to ILVM's advantage is better Along-Track forecast, while having equal or better Cross-Track. 
This implies that our model is able to better estimate overall current and future velocity of the actors while having the same precision on their path as \textsc{ESP}, and significantly better than the other baselines.

\subsubsection{Sample Quality -- Precision Diversity Tradeoff}
\begin{figure}[h!]
    \centering
    \includegraphics[width=.5\textwidth]{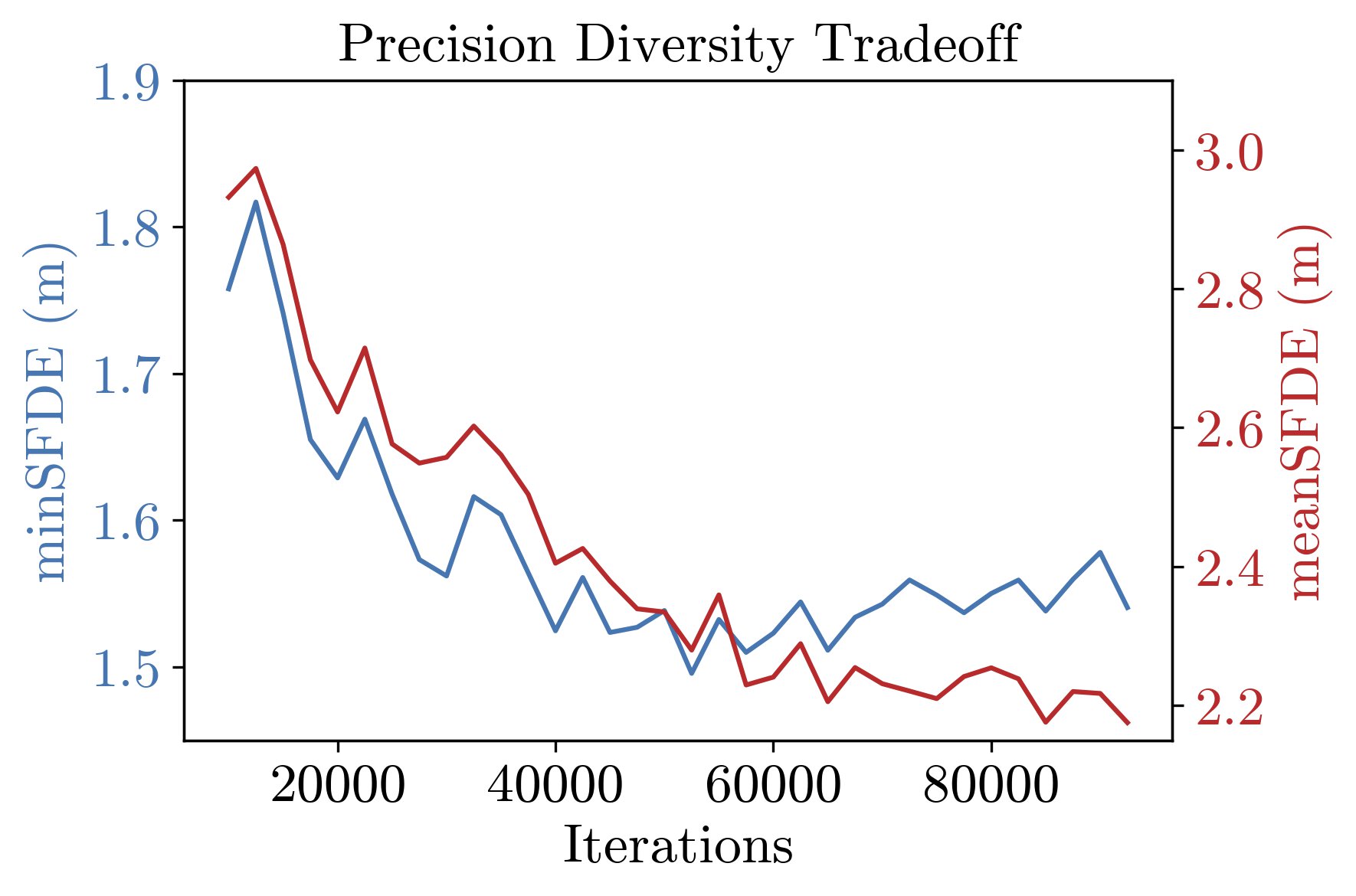}
    \caption{Tradeoff between minSFDE and meanSFDE as model training progresses. Note that the \textcolor{blue}{\emph{minSFDE curve}} follows the y-axis on the left, and the \textcolor{red}{\emph{meanSFDE curve}} follows the y-axis on the right.}
    \label{fig:tradeoff_training}
\end{figure}

In Fig.~\ref{fig:tradeoff_training}, we showcase the progression of scene-level sample quality metrics of Our ILVM during training. While the precision metric (meanSFDE) continues to improve past 50k iterations, the diversity metric (minSFDE) reaches its optimum. This sheds light on the inherent tradeoff between the diversity and precision aspect of sample quality measure, particularly when we only have access to a single ground truth realization of the multiple plausible futures. 

\subsubsection{Ablation -- KL Term}
We include an analysis of how the $beta$ weight on the KL term trades off diversity and precision in Fig.~\ref{table:tor4d_ablation_beta}.

\begin{table}[h]
	\centering
	\begin{threeparttable}
        \begin{tabularx}{\textwidth}{
                        s| %
                        s s s s s %
                        }
		    \toprule
                $\beta$  & %
                $\mathrm{SCR}_{5s}$ (\%) & min SFDE (m) & min SADE (m) & mean SFDE (m) & mean SADE (m) \\ %
            \midrule
            0.01	  & 1.41	& 1.70	& 0.84 & 2.58  & 1.17 \\
            0.03	  & 0.89	& 1.59	& 0.79 & 2.37  & 1.06 \\
            0.05	  & 0.70	& \textbf{1.53}	& \textbf{0.76} & 2.27  & 1.02 \\
            0.5	    & \textbf{0.64}	& 1.85	& 0.85 & \textbf{1.90}  & \textbf{0.86} \\
            1	      & 0.64	& 1.87	& 0.87 & 1.95  & 0.88 \\
	        \bottomrule
		\end{tabularx}
    \end{threeparttable}
	\caption{\textbf{[\ourdataset{}] KL loss ablation study}}
	\label{table:tor4d_ablation_beta}
\end{table}

We observe that:
\begin{enumerate}
    \item \emph{High beta}: model loses multimodality and predicts a single future without variance. Low recall (minSADE) and high precision (meanSADE, collision). High KL loss constrains the posterior to be close to the prior, thus limiting the flexibility to encode useful information.
    \item \emph{Low beta}: model produces very high entropy distributions that try to cover all possible futures at the expense of producing unrealistic samples. High recall (minSADE) and low precision (meanSADE, collision). Low KL loss allows the posterior to diverge from the prior, which creates a gap between training and inference. Then at inference, the decoder struggles to interpret latent samples from the prior distribution, which it’s not trained on. 
\end{enumerate}

\section{Additional Visualizations}
\label{sec:visualizations}
\subsubsection{Scene Consistency:}
\begin{figure}[t]
    
    \centering
    \begin{tabular} {@{}c@{\hspace{.1em}}c@{\hspace{.5em}}c}
        {} & \textbf{Sample 1} & \textbf{Sample 2} \\
        \rotatebox[origin=c]{90}{\textbf{SpAGNN}} &
        \raisebox{-0.5\height}{\includegraphics[width=0.35\linewidth, trim={1cm, 1cm, 1cm, .33cm}, clip]{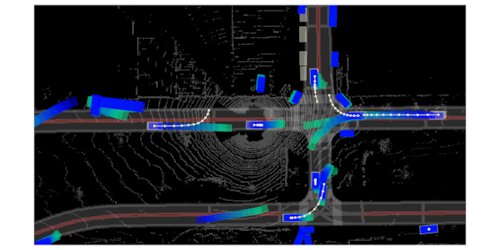}} &
        \raisebox{-0.5\height}{\includegraphics[width=0.35\linewidth, trim={1cm, 1cm, 1cm, .33cm}, clip]{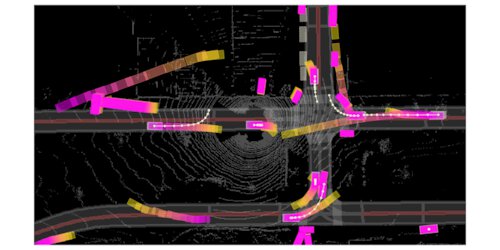}} \\[7.5ex]
        \rotatebox[origin=c]{90}{\textbf{MTP}} &
        \raisebox{-0.5\height}{\includegraphics[width=0.35\linewidth, trim={1cm, 1cm, 1cm, .33cm}, clip]{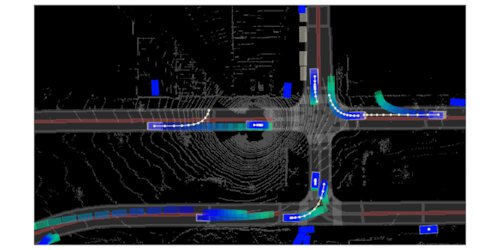}} &
        \raisebox{-0.5\height}{\includegraphics[width=0.35\linewidth, trim={1cm, 1cm, 1cm, .33cm}, clip]{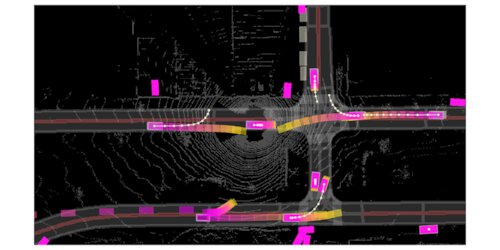}} \\[7.5ex]
        \rotatebox[origin=c]{90}{\textbf{MultiPath}} &
        \raisebox{-0.5\height}{\includegraphics[width=0.35\linewidth, trim={1cm, 1cm, 1cm, .33cm}, clip]{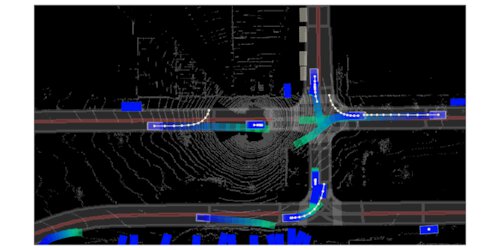}} &
        \raisebox{-0.5\height}{\includegraphics[width=0.35\linewidth, trim={1cm, 1cm, 1cm, .33cm}, clip]{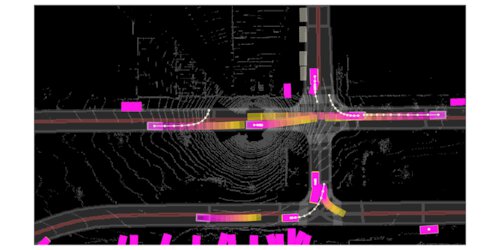}} \\[7.5ex]
        \rotatebox[origin=c]{90}{\textbf{R2P2}} &
        \raisebox{-0.5\height}{\includegraphics[width=0.35\linewidth, trim={1cm, 1cm, 1cm, .33cm}, clip]{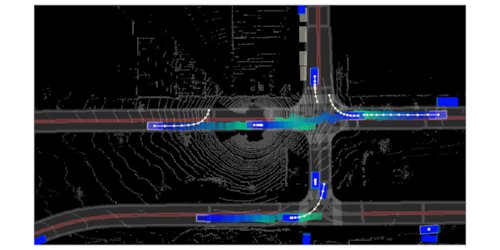}} &
        \raisebox{-0.5\height}{\includegraphics[width=0.35\linewidth, trim={1cm, 1cm, 1cm, .33cm}, clip]{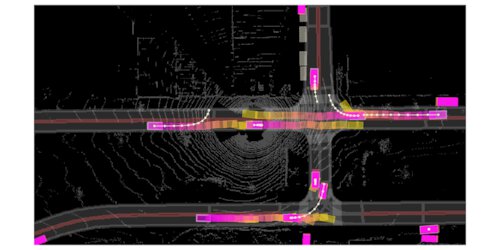}} \\[7.5ex]
        \rotatebox[origin=c]{90}{\textbf{ESP}} &
        \raisebox{-0.5\height}{\includegraphics[width=0.35\linewidth, trim={1cm, 1cm, 1cm, .33cm}, clip]{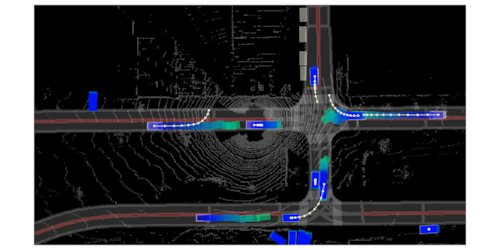}} &
        \raisebox{-0.5\height}{\includegraphics[width=0.35\linewidth, trim={1cm, 1cm, 1cm, .33cm}, clip]{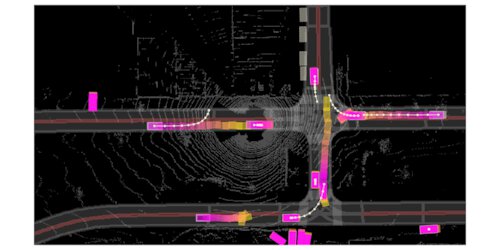}} \\[7.5ex]
        \rotatebox[origin=c]{90}{\textbf{MFP}} &
        \raisebox{-0.5\height}{\includegraphics[width=0.35\linewidth, trim={1cm, 1cm, 1cm, .33cm}, clip]{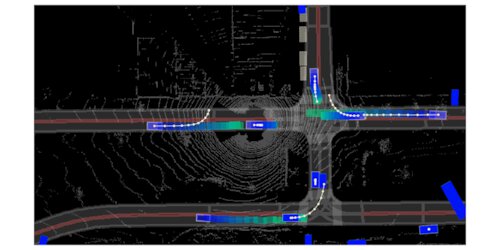}} &
        \raisebox{-0.5\height}{\includegraphics[width=0.35\linewidth, trim={1cm, 1cm, 1cm, .33cm}, clip]{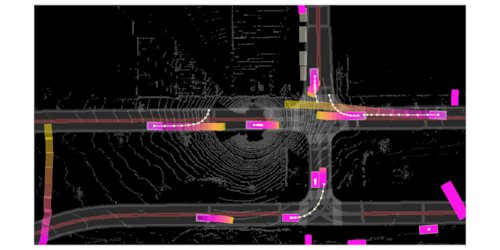}} \\[7.5ex]
        \rotatebox[origin=c]{90}{\textbf{RoR}} &
        \raisebox{-0.5\height}{\includegraphics[width=0.35\linewidth, trim={1cm, 1cm, 1cm, .33cm}, clip]{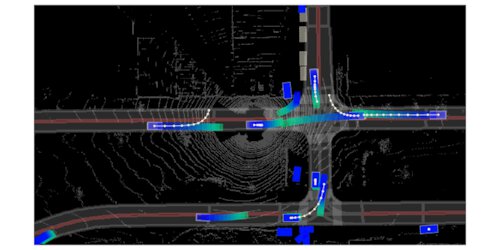}} &
        \raisebox{-0.5\height}{\includegraphics[width=0.35\linewidth, trim={1cm, 1cm, 1cm, .33cm}, clip]{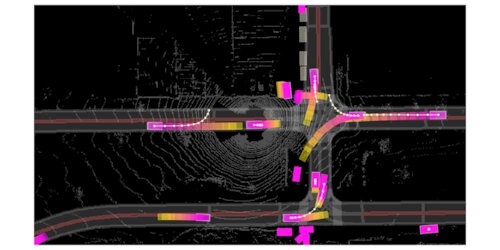}} \\[7.5ex]
        \rotatebox[origin=c]{90}{\textbf{ILVM}} &
        \raisebox{-0.5\height}{\includegraphics[width=0.35\linewidth, trim={1cm, 1cm, 1cm, .33cm}, clip]{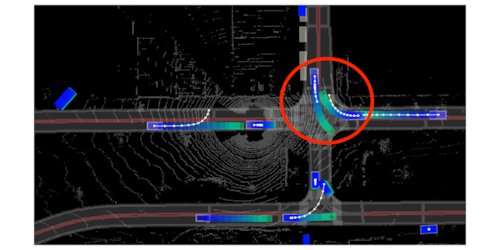}} &
        \raisebox{-0.5\height}{\includegraphics[width=0.35\linewidth, trim={1cm, 1cm, 1cm, .33cm}, clip]{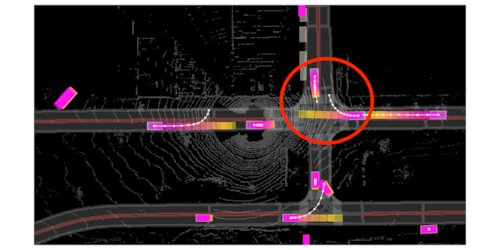}}
    \end{tabular}
    \caption{\textbf{Scene-level samples}. Our latent variable model captures complex interactions at intersections. In this example, the car facing south will yield/go if the car facing west goes straight/turns right, respectively. The baselines do not capture this complex interaction, and most show inconsistent (colliding) samples for the 2 highlighted actors.
    }
    \label{fig:supp_scene_samples_1}
\end{figure}

\begin{figure}[t]
    
    \centering
    \begin{tabular} {@{}c@{\hspace{.1em}}c@{\hspace{.5em}}c}
        {} & \textbf{Sample 1} & \textbf{Sample 2} \\
        \rotatebox[origin=c]{90}{\textbf{SpAGNN}} &
        \raisebox{-0.5\height}{\includegraphics[width=0.35\linewidth, trim={1cm, 1cm, 1cm, .33cm}, clip]{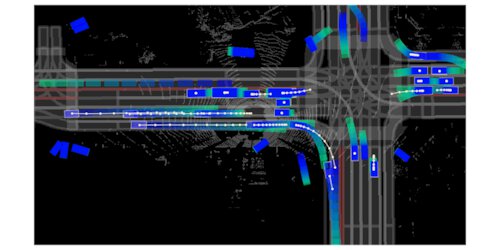}} &
        \raisebox{-0.5\height}{\includegraphics[width=0.35\linewidth, trim={1cm, 1cm, 1cm, .33cm}, clip]{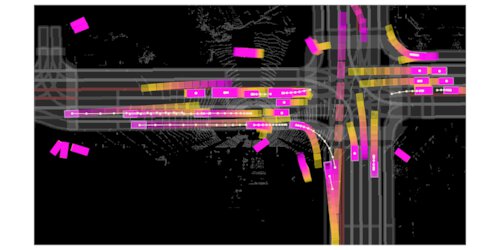}} \\[7.5ex]
        \rotatebox[origin=c]{90}{\textbf{MTP}} &
        \raisebox{-0.5\height}{\includegraphics[width=0.35\linewidth, trim={1cm, 1cm, 1cm, .33cm}, clip]{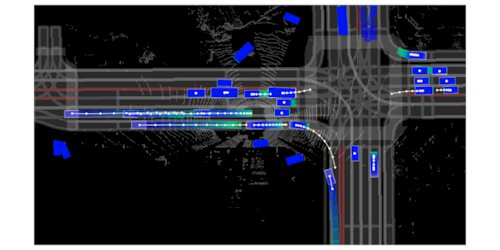}} &
        \raisebox{-0.5\height}{\includegraphics[width=0.35\linewidth, trim={1cm, 1cm, 1cm, .33cm}, clip]{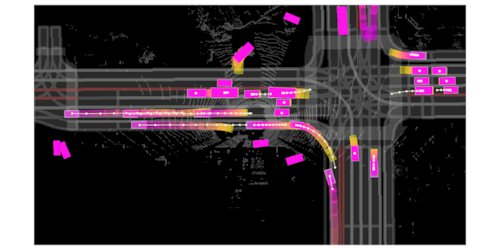}} \\[7.5ex]
        \rotatebox[origin=c]{90}{\textbf{MultiPath}} &
        \raisebox{-0.5\height}{\includegraphics[width=0.35\linewidth, trim={1cm, 1cm, 1cm, .33cm}, clip]{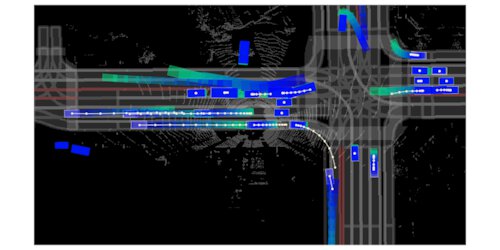}} &
        \raisebox{-0.5\height}{\includegraphics[width=0.35\linewidth, trim={1cm, 1cm, 1cm, .33cm}, clip]{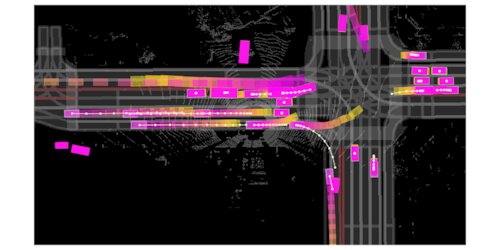}} \\[7.5ex]
        \rotatebox[origin=c]{90}{\textbf{R2P2}} &
        \raisebox{-0.5\height}{\includegraphics[width=0.35\linewidth, trim={1cm, 1cm, 1cm, .33cm}, clip]{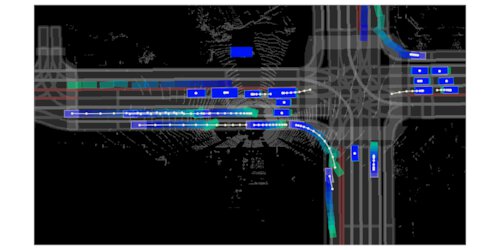}} &
        \raisebox{-0.5\height}{\includegraphics[width=0.35\linewidth, trim={1cm, 1cm, 1cm, .33cm}, clip]{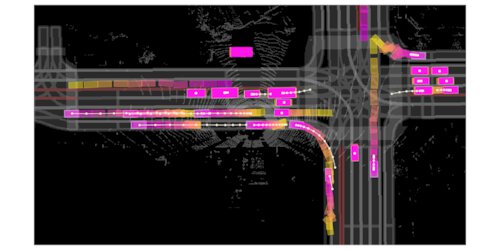}} \\[7.5ex]
        \rotatebox[origin=c]{90}{\textbf{ESP}} &
        \raisebox{-0.5\height}{\includegraphics[width=0.35\linewidth, trim={1cm, 1cm, 1cm, .33cm}, clip]{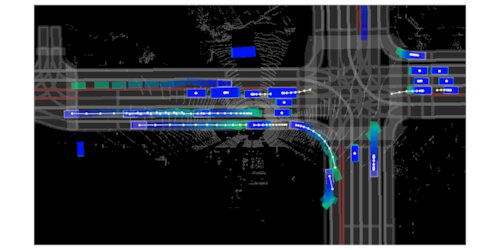}} &
        \raisebox{-0.5\height}{\includegraphics[width=0.35\linewidth, trim={1cm, 1cm, 1cm, .33cm}, clip]{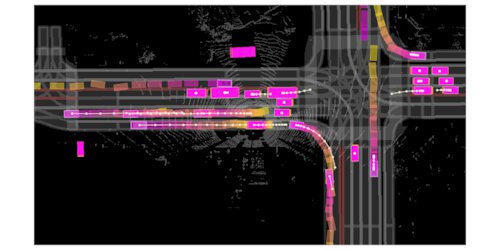}} \\[7.5ex]
        \rotatebox[origin=c]{90}{\textbf{MFP}} &
        \raisebox{-0.5\height}{\includegraphics[width=0.35\linewidth, trim={1cm, 1cm, 1cm, .33cm}, clip]{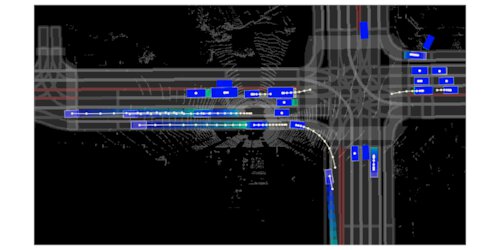}} &
        \raisebox{-0.5\height}{\includegraphics[width=0.35\linewidth, trim={1cm, 1cm, 1cm, .33cm}, clip]{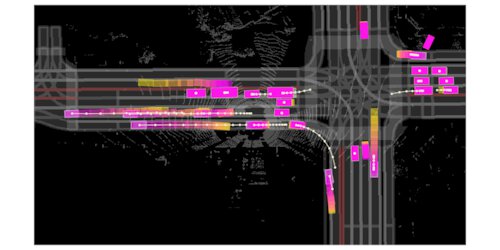}} \\[7.5ex]
        \rotatebox[origin=c]{90}{\textbf{RoR}} &
        \raisebox{-0.5\height}{\includegraphics[width=0.35\linewidth, trim={1cm, 1cm, 1cm, .33cm}, clip]{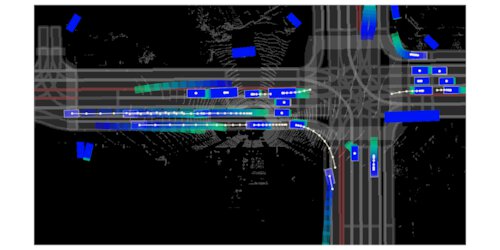}} &
        \raisebox{-0.5\height}{\includegraphics[width=0.35\linewidth, trim={1cm, 1cm, 1cm, .33cm}, clip]{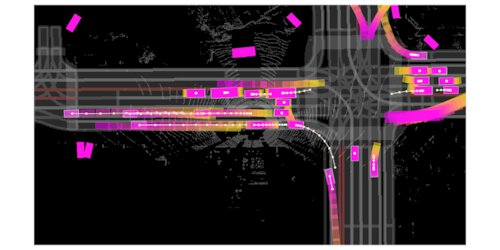}} \\[7.5ex]
        \rotatebox[origin=c]{90}{\textbf{ILVM}} &
        \raisebox{-0.5\height}{\includegraphics[width=0.35\linewidth, trim={1cm, 1cm, 1cm, .33cm}, clip]{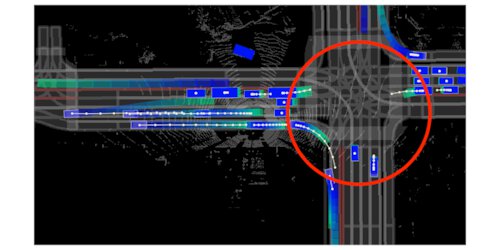}} &
        \raisebox{-0.5\height}{\includegraphics[width=0.35\linewidth, trim={1cm, 1cm, 1cm, .33cm}, clip]{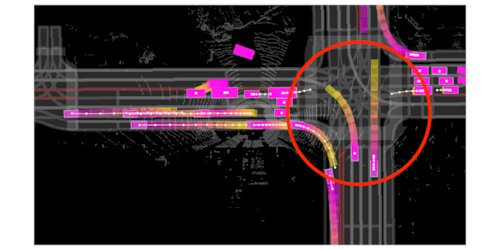}}
    \end{tabular}
    \caption{\textbf{Scene-level samples}. Our latent variable model captures the different scene outcomes for possible states of a given traffic light intersection (vertical vs. horizontal traffic).
    }
    \label{fig:supp_scene_samples_2}
\end{figure}

\begin{figure}[t]
    
    \centering
    \begin{tabular} {@{}c@{\hspace{.1em}}c@{\hspace{.5em}}c}
        {} & \textbf{Sample 1} & \textbf{Sample 2} \\
        \rotatebox[origin=c]{90}{\textbf{SpAGNN}} &
        \raisebox{-0.5\height}{\includegraphics[width=0.35\linewidth, trim={1cm, 1cm, 1cm, .33cm}, clip]{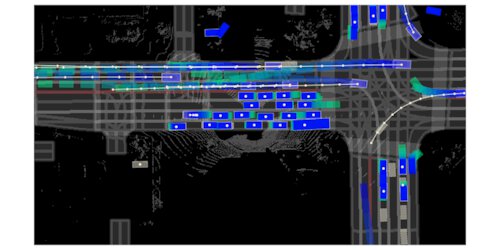}} &
        \raisebox{-0.5\height}{\includegraphics[width=0.35\linewidth, trim={1cm, 1cm, 1cm, .33cm}, clip]{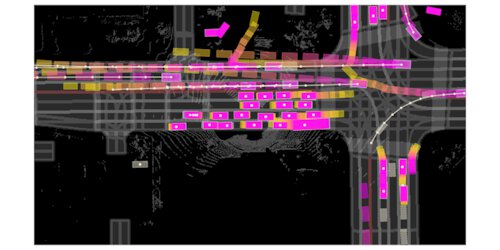}} \\[7.5ex]
        \rotatebox[origin=c]{90}{\textbf{MTP}} &
        \raisebox{-0.5\height}{\includegraphics[width=0.35\linewidth, trim={1cm, 1cm, 1cm, .33cm}, clip]{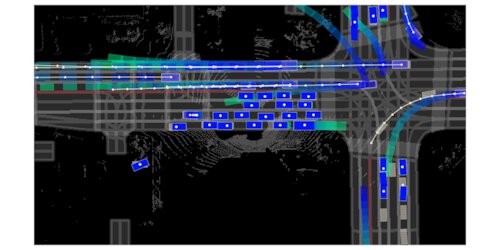}} &
        \raisebox{-0.5\height}{\includegraphics[width=0.35\linewidth, trim={1cm, 1cm, 1cm, .33cm}, clip]{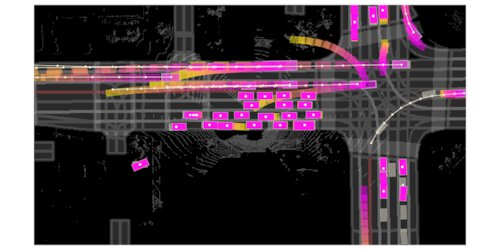}} \\[7.5ex]
        \rotatebox[origin=c]{90}{\textbf{MultiPath}} &
        \raisebox{-0.5\height}{\includegraphics[width=0.35\linewidth, trim={1cm, 1cm, 1cm, .33cm}, clip]{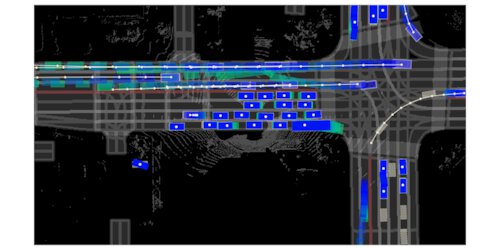}} &
        \raisebox{-0.5\height}{\includegraphics[width=0.35\linewidth, trim={1cm, 1cm, 1cm, .33cm}, clip]{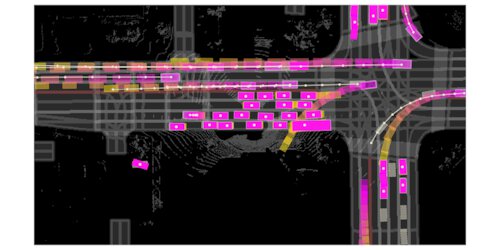}} \\[7.5ex]
        \rotatebox[origin=c]{90}{\textbf{R2P2}} &
        \raisebox{-0.5\height}{\includegraphics[width=0.35\linewidth, trim={1cm, 1cm, 1cm, .33cm}, clip]{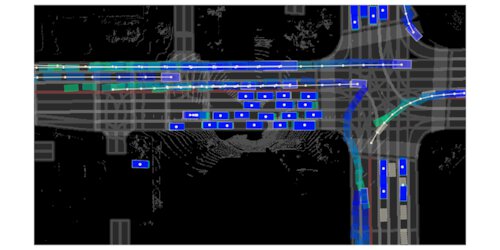}} &
        \raisebox{-0.5\height}{\includegraphics[width=0.35\linewidth, trim={1cm, 1cm, 1cm, .33cm}, clip]{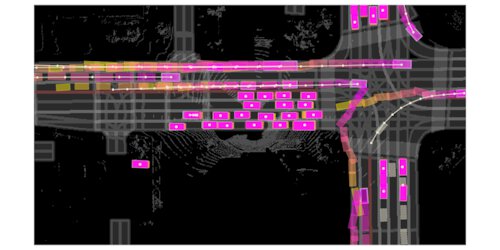}} \\[7.5ex]
        \rotatebox[origin=c]{90}{\textbf{ESP}} &
        \raisebox{-0.5\height}{\includegraphics[width=0.35\linewidth, trim={1cm, 1cm, 1cm, .33cm}, clip]{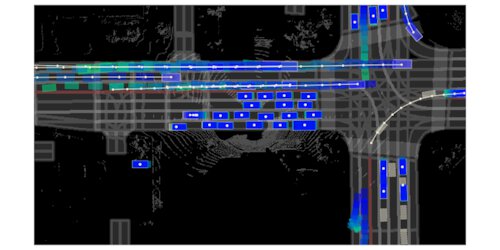}} &
        \raisebox{-0.5\height}{\includegraphics[width=0.35\linewidth, trim={1cm, 1cm, 1cm, .33cm}, clip]{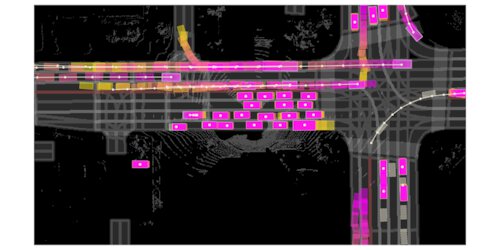}} \\[7.5ex]
        \rotatebox[origin=c]{90}{\textbf{MFP}} &
        \raisebox{-0.5\height}{\includegraphics[width=0.35\linewidth, trim={1cm, 1cm, 1cm, .33cm}, clip]{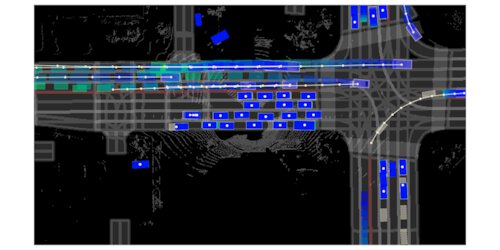}} &
        \raisebox{-0.5\height}{\includegraphics[width=0.35\linewidth, trim={1cm, 1cm, 1cm, .33cm}, clip]{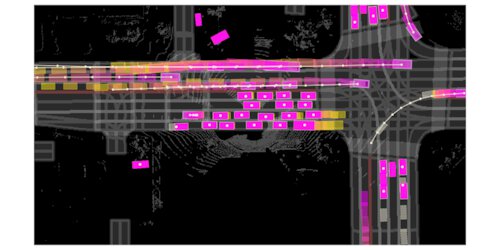}} \\[7.5ex]
        \rotatebox[origin=c]{90}{\textbf{RoR}} &
        \raisebox{-0.5\height}{\includegraphics[width=0.35\linewidth, trim={1cm, 1cm, 1cm, .33cm}, clip]{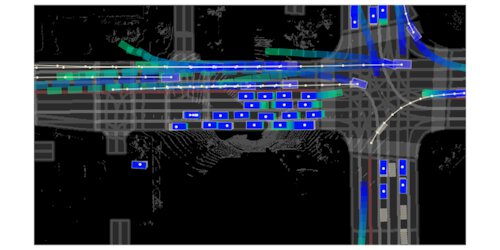}} &
        \raisebox{-0.5\height}{\includegraphics[width=0.35\linewidth, trim={1cm, 1cm, 1cm, .33cm}, clip]{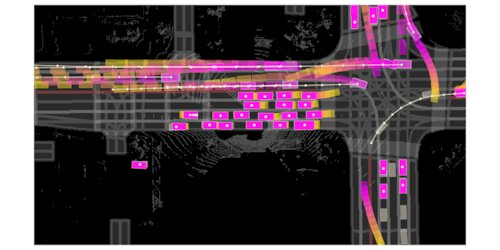}} \\[7.5ex]
        \rotatebox[origin=c]{90}{\textbf{ILVM}} &
        \raisebox{-0.5\height}{\includegraphics[width=0.35\linewidth, trim={1cm, 1cm, 1cm, .33cm}, clip]{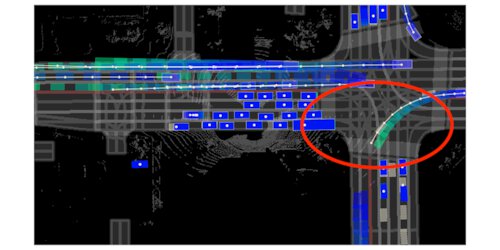}} &
        \raisebox{-0.5\height}{\includegraphics[width=0.35\linewidth, trim={1cm, 1cm, 1cm, .33cm}, clip]{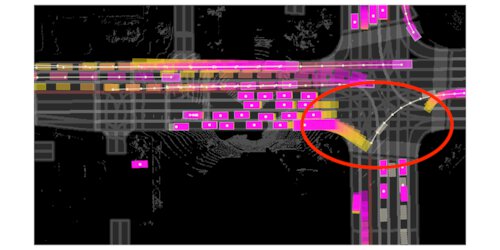}}
    \end{tabular}
    \caption{\textbf{Scene-level samples}. Our latent variable model captures whether the bus will proceed with the right turn, or the left-turning vehicle will.
    }
    \label{fig:supp_scene_samples_3}
\end{figure}

\begin{figure}[t]
    
    \centering
    \begin{tabular} {@{}c@{\hspace{.1em}}c@{\hspace{.5em}}c}
        {} & \textbf{Sample 1} & \textbf{Sample 2} \\
        \rotatebox[origin=c]{90}{\textbf{SpAGNN}} &
        \raisebox{-0.5\height}{\includegraphics[width=0.35\linewidth, trim={1cm, 1cm, 1cm, .33cm}, clip]{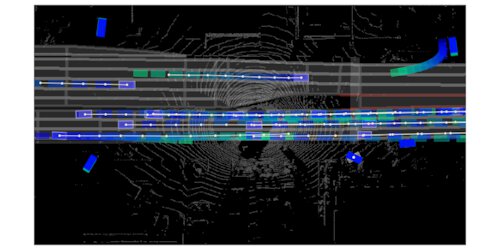}} &
        \raisebox{-0.5\height}{\includegraphics[width=0.35\linewidth, trim={1cm, 1cm, 1cm, .33cm}, clip]{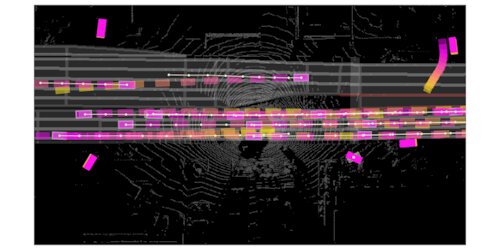}} \\[7.5ex]
        \rotatebox[origin=c]{90}{\textbf{MTP}} &
        \raisebox{-0.5\height}{\includegraphics[width=0.35\linewidth, trim={1cm, 1cm, 1cm, .33cm}, clip]{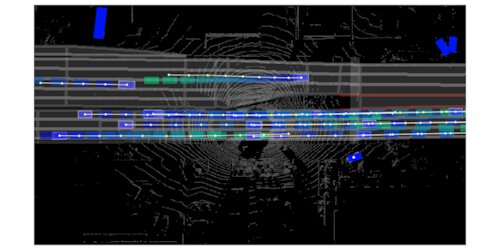}} &
        \raisebox{-0.5\height}{\includegraphics[width=0.35\linewidth, trim={1cm, 1cm, 1cm, .33cm}, clip]{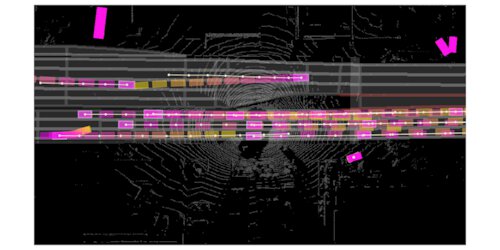}} \\[7.5ex]
        \rotatebox[origin=c]{90}{\textbf{MultiPath}} &
        \raisebox{-0.5\height}{\includegraphics[width=0.35\linewidth, trim={1cm, 1cm, 1cm, .33cm}, clip]{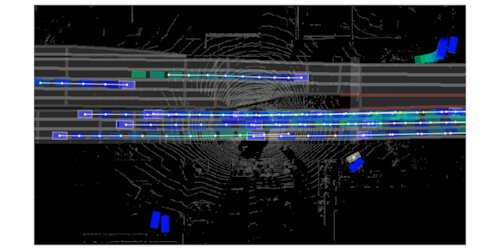}} &
        \raisebox{-0.5\height}{\includegraphics[width=0.35\linewidth, trim={1cm, 1cm, 1cm, .33cm}, clip]{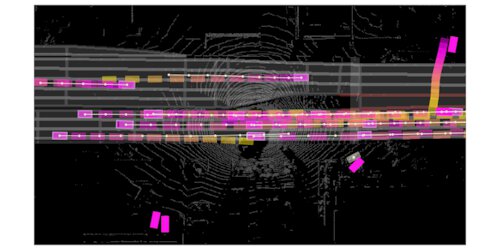}} \\[7.5ex]
        \rotatebox[origin=c]{90}{\textbf{R2P2}} &
        \raisebox{-0.5\height}{\includegraphics[width=0.35\linewidth, trim={1cm, 1cm, 1cm, .33cm}, clip]{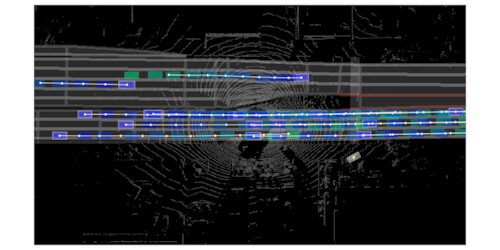}} &
        \raisebox{-0.5\height}{\includegraphics[width=0.35\linewidth, trim={1cm, 1cm, 1cm, .33cm}, clip]{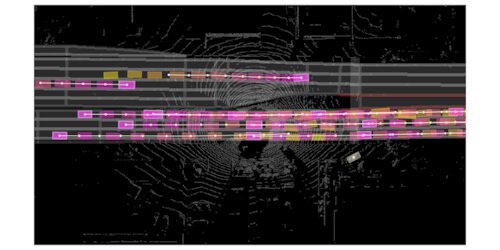}} \\[7.5ex]
        \rotatebox[origin=c]{90}{\textbf{ESP}} &
        \raisebox{-0.5\height}{\includegraphics[width=0.35\linewidth, trim={1cm, 1cm, 1cm, .33cm}, clip]{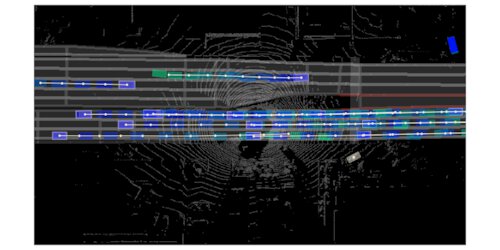}} &
        \raisebox{-0.5\height}{\includegraphics[width=0.35\linewidth, trim={1cm, 1cm, 1cm, .33cm}, clip]{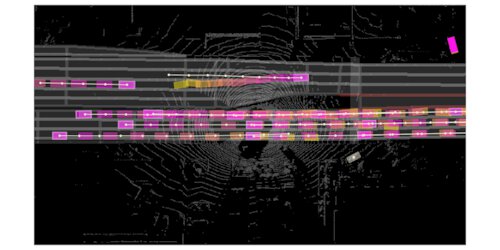}} \\[7.5ex]
        \rotatebox[origin=c]{90}{\textbf{MFP}} &
        \raisebox{-0.5\height}{\includegraphics[width=0.35\linewidth, trim={1cm, 1cm, 1cm, .33cm}, clip]{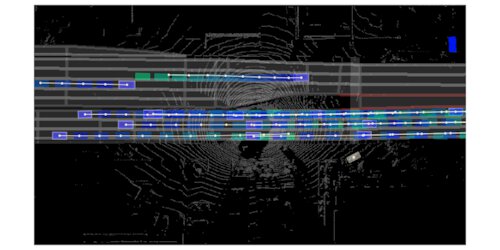}} &
        \raisebox{-0.5\height}{\includegraphics[width=0.35\linewidth, trim={1cm, 1cm, 1cm, .33cm}, clip]{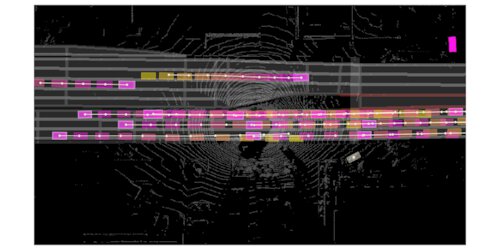}} \\[7.5ex]
        \rotatebox[origin=c]{90}{\textbf{RoR}} &
        \raisebox{-0.5\height}{\includegraphics[width=0.35\linewidth, trim={1cm, 1cm, 1cm, .33cm}, clip]{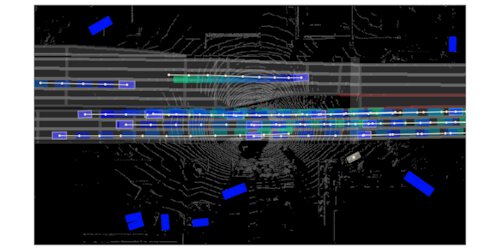}} &
        \raisebox{-0.5\height}{\includegraphics[width=0.35\linewidth, trim={1cm, 1cm, 1cm, .33cm}, clip]{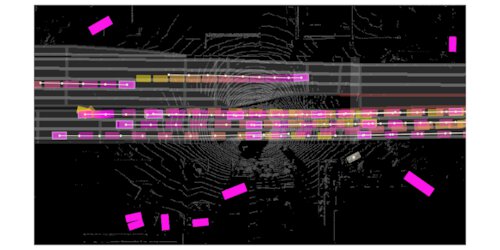}} \\[7.5ex]
        \rotatebox[origin=c]{90}{\textbf{ILVM}} &
        \raisebox{-0.5\height}{\includegraphics[width=0.35\linewidth, trim={1cm, 1cm, 1cm, .33cm}, clip]{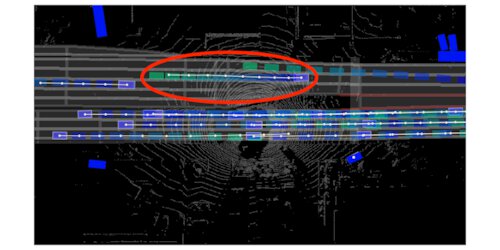}} &
        \raisebox{-0.5\height}{\includegraphics[width=0.35\linewidth, trim={1cm, 1cm, 1cm, .33cm}, clip]{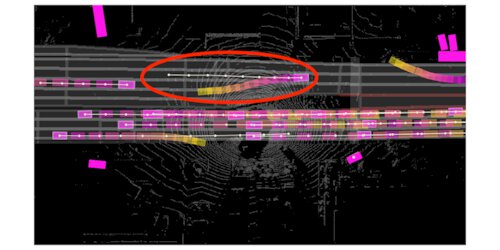}}
    \end{tabular}
    \caption{\textbf{Scene-level samples}. Our latent variable model captures multiple realistic futures (including lane changes) that respect the map geometries and are dynamically feasible.
    }
    \label{fig:supp_scene_samples_4}
\end{figure}

In Figure~\ref{fig:supp_scene_samples_1}, \ref{fig:supp_scene_samples_2}, \ref{fig:supp_scene_samples_3}, \ref{fig:supp_scene_samples_4} we showcase the scene consistency of the samples generated from our model.
For these visualizations, each row corresponds to a model, and we show 2 scene-level samples for each model to characterize the joint distribution. 

More concretely, 
we show the two \emph{most distinct} samples by averaging the pairwise Euclidean distance between all samples.
We empirically find that this selection methodology yields representative samples and insight into how well the models learn scene-level social interaction between agents.

\subsubsection{Latent Space Interpolation:}
\begin{figure}[t]
    \centering
    \begin{tabular} {@{}c@{\hspace{.1em}}c@{\hspace{.5em}}c@{\hspace{.5em}}c}
        {} & \textbf{Scenario 1} & \textbf{Scenario 2} & \textbf{Scenario 3} \\
        \rotatebox[origin=c]{0}{\textbf{$Z^1$}} &
        \raisebox{-0.5\height}{\includegraphics[width=0.325\linewidth, trim={1cm, 1cm, 1cm, .33cm}, clip]{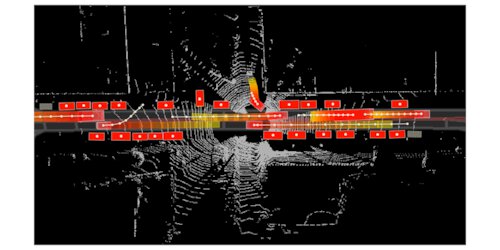}} &
        \raisebox{-0.5\height}{\includegraphics[width=0.325\linewidth, trim={1cm, 1cm, 1cm, .33cm}, clip]{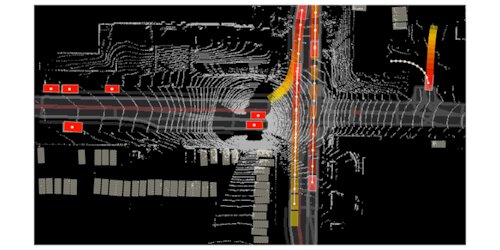}} &
        \raisebox{-0.5\height}{\includegraphics[width=0.325\linewidth, trim={1cm, 1cm, 1cm, .33cm}, clip]{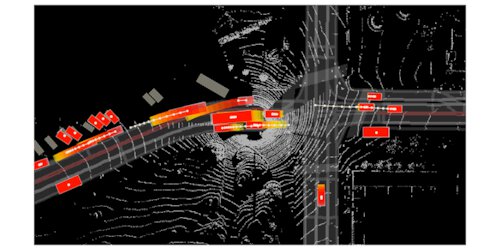}} \\[7.5ex]
        \rotatebox[origin=c]{90}{} &
        \raisebox{-0.5\height}{\includegraphics[width=0.325\linewidth, trim={1cm, 1cm, 1cm, .33cm}, clip]{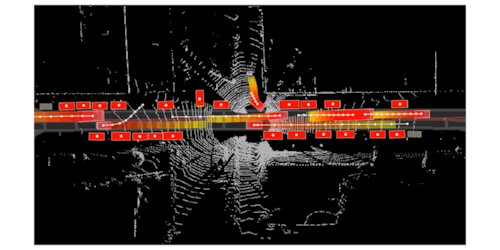}} &
        \raisebox{-0.5\height}{\includegraphics[width=0.325\linewidth, trim={1cm, 1cm, 1cm, .33cm}, clip]{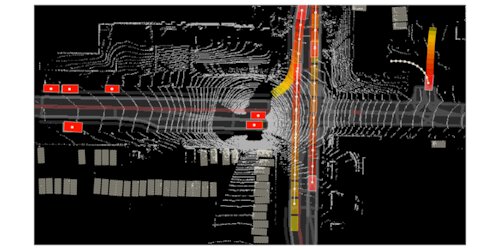}} &
        \raisebox{-0.5\height}{\includegraphics[width=0.325\linewidth, trim={1cm, 1cm, 1cm, .33cm}, clip]{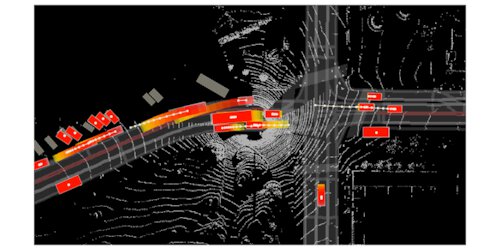}} \\[7.5ex]
        \rotatebox[origin=c]{90}{} &
        \raisebox{-0.5\height}{\includegraphics[width=0.325\linewidth, trim={1cm, 1cm, 1cm, .33cm}, clip]{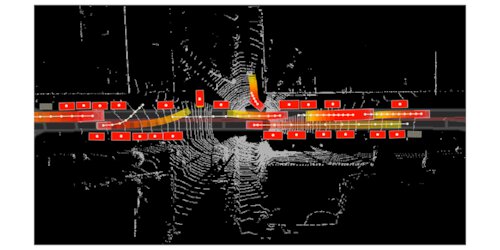}} &
        \raisebox{-0.5\height}{\includegraphics[width=0.325\linewidth, trim={1cm, 1cm, 1cm, .33cm}, clip]{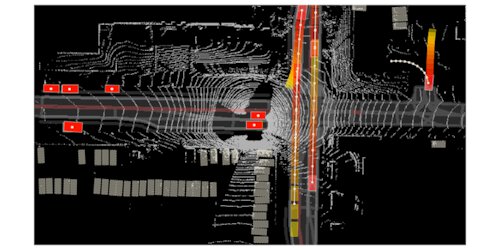}} &
        \raisebox{-0.5\height}{\includegraphics[width=0.325\linewidth, trim={1cm, 1cm, 1cm, .33cm}, clip]{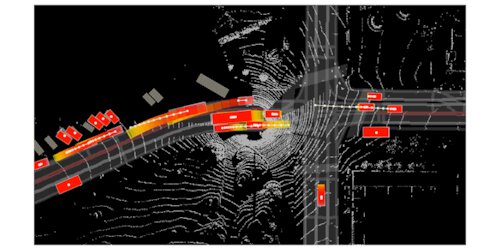}} \\[7.5ex]
        \rotatebox[origin=c]{0}{\textbf{$Z^2$}} &
        \raisebox{-0.5\height}{\includegraphics[width=0.325\linewidth, trim={1cm, 1cm, 1cm, .33cm}, clip]{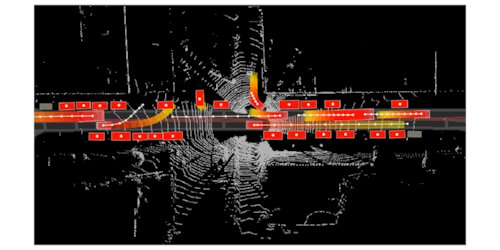}} &
        \raisebox{-0.5\height}{\includegraphics[width=0.325\linewidth, trim={1cm, 1cm, 1cm, .33cm}, clip]{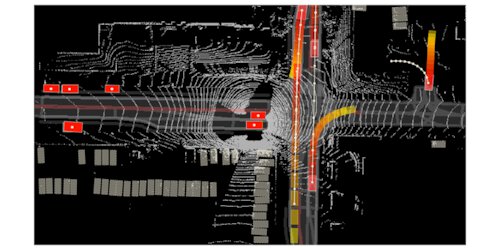}} &
        \raisebox{-0.5\height}{\includegraphics[width=0.325\linewidth, trim={1cm, 1cm, 1cm, .33cm}, clip]{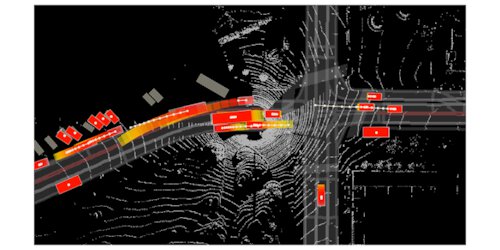}} \\[7.5ex]
    \end{tabular}
    \caption{\textbf{Latent space interpolation:} \emph{Scenario 1} showcases a complex interaction between 3 vehicles: when the 2 vehicles in the road predict turning or slow moving trajectories, the third vehicle pulls out of the driveway, and when the 2 vehicles in the road keep constant velocity to go straight, the vehicle in the driveway yields. \emph{Scenario 2}  turning vs. going straight behavior with smooth transitions. \emph{Scenario 3} we can see how the speed of 2 cars that follow each other vary consistently.} 
    \label{fig:interpolated_samples_0}
\end{figure}

\begin{figure}[t]
    \centering
    \begin{tabular} {@{}c@{\hspace{.1em}}c@{\hspace{.5em}}c@{\hspace{.5em}}c}
        {} & \textbf{Scenario 4} & \textbf{Scenario 5} & \textbf{Scenario 6} \\
        \rotatebox[origin=c]{0}{\textbf{$Z^1$}} &
        \raisebox{-0.5\height}{\includegraphics[width=0.325\linewidth, trim={1cm, 1cm, 1cm, .33cm}, clip]{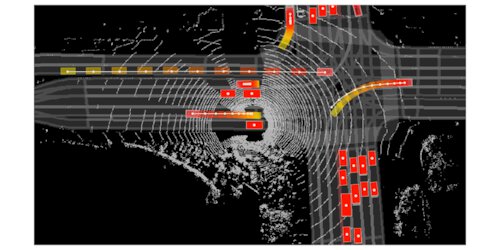}} &
        \raisebox{-0.5\height}{\includegraphics[width=0.325\linewidth, trim={1cm, 1cm, 1cm, .33cm}, clip]{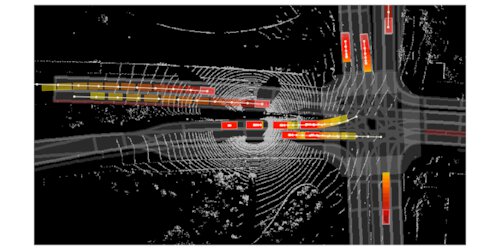}} &
        \raisebox{-0.5\height}{\includegraphics[width=0.325\linewidth, trim={1cm, 1cm, 1cm, .33cm}, clip]{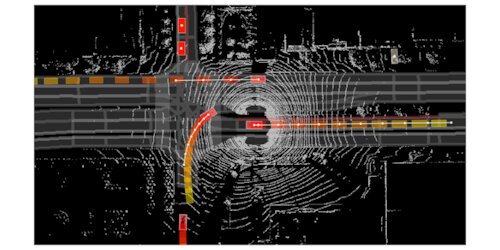}} \\[7.5ex]
        \rotatebox[origin=c]{90}{} &
        \raisebox{-0.5\height}{\includegraphics[width=0.325\linewidth, trim={1cm, 1cm, 1cm, .33cm}, clip]{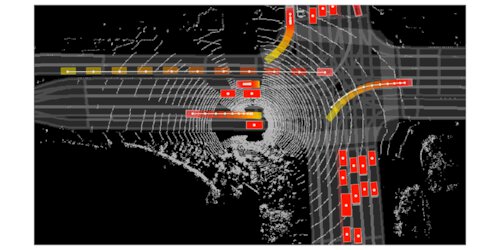}} &
        \raisebox{-0.5\height}{\includegraphics[width=0.325\linewidth, trim={1cm, 1cm, 1cm, .33cm}, clip]{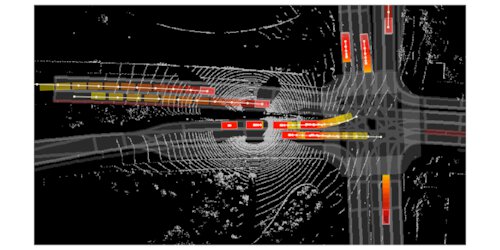}} &
        \raisebox{-0.5\height}{\includegraphics[width=0.325\linewidth, trim={1cm, 1cm, 1cm, .33cm}, clip]{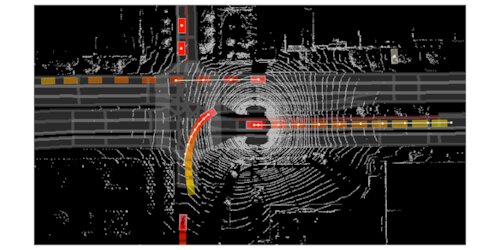}} \\[7.5ex]
        \rotatebox[origin=c]{90}{} &
        \raisebox{-0.5\height}{\includegraphics[width=0.325\linewidth, trim={1cm, 1cm, 1cm, .33cm}, clip]{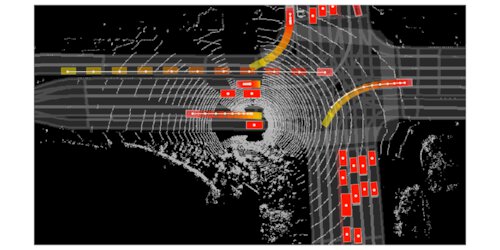}} &
        \raisebox{-0.5\height}{\includegraphics[width=0.325\linewidth, trim={1cm, 1cm, 1cm, .33cm}, clip]{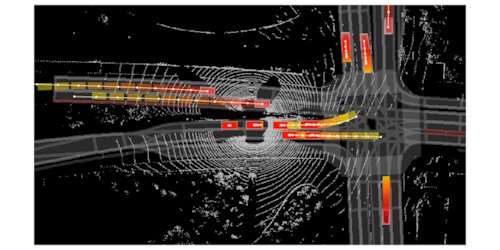}} &
        \raisebox{-0.5\height}{\includegraphics[width=0.325\linewidth, trim={1cm, 1cm, 1cm, .33cm}, clip]{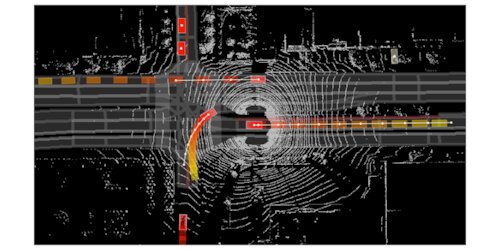}} \\[7.5ex]
        \rotatebox[origin=c]{0}{\textbf{$Z^2$}} &
        \raisebox{-0.5\height}{\includegraphics[width=0.325\linewidth, trim={1cm, 1cm, 1cm, .33cm}, clip]{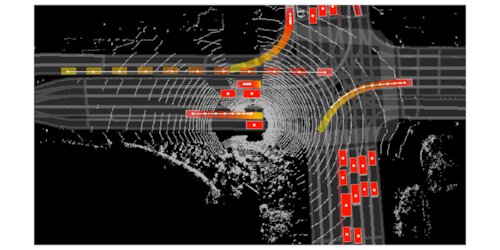}} &
        \raisebox{-0.5\height}{\includegraphics[width=0.325\linewidth, trim={1cm, 1cm, 1cm, .33cm}, clip]{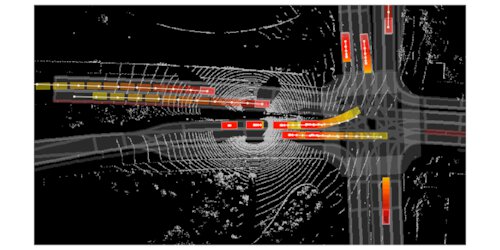}} &
        \raisebox{-0.5\height}{\includegraphics[width=0.325\linewidth, trim={1cm, 1cm, 1cm, .33cm}, clip]{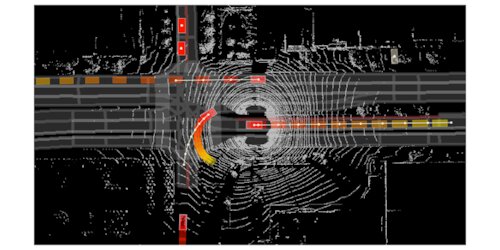}} \\[7.5ex]
    \end{tabular}
    \caption{\textbf{Latent space interpolation:} \emph{Scenario 4} and \emph{Scenario 5} showcase smooth transitions between different speed profiles when turning and going straight at an intersection. \emph{Scenario 6} we can see all the range of possibilities from a left-turn to a u-turn, which is a pretty rare event.}
    \label{fig:interpolated_samples_1}
\end{figure}

In Figs.~\ref{fig:interpolated_samples_0} and \ref{fig:interpolated_samples_1}, we take the 2 most distinct samples as in the previous scene sample visualizations, and show the resulting futures when performing linear interpolation in the latent space. We show that the interpolated latent points still produce semantically meaningful trajectories for all the actors in the scene, and capture scene level variations including multi-agent interactions.
More precisely, $Z^1$ and $Z^2$ are the latent samples that map into the most distinct futures out of 50. The rows in between correspond to the linear interpolation of the latent space, and different columns to different scenarios.

\subsubsection{Overall Sample Quality:}
\begin{figure}[t]
    \centering
    \begin{tabular} {@{}c@{\hspace{.1em}}c@{\hspace{.5em}}c@{\hspace{.5em}}c}
        {} & \textbf{Scenario 1} & \textbf{Scenario 2} & \textbf{Scenario 3} \\
        \rotatebox[origin=c]{90}{\textbf{SpAGNN}} &
        \raisebox{-0.5\height}{\includegraphics[width=0.325\linewidth, trim={1cm, 1cm, 1cm, .33cm}, clip]{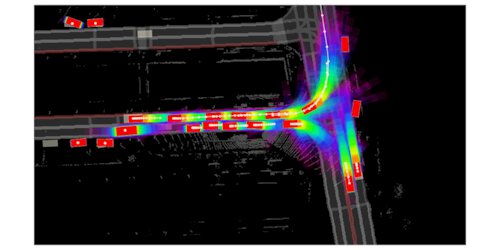}} &
        \raisebox{-0.5\height}{\includegraphics[width=0.325\linewidth, trim={1cm, 1cm, 1cm, .33cm}, clip]{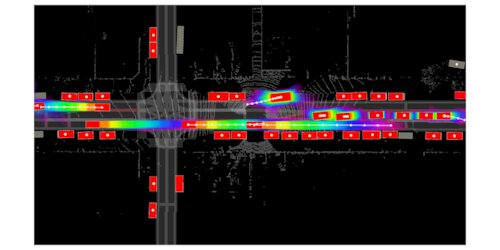}} &
        \raisebox{-0.5\height}{\includegraphics[width=0.325\linewidth, trim={1cm, 1cm, 1cm, .33cm}, clip]{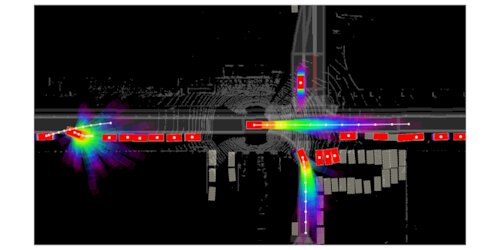}} \\[7.5ex]
        \rotatebox[origin=c]{90}{\textbf{MTP}} &
        \raisebox{-0.5\height}{\includegraphics[width=0.325\linewidth, trim={1cm, 1cm, 1cm, .33cm}, clip]{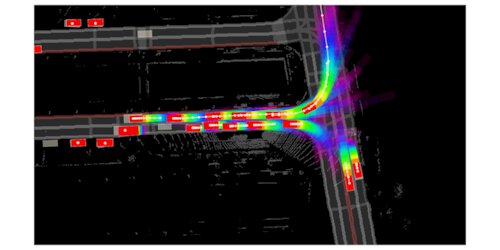}} &
        \raisebox{-0.5\height}{\includegraphics[width=0.325\linewidth, trim={1cm, 1cm, 1cm, .33cm}, clip]{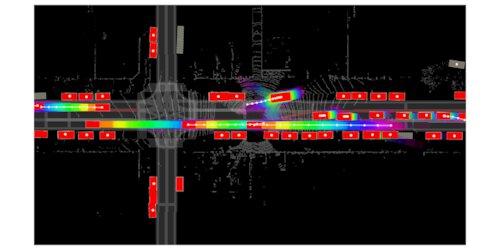}} &
        \raisebox{-0.5\height}{\includegraphics[width=0.325\linewidth, trim={1cm, 1cm, 1cm, .33cm}, clip]{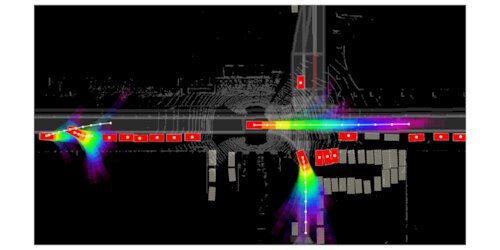}} \\[7.5ex]
        \rotatebox[origin=c]{90}{\textbf{MultiPath}} &
        \raisebox{-0.5\height}{\includegraphics[width=0.325\linewidth, trim={1cm, 1cm, 1cm, .33cm}, clip]{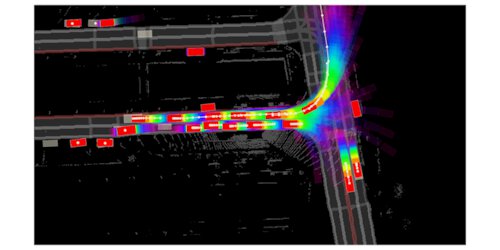}} &
        \raisebox{-0.5\height}{\includegraphics[width=0.325\linewidth, trim={1cm, 1cm, 1cm, .33cm}, clip]{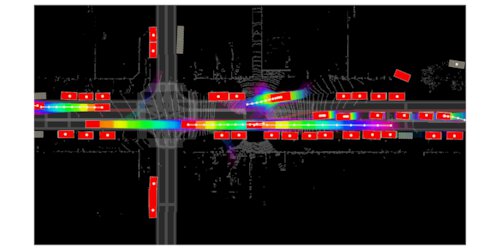}} &
        \raisebox{-0.5\height}{\includegraphics[width=0.325\linewidth, trim={1cm, 1cm, 1cm, .33cm}, clip]{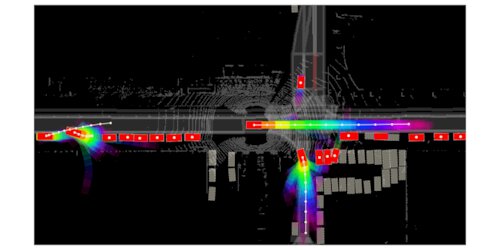}} \\[7.5ex]
        \rotatebox[origin=c]{90}{\textbf{R2P2}} &
        \raisebox{-0.5\height}{\includegraphics[width=0.325\linewidth, trim={1cm, 1cm, 1cm, .33cm}, clip]{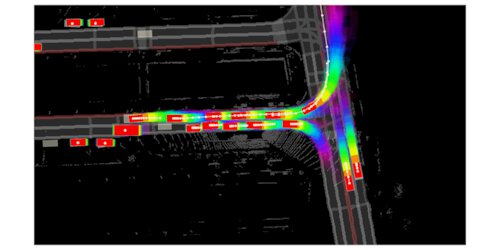}} &
        \raisebox{-0.5\height}{\includegraphics[width=0.325\linewidth, trim={1cm, 1cm, 1cm, .33cm}, clip]{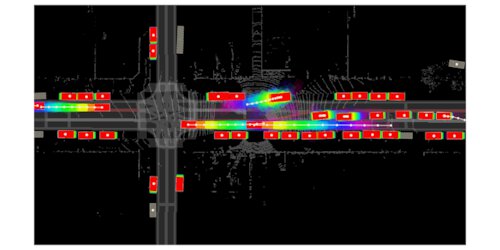}} &
        \raisebox{-0.5\height}{\includegraphics[width=0.325\linewidth, trim={1cm, 1cm, 1cm, .33cm}, clip]{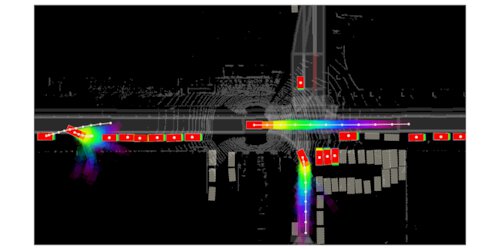}} \\[7.5ex]
        \rotatebox[origin=c]{90}{\textbf{ESP}} &
        \raisebox{-0.5\height}{\includegraphics[width=0.325\linewidth, trim={1cm, 1cm, 1cm, .33cm}, clip]{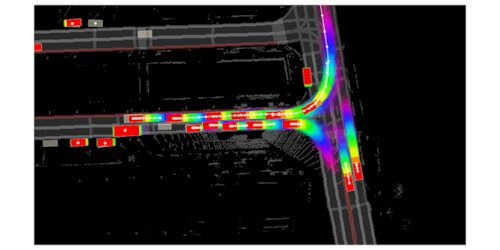}} &
        \raisebox{-0.5\height}{\includegraphics[width=0.325\linewidth, trim={1cm, 1cm, 1cm, .33cm}, clip]{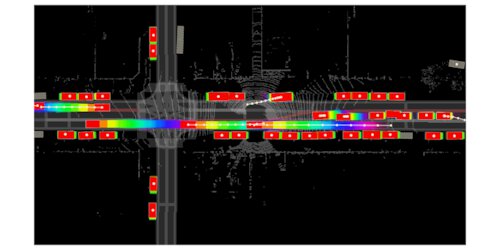}} &
        \raisebox{-0.5\height}{\includegraphics[width=0.325\linewidth, trim={1cm, 1cm, 1cm, .33cm}, clip]{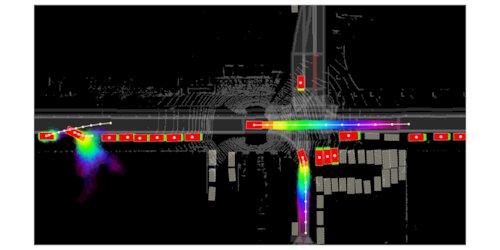}} \\[7.5ex]
        \rotatebox[origin=c]{90}{\textbf{MFP}} &
        \raisebox{-0.5\height}{\includegraphics[width=0.325\linewidth, trim={1cm, 1cm, 1cm, .33cm}, clip]{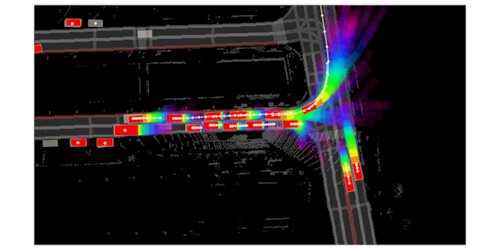}} &
        \raisebox{-0.5\height}{\includegraphics[width=0.325\linewidth, trim={1cm, 1cm, 1cm, .33cm}, clip]{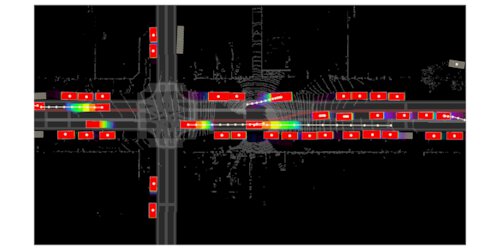}} &
        \raisebox{-0.5\height}{\includegraphics[width=0.325\linewidth, trim={1cm, 1cm, 1cm, .33cm}, clip]{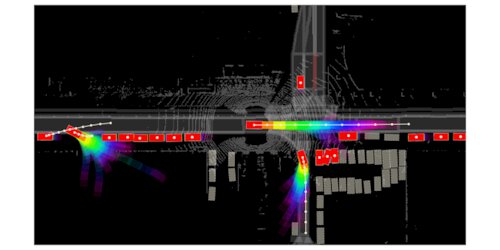}} \\[7.5ex]
        \rotatebox[origin=c]{90}{\textbf{RoR}} &
        \raisebox{-0.5\height}{\includegraphics[width=0.325\linewidth, trim={1cm, 1cm, 1cm, .33cm}, clip]{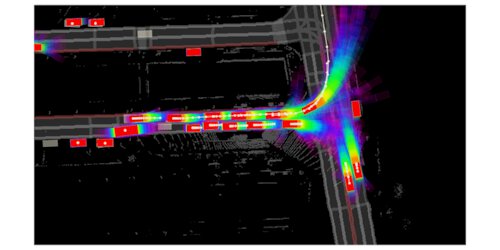}} &
        \raisebox{-0.5\height}{\includegraphics[width=0.325\linewidth, trim={1cm, 1cm, 1cm, .33cm}, clip]{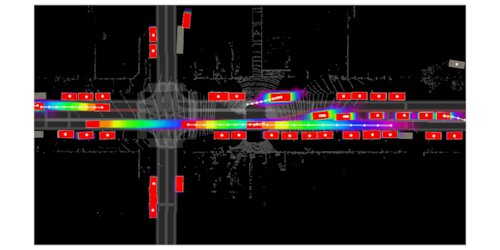}} &
        \raisebox{-0.5\height}{\includegraphics[width=0.325\linewidth, trim={1cm, 1cm, 1cm, .33cm}, clip]{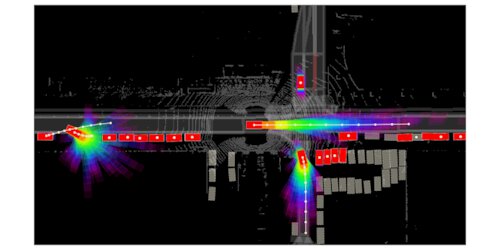}} \\[7.5ex]
        \rotatebox[origin=c]{90}{\textbf{ILVM}} &
        \raisebox{-0.5\height}{\includegraphics[width=0.325\linewidth, trim={1cm, 1cm, 1cm, .33cm}, clip]{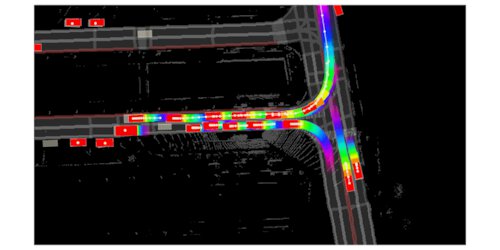}} &
        \raisebox{-0.5\height}{\includegraphics[width=0.325\linewidth, trim={1cm, 1cm, 1cm, .33cm}, clip]{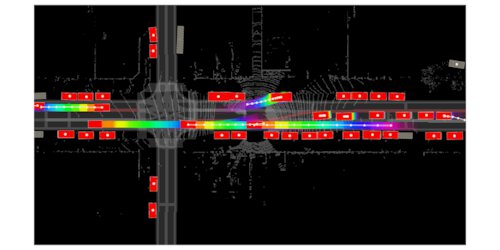}} &
        \raisebox{-0.5\height}{\includegraphics[width=0.325\linewidth, trim={1cm, 1cm, 1cm, .33cm}, clip]{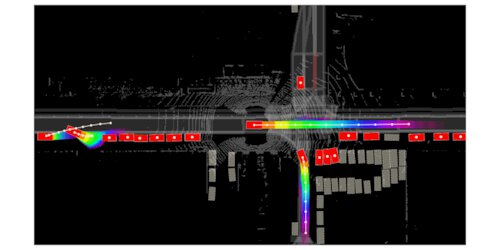}} \\[7.5ex]
    \end{tabular}
    \caption{\textbf{Overall Sample Quality:} \emph{Scenario 1} showcases a T-intersection with fast-moving turns. \emph{Scenario 2} is interesting because there is a vehicle coming out of a parking spot, which is not very frequent in driving logs. \emph{Scenario 3} captures a vehicle maneuvering into a parking spot, also an unusual event.}
    \label{fig:qualitative_prediction_supplementary_1}
\end{figure}

\begin{figure}[t]
    \centering
    \begin{tabular} {@{}c@{\hspace{.1em}}c@{\hspace{.5em}}c@{\hspace{.5em}}c}
        {} & \textbf{Scenario 4} & \textbf{Scenario 5} & \textbf{Scenario 6} \\
        \rotatebox[origin=c]{90}{\textbf{SpAGNN}} &
        \raisebox{-0.5\height}{\includegraphics[width=0.325\linewidth, trim={1cm, 1cm, 1cm, .33cm}, clip]{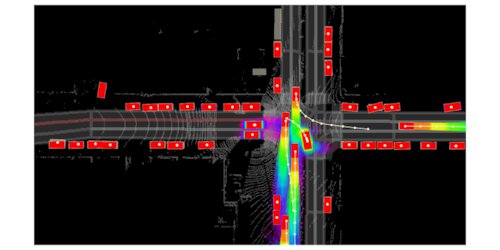}} &
        \raisebox{-0.5\height}{\includegraphics[width=0.325\linewidth, trim={1cm, 1cm, 1cm, .33cm}, clip]{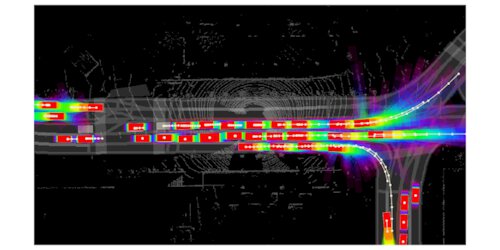}} &
        \raisebox{-0.5\height}{\includegraphics[width=0.325\linewidth, trim={1cm, 1cm, 1cm, .33cm}, clip]{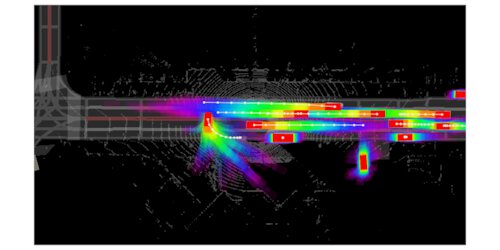}} \\[7.5ex]
        \rotatebox[origin=c]{90}{\textbf{MTP}} &
        \raisebox{-0.5\height}{\includegraphics[width=0.325\linewidth, trim={1cm, 1cm, 1cm, .33cm}, clip]{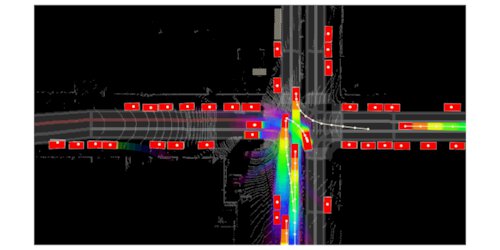}} &
        \raisebox{-0.5\height}{\includegraphics[width=0.325\linewidth, trim={1cm, 1cm, 1cm, .33cm}, clip]{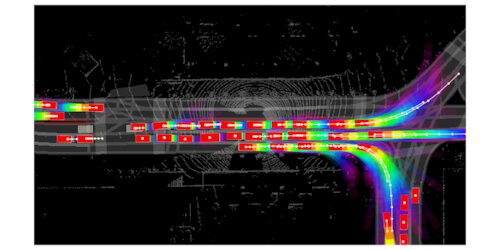}} &
        \raisebox{-0.5\height}{\includegraphics[width=0.325\linewidth, trim={1cm, 1cm, 1cm, .33cm}, clip]{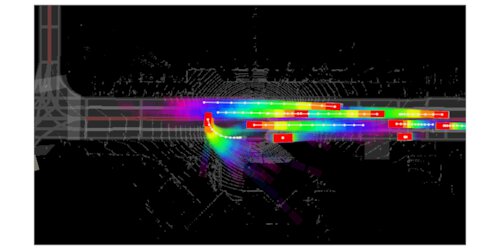}} \\[7.5ex]
        \rotatebox[origin=c]{90}{\textbf{MultiPath}} &
        \raisebox{-0.5\height}{\includegraphics[width=0.325\linewidth, trim={1cm, 1cm, 1cm, .33cm}, clip]{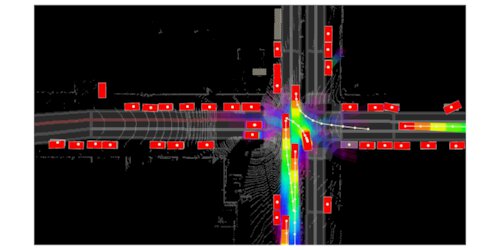}} &
        \raisebox{-0.5\height}{\includegraphics[width=0.325\linewidth, trim={1cm, 1cm, 1cm, .33cm}, clip]{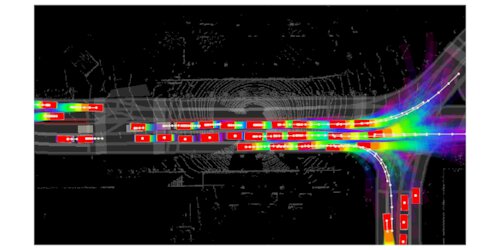}} &
        \raisebox{-0.5\height}{\includegraphics[width=0.325\linewidth, trim={1cm, 1cm, 1cm, .33cm}, clip]{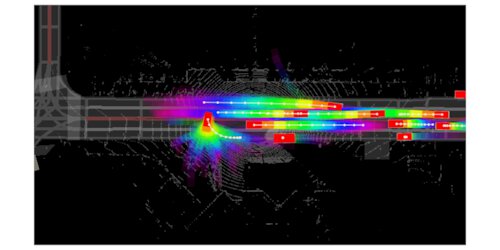}} \\[7.5ex]
        \rotatebox[origin=c]{90}{\textbf{R2P2}} &
        \raisebox{-0.5\height}{\includegraphics[width=0.325\linewidth, trim={1cm, 1cm, 1cm, .33cm}, clip]{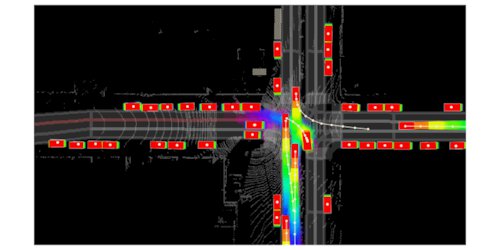}} &
        \raisebox{-0.5\height}{\includegraphics[width=0.325\linewidth, trim={1cm, 1cm, 1cm, .33cm}, clip]{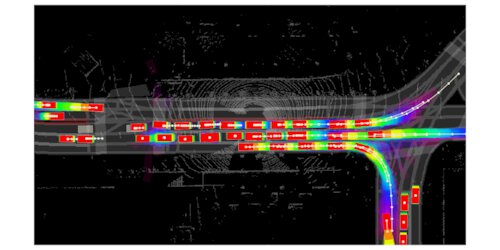}} &
        \raisebox{-0.5\height}{\includegraphics[width=0.325\linewidth, trim={1cm, 1cm, 1cm, .33cm}, clip]{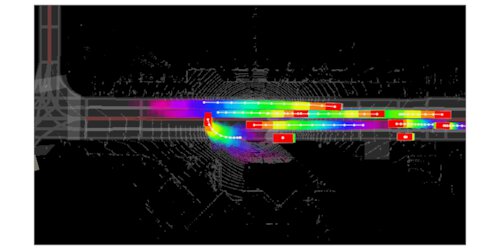}} \\[7.5ex]
        \rotatebox[origin=c]{90}{\textbf{ESP}} &
        \raisebox{-0.5\height}{\includegraphics[width=0.325\linewidth, trim={1cm, 1cm, 1cm, .33cm}, clip]{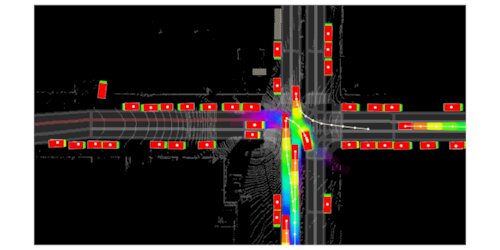}} &
        \raisebox{-0.5\height}{\includegraphics[width=0.325\linewidth, trim={1cm, 1cm, 1cm, .33cm}, clip]{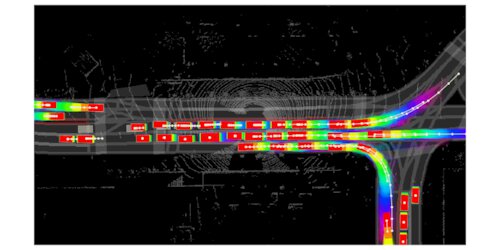}} &
        \raisebox{-0.5\height}{\includegraphics[width=0.325\linewidth, trim={1cm, 1cm, 1cm, .33cm}, clip]{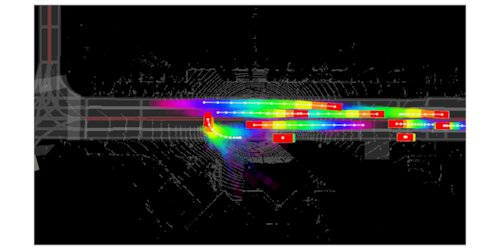}} \\[7.5ex]
        \rotatebox[origin=c]{90}{\textbf{MFP}} &
        \raisebox{-0.5\height}{\includegraphics[width=0.325\linewidth, trim={1cm, 1cm, 1cm, .33cm}, clip]{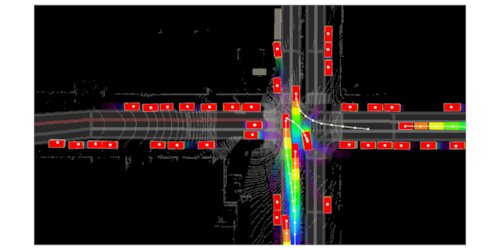}} &
        \raisebox{-0.5\height}{\includegraphics[width=0.325\linewidth, trim={1cm, 1cm, 1cm, .33cm}, clip]{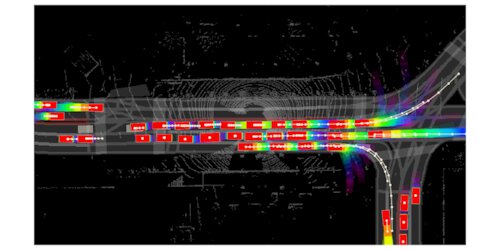}} &
        \raisebox{-0.5\height}{\includegraphics[width=0.325\linewidth, trim={1cm, 1cm, 1cm, .33cm}, clip]{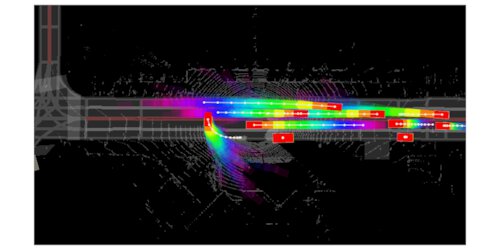}} \\[7.5ex]
        \rotatebox[origin=c]{90}{\textbf{RoR}} &
        \raisebox{-0.5\height}{\includegraphics[width=0.325\linewidth, trim={1cm, 1cm, 1cm, .33cm}, clip]{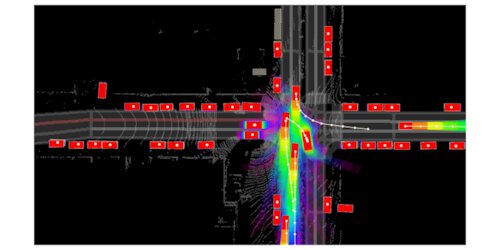}} &
        \raisebox{-0.5\height}{\includegraphics[width=0.325\linewidth, trim={1cm, 1cm, 1cm, .33cm}, clip]{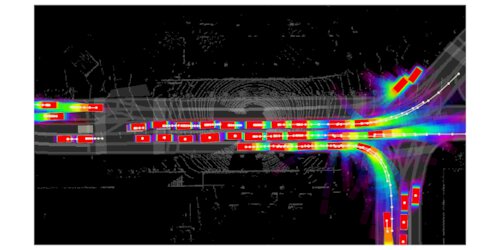}} &
        \raisebox{-0.5\height}{\includegraphics[width=0.325\linewidth, trim={1cm, 1cm, 1cm, .33cm}, clip]{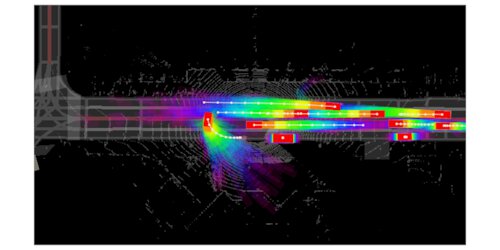}} \\[7.5ex]
        \rotatebox[origin=c]{90}{\textbf{ILVM}} &
        \raisebox{-0.5\height}{\includegraphics[width=0.325\linewidth, trim={1cm, 1cm, 1cm, .33cm}, clip]{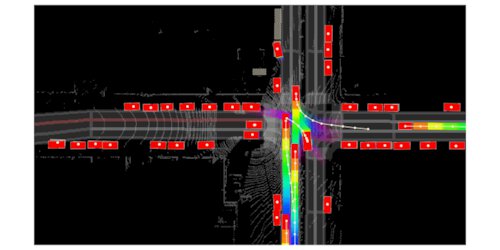}} &
        \raisebox{-0.5\height}{\includegraphics[width=0.325\linewidth, trim={1cm, 1cm, 1cm, .33cm}, clip]{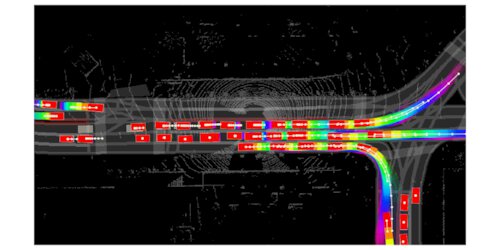}} &
        \raisebox{-0.5\height}{\includegraphics[width=0.325\linewidth, trim={1cm, 1cm, 1cm, .33cm}, clip]{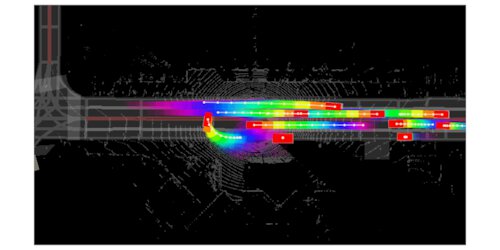}} \\[7.5ex]
    \end{tabular}
    \caption{\textbf{Overall Sample Quality:} \emph{Scenario 4} showcases a complex interaction between 3 vehicles at a 4-way intersection, where our model identifies sharp "modes". However, like all other baselines, it misses (or predicts with very low probability) the true mode of the vehicle facing south and left-turning. \emph{Scenario 5} showcases fast moving traffic, where our model can predict an accurate distribution with very low entropy even at 5 seconds into the future. \emph{Scenario 6} Our model captures a vehicle performing a U-turn.}
    \label{fig:qualitative_prediction_supplementary_2}
\end{figure}

\begin{figure}[t]
    \centering
    \begin{tabular} {@{}c@{\hspace{.1em}}c@{\hspace{.5em}}c@{\hspace{.5em}}c}
        {} & \textbf{Scenario 7} & \textbf{Scenario 8} & \textbf{Scenario 9} \\
        \rotatebox[origin=c]{90}{\textbf{SpAGNN}} &
        \raisebox{-0.5\height}{\includegraphics[width=0.325\linewidth, trim={1cm, 1cm, 1cm, .33cm}, clip]{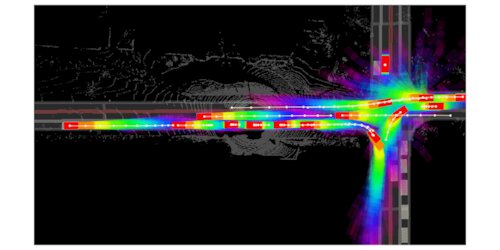}} &
        \raisebox{-0.5\height}{\includegraphics[width=0.325\linewidth, trim={1cm, 1cm, 1cm, .33cm}, clip]{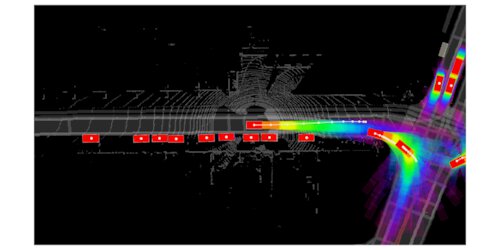}} &
        \raisebox{-0.5\height}{\includegraphics[width=0.325\linewidth, trim={1cm, 1cm, 1cm, .33cm}, clip]{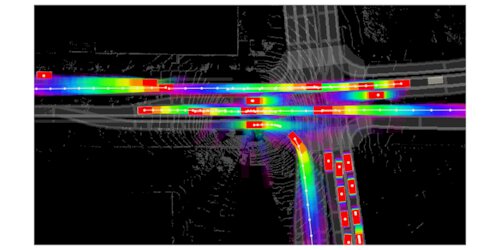}} \\[7.5ex]
        \rotatebox[origin=c]{90}{\textbf{MTP}} &
        \raisebox{-0.5\height}{\includegraphics[width=0.325\linewidth, trim={1cm, 1cm, 1cm, .33cm}, clip]{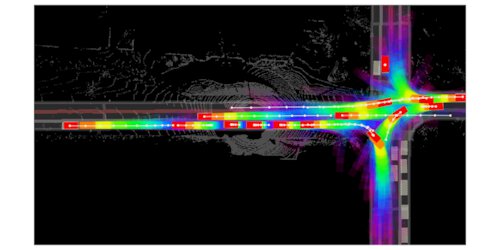}} &
        \raisebox{-0.5\height}{\includegraphics[width=0.325\linewidth, trim={1cm, 1cm, 1cm, .33cm}, clip]{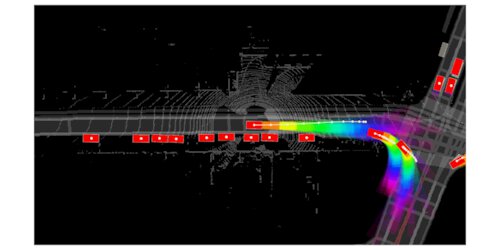}} &
        \raisebox{-0.5\height}{\includegraphics[width=0.325\linewidth, trim={1cm, 1cm, 1cm, .33cm}, clip]{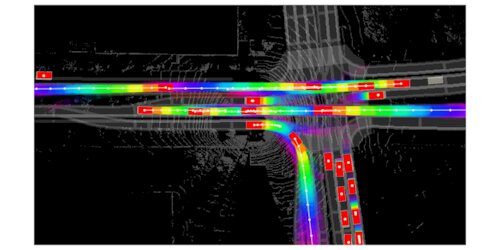}} \\[7.5ex]
        \rotatebox[origin=c]{90}{\textbf{MultiPath}} &
        \raisebox{-0.5\height}{\includegraphics[width=0.325\linewidth, trim={1cm, 1cm, 1cm, .33cm}, clip]{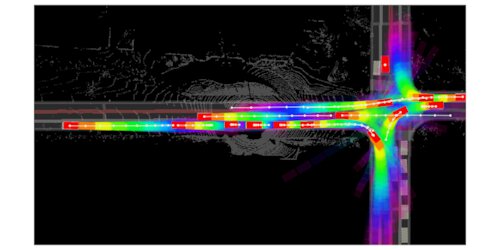}} &
        \raisebox{-0.5\height}{\includegraphics[width=0.325\linewidth, trim={1cm, 1cm, 1cm, .33cm}, clip]{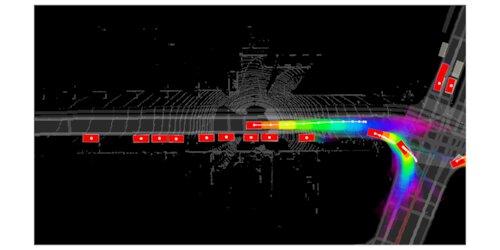}} &
        \raisebox{-0.5\height}{\includegraphics[width=0.325\linewidth, trim={1cm, 1cm, 1cm, .33cm}, clip]{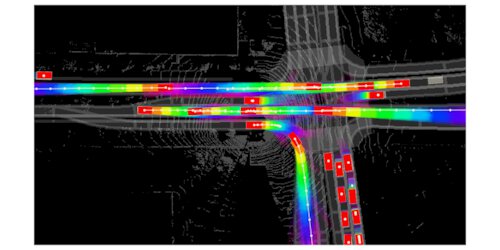}} \\[7.5ex]
        \rotatebox[origin=c]{90}{\textbf{R2P2}} &
        \raisebox{-0.5\height}{\includegraphics[width=0.325\linewidth, trim={1cm, 1cm, 1cm, .33cm}, clip]{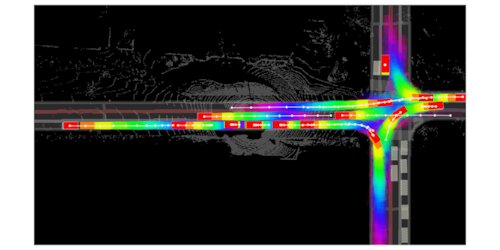}} &
        \raisebox{-0.5\height}{\includegraphics[width=0.325\linewidth, trim={1cm, 1cm, 1cm, .33cm}, clip]{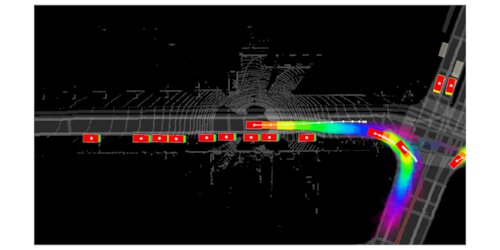}} &
        \raisebox{-0.5\height}{\includegraphics[width=0.325\linewidth, trim={1cm, 1cm, 1cm, .33cm}, clip]{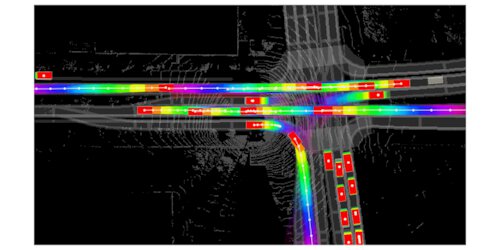}} \\[7.5ex]
        \rotatebox[origin=c]{90}{\textbf{ESP}} &
        \raisebox{-0.5\height}{\includegraphics[width=0.325\linewidth, trim={1cm, 1cm, 1cm, .33cm}, clip]{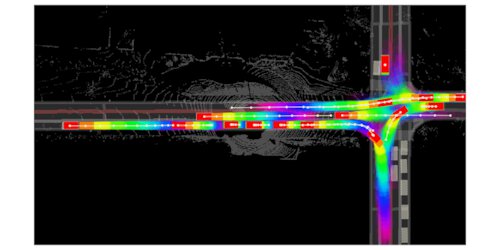}} &
        \raisebox{-0.5\height}{\includegraphics[width=0.325\linewidth, trim={1cm, 1cm, 1cm, .33cm}, clip]{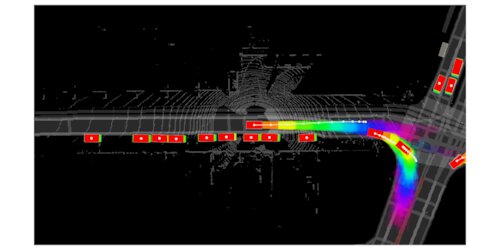}} &
        \raisebox{-0.5\height}{\includegraphics[width=0.325\linewidth, trim={1cm, 1cm, 1cm, .33cm}, clip]{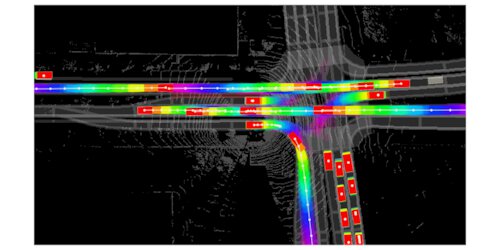}} \\[7.5ex]
        \rotatebox[origin=c]{90}{\textbf{MFP}} &
        \raisebox{-0.5\height}{\includegraphics[width=0.325\linewidth, trim={1cm, 1cm, 1cm, .33cm}, clip]{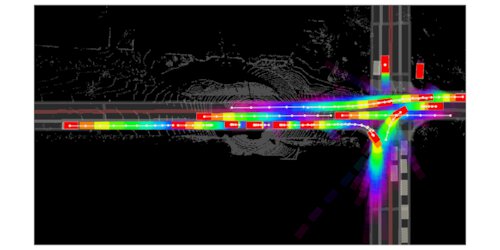}} &
        \raisebox{-0.5\height}{\includegraphics[width=0.325\linewidth, trim={1cm, 1cm, 1cm, .33cm}, clip]{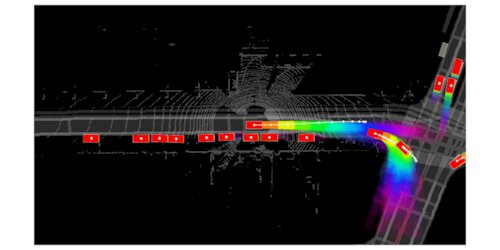}} &
        \raisebox{-0.5\height}{\includegraphics[width=0.325\linewidth, trim={1cm, 1cm, 1cm, .33cm}, clip]{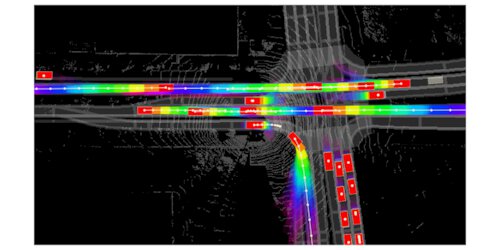}} \\[7.5ex]
        \rotatebox[origin=c]{90}{\textbf{RoR}} &
        \raisebox{-0.5\height}{\includegraphics[width=0.325\linewidth, trim={1cm, 1cm, 1cm, .33cm}, clip]{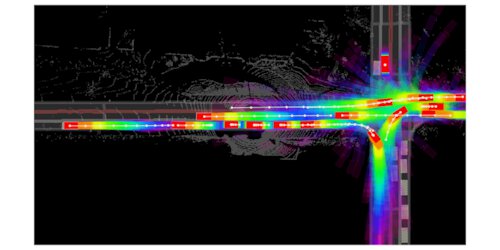}} &
        \raisebox{-0.5\height}{\includegraphics[width=0.325\linewidth, trim={1cm, 1cm, 1cm, .33cm}, clip]{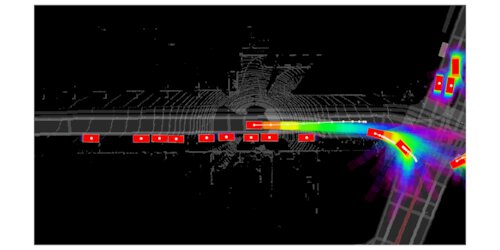}} &
        \raisebox{-0.5\height}{\includegraphics[width=0.325\linewidth, trim={1cm, 1cm, 1cm, .33cm}, clip]{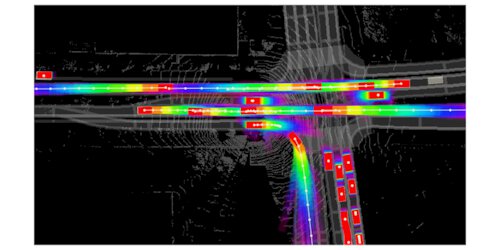}} \\[7.5ex]
        \rotatebox[origin=c]{90}{\textbf{ILVM}} &
        \raisebox{-0.5\height}{\includegraphics[width=0.325\linewidth, trim={1cm, 1cm, 1cm, .33cm}, clip]{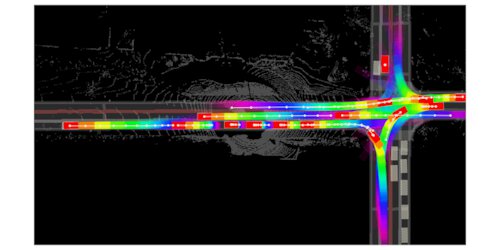}} &
        \raisebox{-0.5\height}{\includegraphics[width=0.325\linewidth, trim={1cm, 1cm, 1cm, .33cm}, clip]{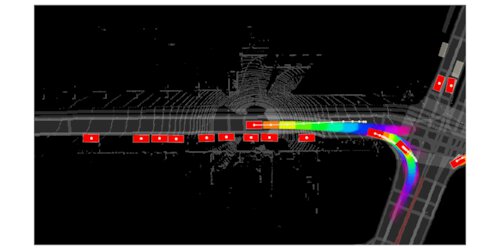}} &
        \raisebox{-0.5\height}{\includegraphics[width=0.325\linewidth, trim={1cm, 1cm, 1cm, .33cm}, clip]{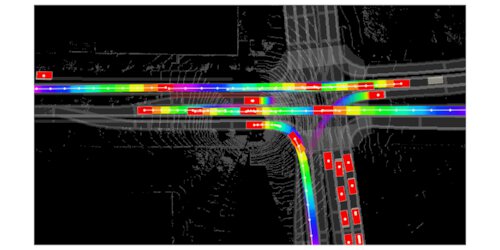}} \\[7.5ex]
    \end{tabular}
    \caption{\textbf{Overall Sample Quality:} We highlighted the accuracy and sharpness of our predictions in \emph{Scenarios 7, 8 and 9}.}
    \label{fig:qualitative_prediction_supplementary_3}
\end{figure}

In Figs.~\ref{fig:qualitative_prediction_supplementary_1}, \ref{fig:qualitative_prediction_supplementary_2}, \ref{fig:qualitative_prediction_supplementary_3}, we show additional qualitative results for motion forecasting, comparing our method to the baselines in a wide range of urban scenarios, one per column. 
We blend 50 scene sample trajectories with transparency.
Time is encoded in the rainbow color map ranging from red (0s) to pink (5s). 
This can be seen as a sample-based characterization of the per-actor marginal distributions.
We can see that our method generally produces more accurate and less entropic distributions that better understand the map topology and multi-agent interactions.

\subsubsection{Ego-motion Planning:}
\begin{figure}[t]
    
    \begin{tabular}{@{}c@{\hspace{.1em}}c@{\hspace{.5em}}c@{\hspace{.5em}}c}
        \textbf{} &
        \textbf{\textsc{Multipath}} &
        \textbf{\textsc{ESP}} &
        \textbf{\textsc{ILVM}} \\

        \rotatebox[origin=c]{90}{\textbf{t = 0 s}} &
        \raisebox{-0.5\height}{\includegraphics[trim={1cm .33cm 1cm .33cm},clip, width=0.33\textwidth]{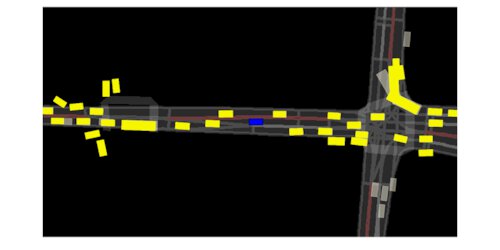}} &
        \raisebox{-0.5\height}{\includegraphics[trim={1cm .33cm 1cm .33cm},clip, width=0.33\textwidth]{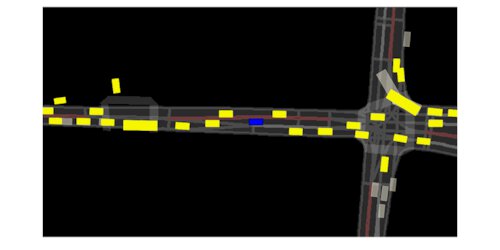}} &
        \raisebox{-0.5\height}{\includegraphics[trim={1cm .33cm 1cm .33cm},clip, width=0.33\textwidth]{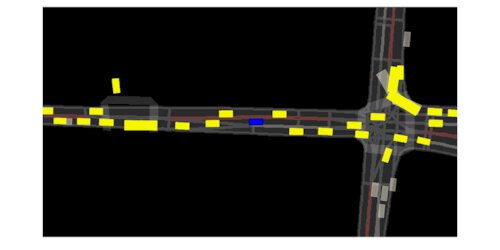}} 
        \\[7.5ex]
        
        \rotatebox[origin=c]{90}{\textbf{t = 1 s}} &
        \raisebox{-0.5\height}{\includegraphics[trim={1cm .33cm 1cm .33cm},clip, width=0.33\textwidth]{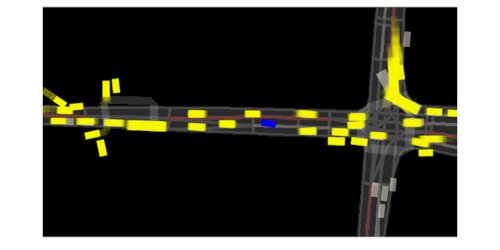}} &
        \raisebox{-0.5\height}{\includegraphics[trim={1cm .33cm 1cm .33cm},clip, width=0.33\textwidth]{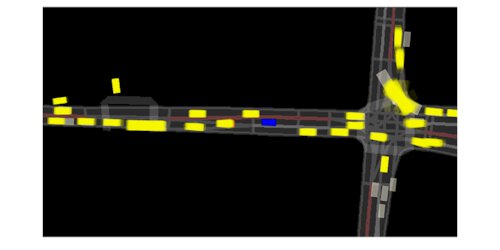}} &
        \raisebox{-0.5\height}{\includegraphics[trim={1cm .33cm 1cm .33cm},clip, width=0.33\textwidth]{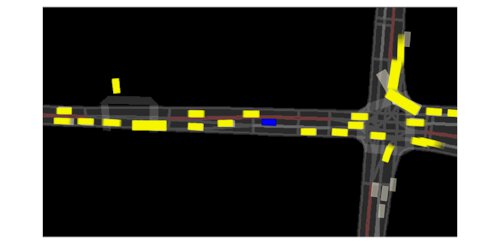}} 
        \\[7.5ex]

        \rotatebox[origin=c]{90}{\textbf{t = 2 s}} &
        \raisebox{-0.5\height}{\includegraphics[trim={1cm .33cm 1cm .33cm},clip, width=0.33\textwidth]{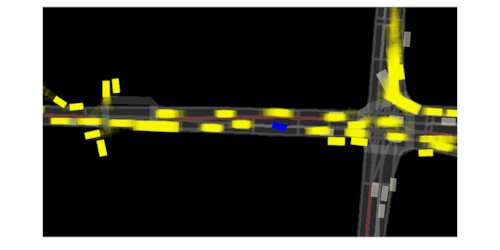}} &
        \raisebox{-0.5\height}{\includegraphics[trim={1cm .33cm 1cm .33cm},clip, width=0.33\textwidth]{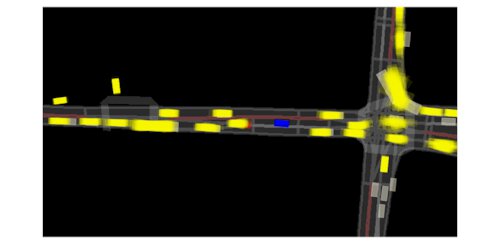}} &
        \raisebox{-0.5\height}{\includegraphics[trim={1cm .33cm 1cm .33cm},clip, width=0.33\textwidth]{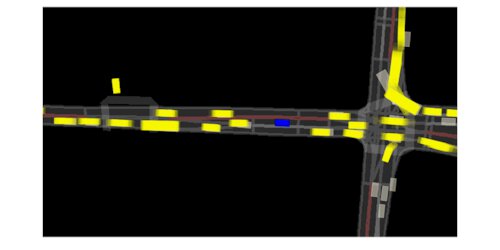}} 
        \\[7.5ex]

        \rotatebox[origin=c]{90}{\textbf{t = 3 s}} &
        \raisebox{-0.5\height}{\includegraphics[trim={1cm .33cm 1cm .33cm},clip, width=0.33\textwidth]{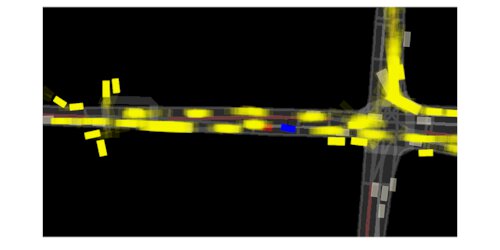}} &
        \raisebox{-0.5\height}{\includegraphics[trim={1cm .33cm 1cm .33cm},clip, width=0.33\textwidth]{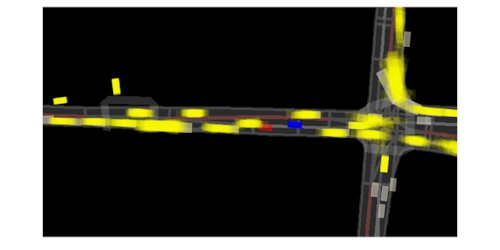}} &
        \raisebox{-0.5\height}{\includegraphics[trim={1cm .33cm 1cm .33cm},clip, width=0.33\textwidth]{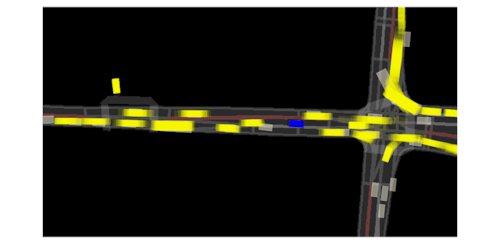}} 
        \\[7.5ex]

        \rotatebox[origin=c]{90}{\textbf{t = 4 s}} &
        \raisebox{-0.5\height}{\includegraphics[trim={1cm .33cm 1cm .33cm},clip, width=0.33\textwidth]{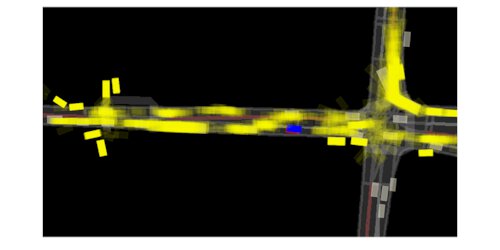}} &
        \raisebox{-0.5\height}{\includegraphics[trim={1cm .33cm 1cm .33cm},clip, width=0.33\textwidth]{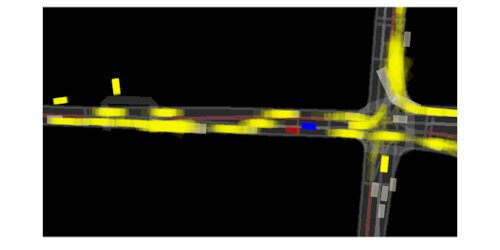}} &
        \raisebox{-0.5\height}{\includegraphics[trim={1cm .33cm 1cm .33cm},clip, width=0.33\textwidth]{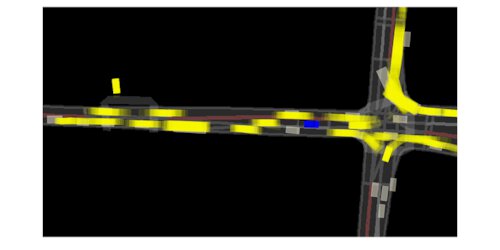}} 
        \\[7.5ex]

        \rotatebox[origin=c]{90}{\textbf{t = 5 s}} &
        \raisebox{-0.5\height}{\includegraphics[trim={1cm .33cm 1cm .33cm},clip, width=0.33\textwidth]{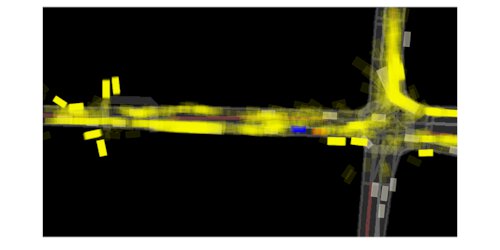}} &
        \raisebox{-0.5\height}{\includegraphics[trim={1cm .33cm 1cm .33cm},clip, width=0.33\textwidth]{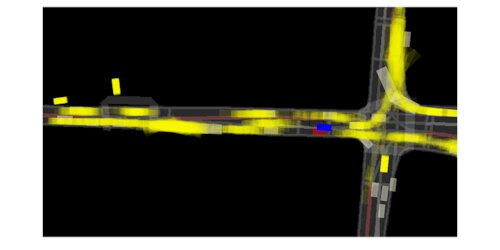}} &
        \raisebox{-0.5\height}{\includegraphics[trim={1cm .33cm 1cm .33cm},clip, width=0.33\textwidth]{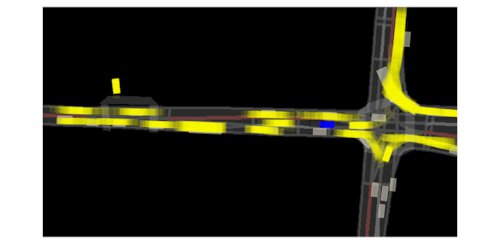}} 
        \\[7.5ex]
    \end{tabular}
    \caption{In this scenario, both MultiPath and ESP generate motion forecasts that get into the SDV lane, forcing it to lane change to its right, where it collides with an actual vehicle that is lane changing from behind the SDV and is not well captured by the prediction models, including ours.}
    \label{figure:planning_comparison_1}
    
\end{figure}

\begin{figure}[t]
    
    \begin{tabular}{@{}c@{\hspace{.1em}}c@{\hspace{.5em}}c@{\hspace{.5em}}c}
        \textbf{} &
        \textbf{\textsc{Multipath}} &
        \textbf{\textsc{ESP}} &
        \textbf{\textsc{ILVM}} \\

        \rotatebox[origin=c]{90}{\textbf{t = 0 s}} &
        \raisebox{-0.5\height}{\includegraphics[trim={1cm .33cm 1cm .33cm},clip, width=0.33\textwidth]{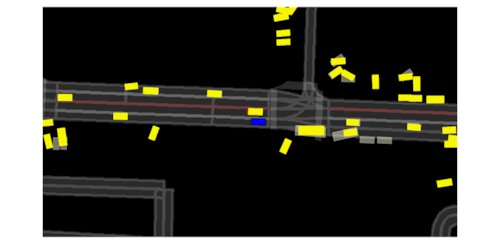}} &
        \raisebox{-0.5\height}{\includegraphics[trim={1cm .33cm 1cm .33cm},clip, width=0.33\textwidth]{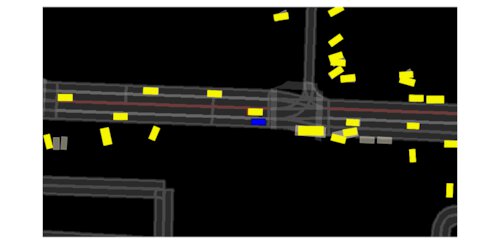}} &
        \raisebox{-0.5\height}{\includegraphics[trim={1cm .33cm 1cm .33cm},clip, width=0.33\textwidth]{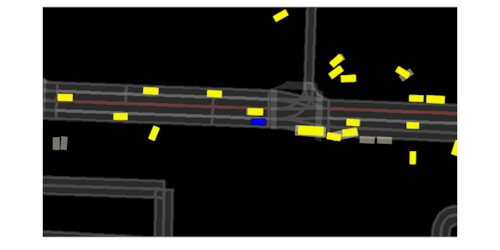}} 
        \\[7.5ex]
        
        \rotatebox[origin=c]{90}{\textbf{t = 1 s}} &
        \raisebox{-0.5\height}{\includegraphics[trim={1cm .33cm 1cm .33cm},clip, width=0.33\textwidth]{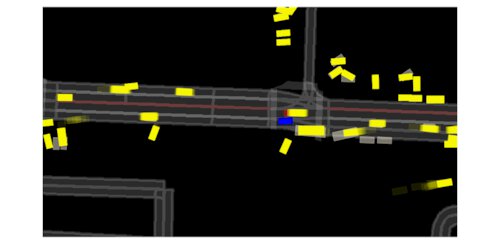}} &
        \raisebox{-0.5\height}{\includegraphics[trim={1cm .33cm 1cm .33cm},clip, width=0.33\textwidth]{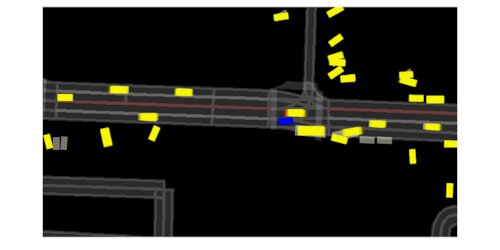}} &
        \raisebox{-0.5\height}{\includegraphics[trim={1cm .33cm 1cm .33cm},clip, width=0.33\textwidth]{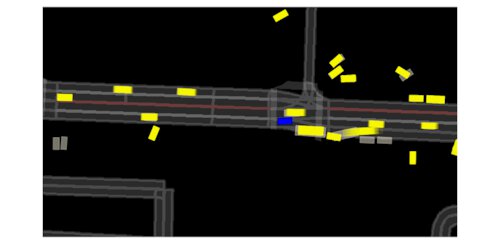}} 
        \\[7.5ex]

        \rotatebox[origin=c]{90}{\textbf{t = 2 s}} &
        \raisebox{-0.5\height}{\includegraphics[trim={1cm .33cm 1cm .33cm},clip, width=0.33\textwidth]{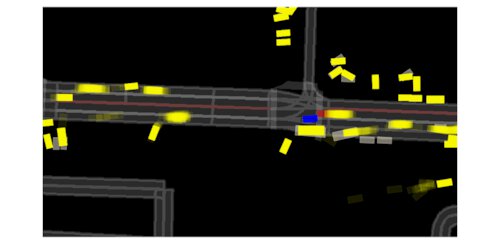}} &
        \raisebox{-0.5\height}{\includegraphics[trim={1cm .33cm 1cm .33cm},clip, width=0.33\textwidth]{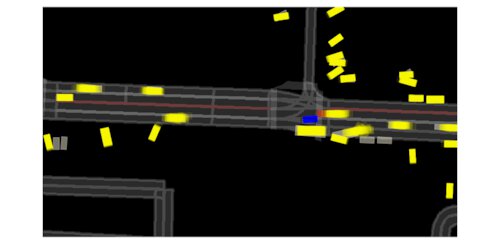}} &
        \raisebox{-0.5\height}{\includegraphics[trim={1cm .33cm 1cm .33cm},clip, width=0.33\textwidth]{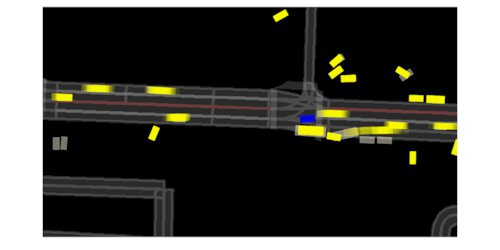}} 
        \\[7.5ex]

        \rotatebox[origin=c]{90}{\textbf{t = 3 s}} &
        \raisebox{-0.5\height}{\includegraphics[trim={1cm .33cm 1cm .33cm},clip, width=0.33\textwidth]{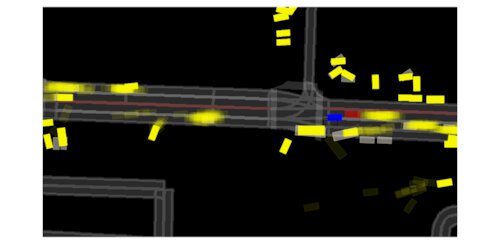}} &
        \raisebox{-0.5\height}{\includegraphics[trim={1cm .33cm 1cm .33cm},clip, width=0.33\textwidth]{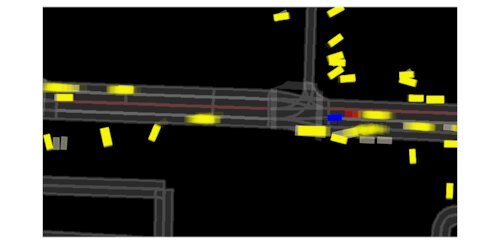}} &
        \raisebox{-0.5\height}{\includegraphics[trim={1cm .33cm 1cm .33cm},clip, width=0.33\textwidth]{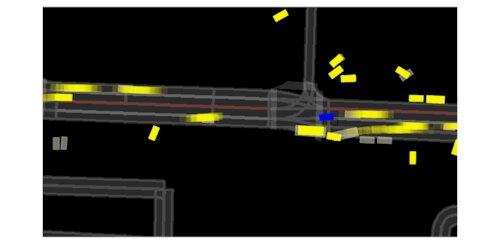}} 
        \\[7.5ex]

        \rotatebox[origin=c]{90}{\textbf{t = 4 s}} &
        \raisebox{-0.5\height}{\includegraphics[trim={1cm .33cm 1cm .33cm},clip, width=0.33\textwidth]{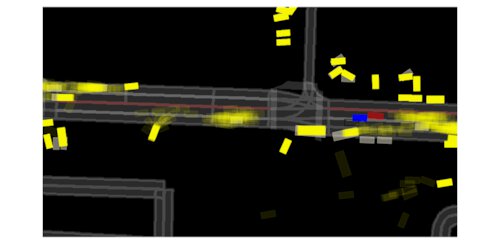}} &
        \raisebox{-0.5\height}{\includegraphics[trim={1cm .33cm 1cm .33cm},clip, width=0.33\textwidth]{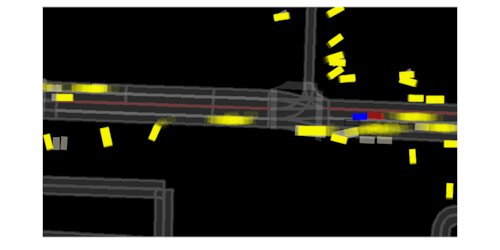}} &
        \raisebox{-0.5\height}{\includegraphics[trim={1cm .33cm 1cm .33cm},clip, width=0.33\textwidth]{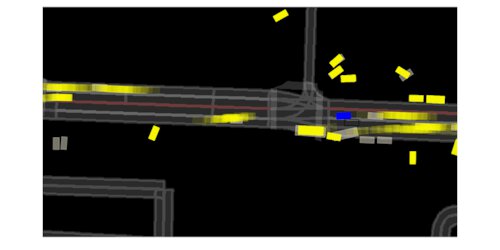}} 
        \\[7.5ex]

        \rotatebox[origin=c]{90}{\textbf{t = 5 s}} &
        \raisebox{-0.5\height}{\includegraphics[trim={1cm .33cm 1cm .33cm},clip, width=0.33\textwidth]{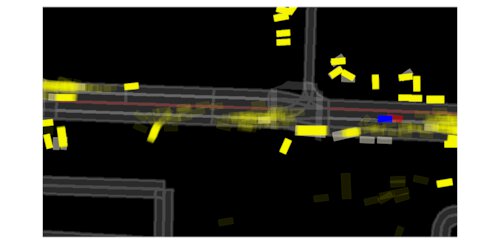}} &
        \raisebox{-0.5\height}{\includegraphics[trim={1cm .33cm 1cm .33cm},clip, width=0.33\textwidth]{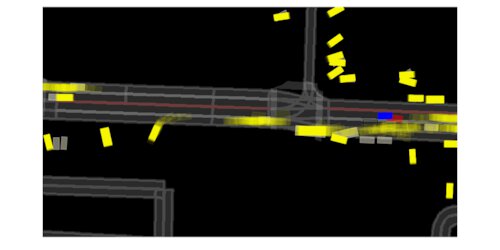}} &
        \raisebox{-0.5\height}{\includegraphics[trim={1cm .33cm 1cm .33cm},clip, width=0.33\textwidth]{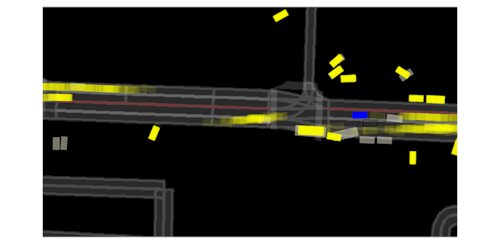}} 
        \\[7.5ex]
    \end{tabular}
    \caption{In this scenario, the 3 models generate pulling out trajectories for a big vehicle, forcing the SDV to maneuver to an unoccupied region. However, ILVM captures well the distribution of the rest of the actors and the SDV performs a safe left lane change. However, in ESP and MultiPath the trajectory of the vehicle to the left is not well captured and the SDV proceeds too aggressively, resulting in a collision.}
    \label{figure:planning_comparison_0}
    
\end{figure}

\begin{figure}[t]
    
    \begin{tabular}{@{}c@{\hspace{.1em}}c@{\hspace{.5em}}c@{\hspace{.5em}}c}
        \textbf{} &
        \textbf{\textsc{Multipath}} &
        \textbf{\textsc{ESP}} &
        \textbf{\textsc{ILVM}} \\

        \rotatebox[origin=c]{90}{\textbf{t = 0 s}} &
        \raisebox{-0.5\height}{\includegraphics[trim={1cm .33cm 1cm .33cm},clip, width=0.33\textwidth]{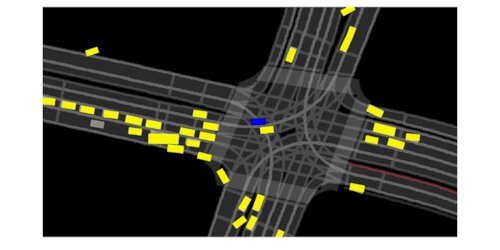}} &
        \raisebox{-0.5\height}{\includegraphics[trim={1cm .33cm 1cm .33cm},clip, width=0.33\textwidth]{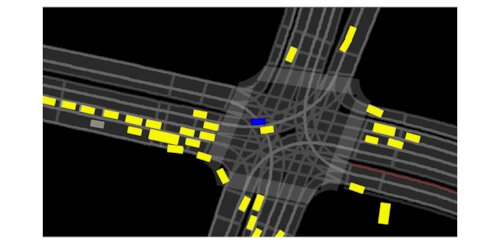}} &
        \raisebox{-0.5\height}{\includegraphics[trim={1cm .33cm 1cm .33cm},clip, width=0.33\textwidth]{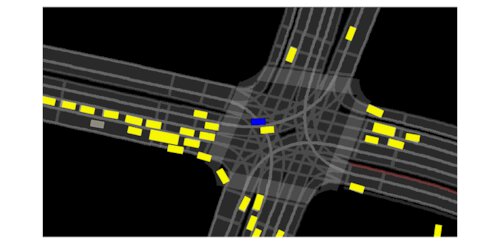}} 
        \\[7.5ex]
        
        \rotatebox[origin=c]{90}{\textbf{t = 1 s}} &
        \raisebox{-0.5\height}{\includegraphics[trim={1cm .33cm 1cm .33cm},clip, width=0.33\textwidth]{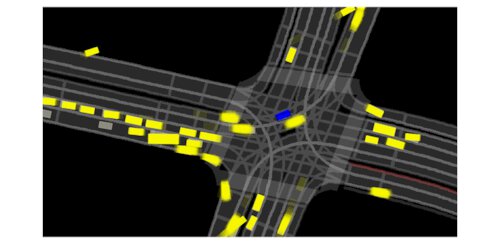}} &
        \raisebox{-0.5\height}{\includegraphics[trim={1cm .33cm 1cm .33cm},clip, width=0.33\textwidth]{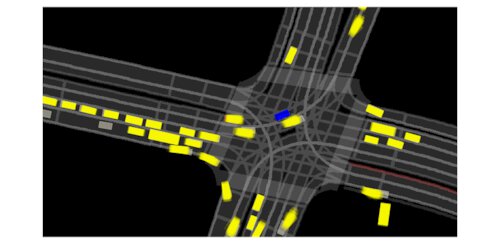}} &
        \raisebox{-0.5\height}{\includegraphics[trim={1cm .33cm 1cm .33cm},clip, width=0.33\textwidth]{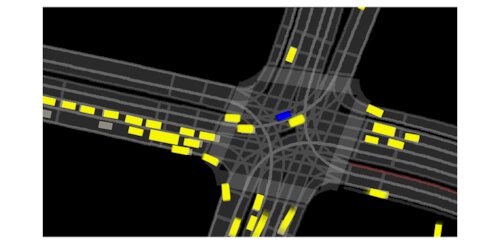}} 
        \\[7.5ex]

        \rotatebox[origin=c]{90}{\textbf{t = 2 s}} &
        \raisebox{-0.5\height}{\includegraphics[trim={1cm .33cm 1cm .33cm},clip, width=0.33\textwidth]{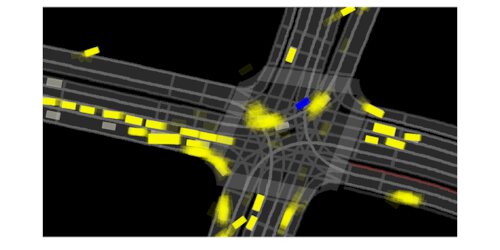}} &
        \raisebox{-0.5\height}{\includegraphics[trim={1cm .33cm 1cm .33cm},clip, width=0.33\textwidth]{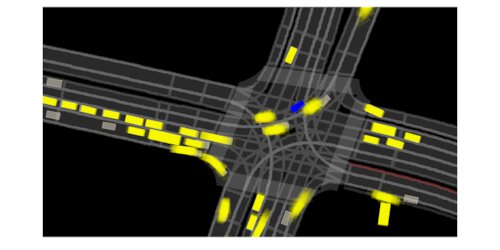}} &
        \raisebox{-0.5\height}{\includegraphics[trim={1cm .33cm 1cm .33cm},clip, width=0.33\textwidth]{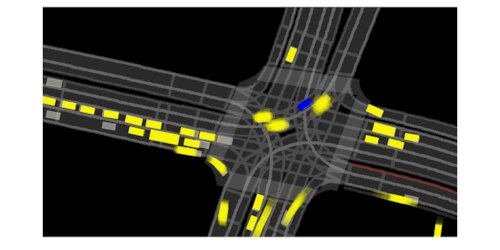}} 
        \\[7.5ex]

        \rotatebox[origin=c]{90}{\textbf{t = 3 s}} &
        \raisebox{-0.5\height}{\includegraphics[trim={1cm .33cm 1cm .33cm},clip, width=0.33\textwidth]{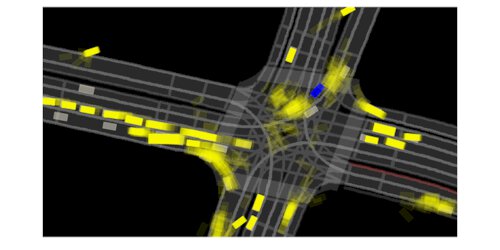}} &
        \raisebox{-0.5\height}{\includegraphics[trim={1cm .33cm 1cm .33cm},clip, width=0.33\textwidth]{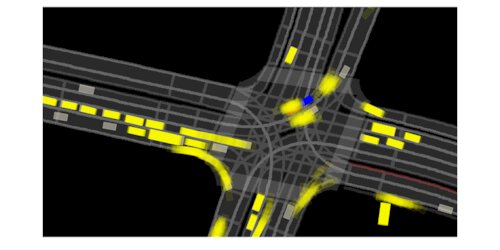}} &
        \raisebox{-0.5\height}{\includegraphics[trim={1cm .33cm 1cm .33cm},clip, width=0.33\textwidth]{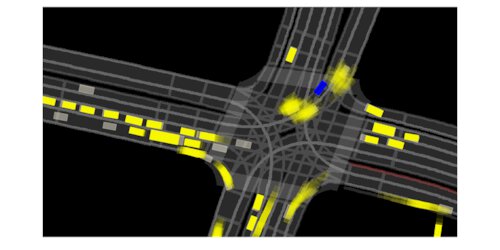}} 
        \\[7.5ex]

        \rotatebox[origin=c]{90}{\textbf{t = 4 s}} &
        \raisebox{-0.5\height}{\includegraphics[trim={1cm .33cm 1cm .33cm},clip, width=0.33\textwidth]{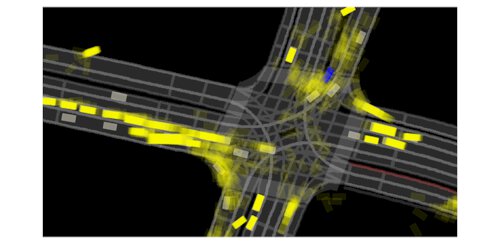}} &
        \raisebox{-0.5\height}{\includegraphics[trim={1cm .33cm 1cm .33cm},clip, width=0.33\textwidth]{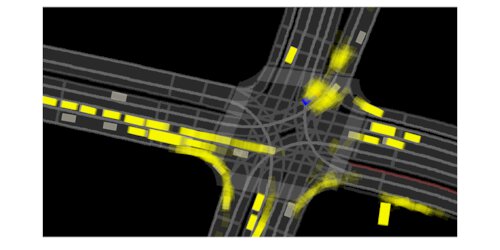}} &
        \raisebox{-0.5\height}{\includegraphics[trim={1cm .33cm 1cm .33cm},clip, width=0.33\textwidth]{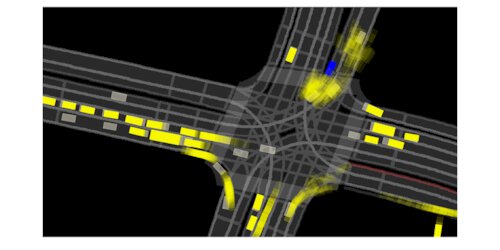}} 
        \\[7.5ex]

        \rotatebox[origin=c]{90}{\textbf{t = 5 s}} &
        \raisebox{-0.5\height}{\includegraphics[trim={1cm .33cm 1cm .33cm},clip, width=0.33\textwidth]{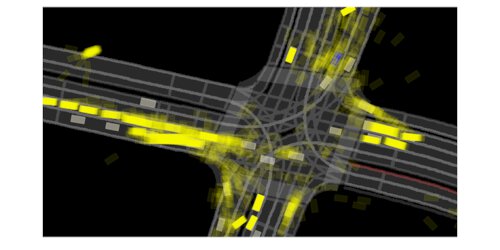}} &
        \raisebox{-0.5\height}{\includegraphics[trim={1cm .33cm 1cm .33cm},clip, width=0.33\textwidth]{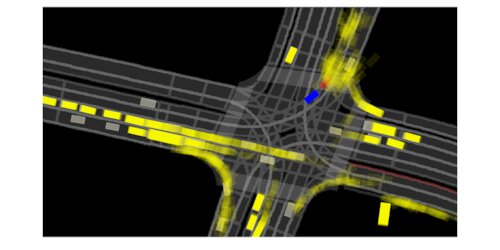}} &
        \raisebox{-0.5\height}{\includegraphics[trim={1cm .33cm 1cm .33cm},clip, width=0.33\textwidth]{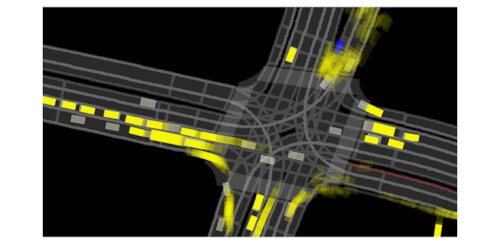}} 
        \\[7.5ex]
    \end{tabular}
    \caption{ESP predicts that the vehicle that starts at the right of the SDV is going to cut-off the SDV by lane changing left, causing the SDV to hard break and causing a collision with the vehicle behind. MultiPath and ILVM successfully drive through the scenario, even though we can see how MultiPath's prediction go even into opposite traffic, but luckily do not interfere the SDV.}
    \label{figure:planning_comparison_2}
    
\end{figure}

In Figs.~\ref{figure:planning_comparison_0}, \ref{figure:planning_comparison_1}, \ref{figure:planning_comparison_2}, 
we show qualitative comparison between ego-motion planning open-loop results when using motion forecast from some of the strongest baselines and our model.
Open-loop means that the SDV acts as if it does not receive new sensor information for the future horizon of 5 seconds, and thus needs to rely completely on the motion forecasts at the start of these scenarios. Thus, many of the collisions on these results could be potentially avoided by obtaining more accurate information in subsequent time steps and re-planning, but closed-loop experiments are out of scope of this paper.

The predicted bounding box samples into the future for other traffic participants are shown in yellow. The ground-truth future trajectories are shown in white if not in collision with the ground-truth SDV trajectory (shown as an empty black box) and in red if colliding with the SDV plan.
Overall, we can see how ego-vehicle harmful events are avoided with more precise motion forecasts from our model.
In particular, we observe that the main reason the baseline motion forecasting models tend to cause more collisions than our predictions is because the entropy of their distributions is too high, leaving the motion planner no space to plan a safe trajectory.

\end{document}